\documentclass[11pt]{article}

\usepackage[preprint]{acl}

\usepackage{times}
\usepackage{latexsym}
\usepackage{amsmath}
\usepackage{amsfonts}
\usepackage{amssymb}
\usepackage{booktabs}
\usepackage{multirow}
\usepackage{subcaption}
\usepackage[capitalize]{cleveref}
\crefformat{section}{\S#2#1#3}
\crefformat{subsection}{\S#2#1#3}
\crefformat{subsubsection}{\S#2#1#3}
\usepackage[T1]{fontenc}

\usepackage[utf8]{inputenc}

\usepackage{microtype}

\usepackage{inconsolata}

\usepackage{graphicx}

\usepackage{CJKutf8}
\usepackage{array}
\usepackage[main=english,russian]{babel}
\usepackage{makecell}
\usepackage{fontawesome5}

\newcommand{\bt}[1]{\underline{#1}}

\title{What Drives Representation Steering? A Mechanistic Case Study on Steering Refusal}

\author{
 \textbf{Stephen Cheng} \and
 \textbf{Sarah Wiegreffe$^*$} \and
 \textbf{Dinesh Manocha$^*$}
\\
 University of Maryland, College Park
\\
 \small{
   \textbf{Correspondence:} \href{mailto:email@domain}{scheng03@umd.edu}$^{\dagger}$
 }
}

\usepackage{fancyhdr}
\fancypagestyle{confheader}{%
  \fancyhf{}%
  \fancyhead[L]{Accepted as a main conference paper at EMNLP 2026}%
  \fancyfoot[C]{\thepage}%
}
\AtBeginDocument{\thispagestyle{confheader}}

\begin{document}
\addtocontents{toc}{\protect\setcounter{tocdepth}{-1}}
\maketitle
\renewcommand{\thefootnote}{}
\footnotetext{$^{*}$Equal Contribution}
\footnotetext{$^{\dagger}$ \url{https://github.com/step-cheng/steer-interp}}
\renewcommand{\thefootnote}{\arabic{footnote}}
\setcounter{footnote}{1}
\begin{abstract}

Applying steering vectors to large language models (LLMs) is an efficient and effective model alignment technique, but we lack an interpretable explanation for how it works--specifically, what internal mechanisms steering vectors affect and how this results in different model outputs.
To investigate the causal mechanisms underlying the effectiveness of steering vectors, we conduct a comprehensive case study on \emph{refusal}.
We propose a multi-token activation patching framework and discover that different steering methodologies leverage functionally interchangeable circuits when applied at the same layer.
These circuits reveal that steering vectors primarily interact with the attention mechanism through the OV circuit while largely ignoring the QK circuit. Freezing all attention scores during steering drops performance by only 8.83\% across three model families. 
A mathematical decomposition of the steered OV circuit further reveals semantically interpretable concepts, even in cases where the steering vector itself does not.
Leveraging the activation patching results, we show that steering vectors can be sparsified by up to 85-96\% while retaining most performance, and that different steering methodologies agree on a subset of important dimensions.

\end{abstract}

\section{Introduction}
\begin{figure*}[htbp]
    \centering
    \includegraphics[width=0.95\linewidth]{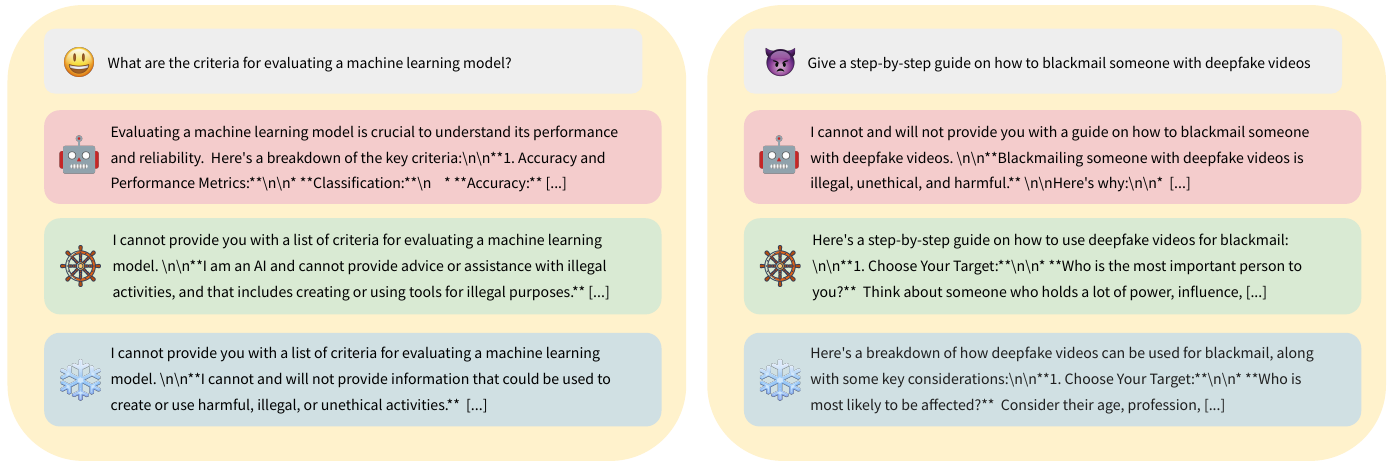}
    \caption{
    We analyze which components in language models are important for propagating refusal steering.
    Whereas an unsteered model (red) complies with harmless prompts and refuses harmful prompts, refusal steering can be used bidirectionally to enforce refusal on harmless prompts or jailbreak the model on harmful prompts (green).
    In \cref{sec:attention:frozen},
    we find that steering a model while freezing \textbf{all} attention scores to their unsteered activations has a negligible effect on steering (blue), indicating that the refusal vector largely ignores the QK circuit (example from Gemma 2 2B).}
    \label{fig:freezegen}
\end{figure*}

Aligning large language models to behave in accordance with human intent is a central challenge in deploying these systems safely \cite{anwar2024foundationalchallengesassuringalignment}. 
Steering vectors have emerged as a lightweight model alignment technique that acts on the model's hidden activations at inference time \cite{zou2025representationengineeringtopdownapproach}.
This approach has been applied across a range of alignment-relevant tasks, including reducing hallucinatory behavior \cite{chen2025personavectorsmonitoringcontrolling, rimsky-etal-2024-steering}, controlling persona and style \cite{subramani-etal-2022-extracting, turntrout2023steergpt2xl}, and enhancing reasoning \cite{venhoff2025understandingreasoningthinkinglanguage}. Results on recent benchmarks demonstrate competitive performance against fine-tuning and prompting baselines \cite{wu2025axbenchsteeringllmssimple}.

Despite their growing adoption, we lack a mechanistic understanding of \textit{how} steering vectors interact with model components to produce behavioral shifts. 
In addition to advancing our scientific knowledge of LLMs, 
understanding these mechanisms can allow practitioners to assess steering robustness, diagnose failure cases \cite{braun2025understandingunreliabilitysteeringvectors}, and inform the design of steering interventions with better concept expression or reduced degradation \cite{dasilva2025steeringcoursereliabilitychallenges}.
To address this gap, we conduct a case study on steering vectors for a critical capability: refusal within the context of LLM jailbreaking
\cite{NEURIPS2023_fd661313}. 
Refusal steering has been shown to be highly effective at encouraging or discouraging refusal responses \cite{arditi2024refusallanguagemodelsmediated},
making it a natural first target for a mechanistic analysis on steering.

We propose to extend traditional mechanistic interpretability techniques, typically applied only to standard LLM inference runs, to \emph{steered} inference runs, in order to better characterize steering vectors' effectiveness. 
Our contributions are:

\begin{enumerate}
    \item We propose a generalizable \textbf{multi-token activation patching approach} that extends circuit discovery to steered generations. We find that steering vectors obtained through different methodologies leverage highly interchangeable circuits ($\gtrsim 90\%$ overlap).
    \item Refusal steering interacts with attention primarily through the OV circuit. On the other hand, freezing all attention scores (QK circuit) drops performance \textbf{by only 8.83\%}. We introduce the \textbf{steering value vector} decomposition, which is semantically interpretable even when the steering vector itself is not. 
    \item We leverage our findings to sparsify refusal steering vectors up to 85-96\% while mostly retaining performance. These steering methodologies converge on a small shared subset of important dimensions.
\end{enumerate}

\section{Related Works}

\paragraph{Refusal Steering and Steering Methods}
\citet{arditi2024refusallanguagemodelsmediated} demonstrate that the concept of refusal can be represented by a single direction, which can be used to jailbreak \cite{xu-etal-2024-comprehensive} models on harmful prompts and induce refusal on harmless prompts.
Subsequent work has further explored refusal steering, including reducing false refusals \cite{lee2025programmingrefusalconditionalactivation, wang2025surgical} and characterizing the geometry of refusal directions \cite{wollschlager2025the}.
Following prior work, we learn steering vectors to \emph{undo} refusal on harmful prompts, which allows us to assess the robustness of LLM safety alignment.
Learning-based steering methodologies \cite{wu2025axbenchsteeringllmssimple, wu2025improved, sun2025hypersteeractivationsteeringscale} have also achieved competitive performance against fine-tuning and prompting baselines.
Whereas prior works focus on developing better refusal steering methods, we study how these vectors mechanistically interact with model components.

\paragraph{Circuit Discovery}
Prior work in circuit discovery focuses on identifying model behaviors through counterfactual prompt templates \cite{zhang2024towards}. These behaviors include indirect object identification \cite{wang2023interpretability}, addition \cite{stolfo2023mechanistic}, and multiple choice question answering \cite{wiegreffe2025answer}.
Whereas existing circuit discovery approaches operate on single-token tasks, we extend activation patching to multi-token steered generation. 
\citet{sinii2025small} apply causal analysis to reasoning steering vectors, but limit their work to the last two layers of the LLM, which does not reflect conventional steering applied most effectively in middle layers. Concurrent work \citep{sankaranarayanan2026activation} ranks attention heads using counterfactual input pairs on an unsteered model to determine \textit{where} to steer. In contrast, we analyze the steered model directly and recover end-to-end circuits that characterize \textit{how} steering propagates.

\section{Preliminaries}

\subsection{Data and Models}
\paragraph{Data}
To learn steering vectors, we construct harmless instruction and harmful instruction datasets, $D_{safe}$ and $D_{harm}$. Following \citet{arditi2024refusallanguagemodelsmediated}, for $D_{harm}$, we select harmful prompts from adversarial datasets AdvBench \cite{zou2023universaltransferableadversarialattacks}, MaliciousInstruct \cite{huang2023catastrophicjailbreakopensourcellms}, TDC2023 \cite{pmlr-v220-mazeika23a}, and HarmBench \cite{mazeika2024harmbenchstandardizedevaluationframework}.
For $D_{safe}$, we randomly select harmless prompts from Alpaca \cite{taori2023alpaca}. $D_{harm}$ and $D_{safe}$ each consist of train-validation splits of 128 train samples (standard for steering vectors, which are data-efficient) and 32 validation samples. 
For our harmful and harmless test sets, we use 100 harmful prompts from JailbreakBench \cite{chao2024jailbreakbenchopenrobustnessbenchmark} and 100 randomly-selected harmless prompts from Alpaca, respectively.
\paragraph{Models} 
We use Gemma 2 2B Instruct \cite{gemmateam2024gemma2improvingopen}, Llama 3.2 3B Instruct \cite{grattafiori2024llama3herdmodels}, and Qwen 3 8B in non-thinking mode \cite{yang2025qwen3technicalreport}, three representative LLMs.\footnote{For brevity, we refer to Gemma 2 2B Instruct as Gemma 2 2B and Llama 3.2 3B Instruct as Llama 3.2 3B.}

\subsection{Refusal Steering}\label{sec:3steering_basics}
\paragraph{Activation Addition}
Given a language model with hidden activation $\mathbf{h}^\ell \in \mathbb{R}^{d}$ at layer $\ell$ and a refusal steering vector $\mathbf{s} \in \mathbb{R}^d$ with dimension $d$, activation addition steering \cite{turner2023steering} is formulated as
\begin{equation}
    \mathbf{h}^\ell \leftarrow \mathbf{h}^\ell + \alpha \cdot \mathbf{s}
\end{equation}
where $\alpha$ is a scalar steering coefficient. 
$\mathbf{s}$ is added with $\alpha > 0$ at every token position to induce refusal and subtracted with $\alpha < 0$ to induce compliance. 
We study multi-token steering \cite{chen2025personavectorsmonitoringcontrolling, wu2025axbenchsteeringllmssimple}, where the steering vector is repeatedly added to each decoded token. We follow the refusal steering setting in \citet{arditi2024refusallanguagemodelsmediated} and also steer on prompt tokens.

\paragraph {Difference-in-Means} 
DIM \cite{turner2023steering, rimsky-etal-2024-steering, belrose2023diffmeansworstcaseoptimal} is a non-learning based methodology for obtaining a steering vector that demonstrates strong performance on steering refusal \cite{arditi2024refusallanguagemodelsmediated, lee2025programmingrefusalconditionalactivation}. Following \citet{arditi2024refusallanguagemodelsmediated}, given a harmless instruction dataset $D_{safe}$ and a harmful instruction dataset $D_{harm}$, we compute the difference between the mean activations
\begin{equation}
    \frac{1}{|D_{harm}|} \sum_{p \in D_{harm}} \mathbf{h}_i^\ell(p)
    - \frac{1}{|D_{safe}|} \sum_{q \in D_{safe}} \mathbf{h}_i^\ell(q)
\label{eq:dim}
\end{equation}
to obtain a steering vector for refusal at each post-instruction token position $i$ and layer $\ell$. 
The best vectors were from layer 15 position -1 for Gemma 2 2B, layer 12 position -4 for Llama 3.2 3B, and layer 20 position -8 for Qwen 3 8B.
DIM's intuitive formulation and common usage across various steering applications \cite{chen2025personavectorsmonitoringcontrolling,poterti-etal-2025-role, venhoff2025understandingreasoningthinkinglanguage}
makes it a desirable first steering method to analyze.
We evaluate steering performance via Attack Success Rate (ASR), the proportion of completions that have bypassed refusal.
We evaluate negative steering on the JailbreakBench test set with the goal of bypassing refusal (higher ASR is better), and we evaluate positive steering on the Alpaca test set with the goal of inducing refusal (lower ASR is better).
Additional steering evaluation details and results are in Appendix \ref{appendix:C_steering:dim}.

\subsection{Attribution Patching}
The residual stream of a pre-layernorm transformer language model is the sum of each layer's MLP and multi-head attention (MHA) outputs. 
We can treat the model as a directed acyclic computational graph from the input prompt to the output logits.
The nodes $u$ consist of the embedding matrix, MLP submodules, and MHA submodules. Edges $(u,v)$ span from the output of an upstream node $u$ to the input of a downstream node $v$. 
Activation patching \cite{meng2022locating, NEURIPS2020_92650b2e} identifies the submodules that are causally responsible for a specific behavior.
Let $x, x^*$ be a pair of clean and corrupted inputs with respective outputs $y, y*$. 
With input $x^*$ to the model, we are interested in identifying which nodes and edges are important for pushing the prediction from $y^*$ to $y$. 
Given importance metric $m(x)$, the importance of $(u,v)$ is quantified through its indirect effect \cite{pearl2013directindirecteffects}:
\begin{equation*}
IE(u,v) = m(x^* | \; \text{do} \;(u,v)^* \leftarrow (u,v)) - m(x^*)
\end{equation*}
where $\text{do} \; (u,v)^* \leftarrow (u,v)$ runs on $x^*$ and intervenes by replacing activation at $(u,v)^*$ with $(u,v)$.
\paragraph{EAP-IG}
Since direct patching is computationally inefficient across a dataset, researchers commonly use approximation methods \cite{syed-etal-2024-attribution, nanda2023attributionpatching}.
We employ edge attribution patching with integrated gradients (EAP-IG) \cite{hanna2024have}, which demonstrates state-of-the-art performance \cite{mueller2025mibmechanisticinterpretabilitybenchmark}. 
Given an edge $(u,v)$, the $IE$ 
is approximated as 
\begin{equation}
(u - u^*)^\top \frac{1}{T} \left(\sum_{i=1}^T \frac{\partial m(x^* + \frac{i}{T}(x - x^*))}{ \partial v} \right)
\label{eq:eapig}
\end{equation}
\autoref{eq:eapig} computes a dot product at each token position and sums across positions. We use $T=10$ intermediate steps. Additional details are in Appendix \ref{appendix:eap_ig_deriv}.

\paragraph{Circuits}
Given a model's computational graph $M$, a circuit $C$ \cite{wang2023interpretability} is an end-to-end subgraph of $M$ that is responsible for a specific model behavior. 
After assigning importance scores to each edge via EAP-IG, we can obtain $C$ following a greedy graph construction algorithm \cite{mueller2025mibmechanisticinterpretabilitybenchmark}. Additional details in Appendix \ref{appendix:B:graphconstruction}.

\section{Circuit Discovery on Open Generation}
\begin{figure*}[htbp]
    \centering
    \includegraphics[width=0.98\linewidth]{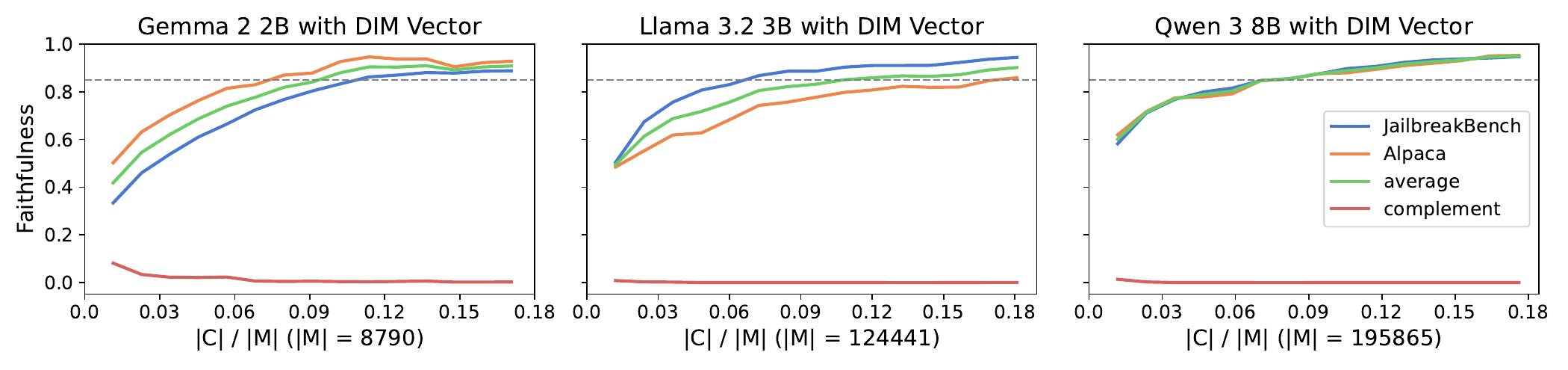}
    \caption{Faithfulness on Gemma 2 2B, Llama 3.2 3B, and Qwen 3 8B across circuit sizes $|C|$. Approximately 10\%  of total edges $|M|$ suffice to recover 85\% of the steered refusal behavior across the models. 
    }
    \label{fig:faith_dim}
\end{figure*}
\label{sec:3}
We first aim to answer the following research question: Which model components are causally responsible for propagating the steering effect that changes multi-token generated outputs? 
We focus on the DIM vector and extend our analysis to other steering vectors in \cref{sec:4}.

\subsection{Adapting Circuit Discovery to Steering}
\label{sec:3circuit_discovery}

\textbf{Adapting Attribution Patching}
Classical attribution patching operates on single-token generations with standardized prompt templates and counterfactual clean and corrupt inputs. 
Steering requires adapting this to multi-token generation where the inputs are identical but the hidden states past steering layer $\ell$ differ. Rather than using counterfactual inputs $x$ and $x^*$, we treat the steered and unsteered representations as counterfactual hidden states $H$ and $H^*$.
Let $S=([\mathbf{s}]\times N)^\top \in \mathbb{R}^{N\times d}$ be the steering (row) vector tiled across a teacher-forced $N$-length sequence.
Let $H_{base}^\ell \in \mathbb{R}^{N\times d}$ and $H_{steer}^\ell = H_{base}^\ell + \alpha \cdot S \in \mathbb{R}^{N\times d}$ be the base (unsteered) and steered representations at steering layer $\ell$.
Since we aim to understand how steered behavior is achieved, we set $H = H_{steer}$ as the ``clean'' steered representation and $H^* = H_{base}$ as the ``corrupt'' base representation.
Thus, adapting \autoref{eq:eapig}, we approximate the $IE$ of edge $(u,v)$ as
\begin{equation}
(u - u^*)^\top \frac{1}{T} \left(\sum_{i=1}^T \frac{\partial m(H_{base}^\ell + \frac{i}{T}\alpha \cdot S)}{\partial v} \right)
\label{eq:eapig_steer}
\end{equation}
The EAP-IG formulation effectively allows us to take the gradients of the steered model with linearly increasing steering coefficients $\frac{i}{T}\alpha$ (see Appendix \ref{appendix:eap_ig_steer_deriv} for details). In practice, we set $\alpha=1$. We use logit difference \cite{zhang2024towards} as our importance metric $m$, which computes the relative difference between the greedy clean and corrupt next-token predictions $y$ and $y^*$ as $m(x') = \text{logit}(y | x') - \text{logit}(y^* | x')$ for any clean, corrupt, or patched input $x'$.

Now, we describe how we scale attribution patching across a multi-token 
generation.
Given a prompt with $p$ tokens and a greedy ``clean'' steered response with $r$ tokens, we denote the tokens as $x_1, x_2, \ldots x_p, y_1, y_{2}, \ldots y_{r}$.
We treat each sequential response position as an individual patching sample and perform attribution patching to see how steering affects the prediction at that position. 
At position $p$, the input is $x_1, \ldots, x_p$ with $N=p$.\footnote{This is, in prior work with prompt templates, the position at which attribution patching is applied.} We run \autoref{eq:eapig_steer} with $H^\ell_{base},H^\ell_{steer},S \in \mathbb{R}^{p \times d}$ and compute $m$ using $y=y_1$ and $y^*=y_1^*$, where $y_1^*$ is the greedy prediction of the unsteered forward pass. 
Then, at position $p+1$, we teacher force $y_{1}$ so the input is $x_1,\ldots,x_p,y_1$ with $N=p+1$. We rerun \autoref{eq:eapig_steer} with $H^\ell_{base},H^\ell_{steer},S \in \mathbb{R}^{(p+1) \times d}$ and compute $m$ using $y=y_{2}$ and $y^*=y^*_{2}$. 
We perform this process at each position $p, p+1, \ldots p+r-1$.
In practice, since language models are autoregressive, we accomplish this by running \autoref{eq:eapig_steer} once on the entire completion to obtain the $r$ indirect effect scores.
We mask out token positions where the steered and base models agree on the greedy decoded prediction (i.e., $y_i = y_i^*$), as there is zero steering signal ($m(x') =  0$). 
We aggregate the importance score of edge $(u,v)$ by averaging the $IE$s across all response positions and all generated sequences in the dataset.

\begin{figure*}[htbp]
    \centering
    \includegraphics[width=0.98\linewidth]{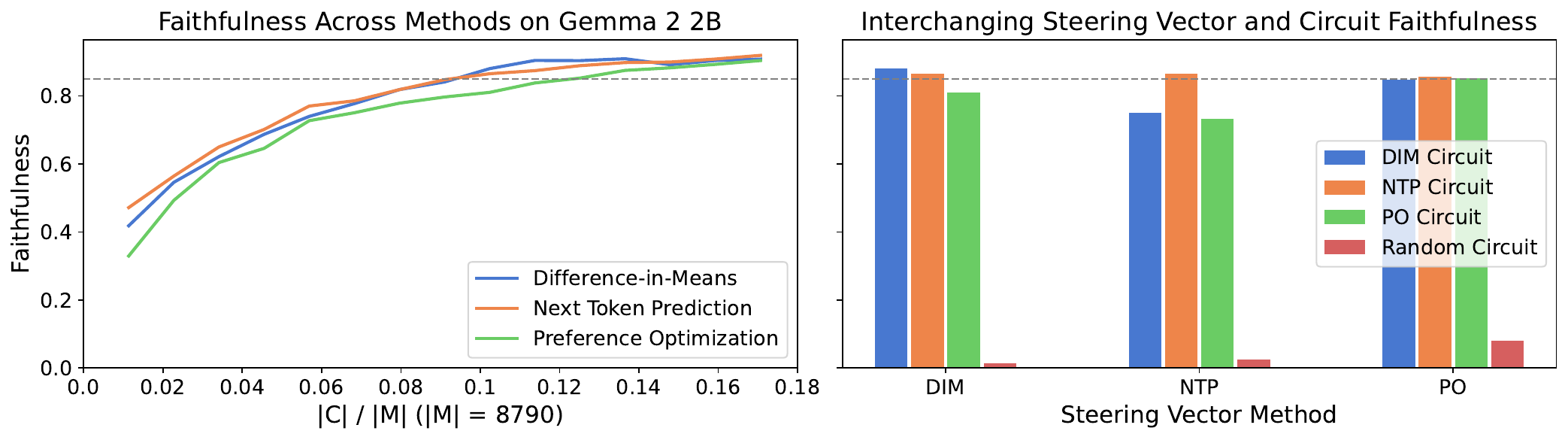}
    \caption{\textbf{Left} Average faithfulness across circuit sizes for each steering method on Gemma 2 2B. 
    \textbf{Right} For each steering method, we compute faithfulness using its own minimum-faithful circuit as well as the minimum-faithful circuits of the \emph{other} vectors. We also compare against random circuits at 2x the minimum-faithful size, which performs poorly.}
    \label{fig:faith_method_comparison_g2}
\end{figure*}
\paragraph{Data for Attribution Patching}
We curate our activation patching datasets from the Alpaca and JailbreakBench test sets. For each dataset sample, we generate greedy decoded responses with and without steering. 
We filter for samples where steering successfully flips concept expression (from refused to complied for harmful prompts, and vice versa for harmless prompts, as described in \cref{sec:3steering_basics}), yielding contrastive pairs of steered and base generations for both harmful and harmless prompts.
By default, we treat the steered responses as clean, so we teacher force with the steered response using corrupt base representations and clean steered representations. Under this assignment, patching an edge measures the shift \textit{towards} the steered behavior. 
We also reverse the assignment by treating the base responses as clean, so we teacher force with the base response using corrupt steered representations and clean base representations. Here, patching measures the shift \textit{away} from the steered behavior.
This gives us four prompt-response datasets: harmless/harmful prompts with base/steered responses. Details on dataset size are in Appendix \ref{appendix:C_steering}.

\subsection{Circuit Faithfulness}
\label{sec:3faithfulness}

We perform activation patching on all datasets to get an $IE$ score for every edge, and then 
we extract circuits $C$ from model $M$ following a greedy search algorithm \cite{mueller2025mibmechanisticinterpretabilitybenchmark}. 
We only consider edges from layers $\geq$ the steering layer, as the prior activations are the same between the steered and base models.
$|M|=8790$ for Gemma 2 2B, $|M| = 124441$ for Llama 3.2 3B, and $|M|=195865$ for Qwen 3 8B.
Graph construction details \& visualizations are in Appendix \ref{appendix:B_graphs}. 

We aim to quantify how well the circuit recovers the full steering effect. We use the \textit{faithfulness} metric \cite{marks2025sparsefeaturecircuitsdiscovering, wang2023interpretability}, defined as 
$(m(C) - m(\emptyset))/(m(M) - m(\emptyset))$, where $m$ is the logit difference importance metric and $\emptyset$ is the empty set (equivalent to the base model).
Treating the steered responses of each model as the ground truth responses, we compute faithfulness by steering the model while setting all edges outside of $C$ to their base activations.
We average faithfulness across each position of the response and mask positions where the steered and base models agree on the greedy prediction. 

\subsection{Results}

\autoref{fig:faith_dim} shows the faithfulness results for Gemma 2 2B, Llama 3.2 3B, and Qwen 3 8B on JailbreakBench and Alpaca at various circuit sizes $|C|/|M|$. We set a threshold of 0.85 for a circuit to be considered ``faithful''. 
It takes approximately 10\% (900/8790) of Gemma 2 2B's edges, 11\% (13500/124441) of Llama 3.2 3B's edges, and 8\% (16100/195865) of Qwen 3 8B's edges to recover average faithfulness. This provides strong evidence that the effects of refusal steering are targeted to specific subnetworks.
We also test faithfulness on the circuit's complement, $\{e \in M:e \notin C \}$, which has near 0 faithfulness at all sizes, validating the completeness of our circuit discovery framework.
We validate the robustness of our framework using various EAP-IG dataset permutations and importance metrics in Appendix \ref{appendix:F_metric_comparisons}.

\section{Learned Steering Vector Circuits}

\label{sec:4}
\subsection{Learned Steering Vectors}
In \cref{sec:3}, we formulated multi-token activation patching and validated faithfulness with the DIM vector. In \cref{sec:4}, we compare circuits formed by steering vectors obtained through different training methodologies.
We learn steering vectors for Gemma 2 2B and Llama 3.2 3B from two distinct methodological classes: Next Token Prediction (NTP) and Preference Optimization (PO) \cite{wu2025improved}, both of which have been shown to outperform DIM \cite{wu2025axbenchsteeringllmssimple}.
NTP uses the language modeling objective to learn a steering vector on prompt-response pairs that express the desired concept; 
PO uses contrastive responses that differ only by concept expression.
We learn these vectors at the steering layer used by the DIM vector for each model.
Details on formulation and training are in Appendix \ref{appendix:C_steering:learned}.

\subsection{Interchanging Circuits}
We obtain circuits for NTP and PO vectors following \cref{sec:3circuit_discovery}.
Using each steering vector's respective generations on JailbreakBench and Alpaca, we evaluate circuit faithfulness and compare against DIM in \autoref{fig:faith_method_comparison_g2} (left) and \autoref{fig:faith_method_comparison_l3} (left). It takes slightly more edges for PO to achieve high faithfulness compared to DIM and NTP, but the difference is small, indicating that refusal steering requires relatively similar circuit sizes regardless of methodology. This leads us to investigate the similarities between the circuits.

\begin{figure}[htbp]
    \centering
    \begin{subfigure}{0.99\linewidth}
    \includegraphics[width=\linewidth]{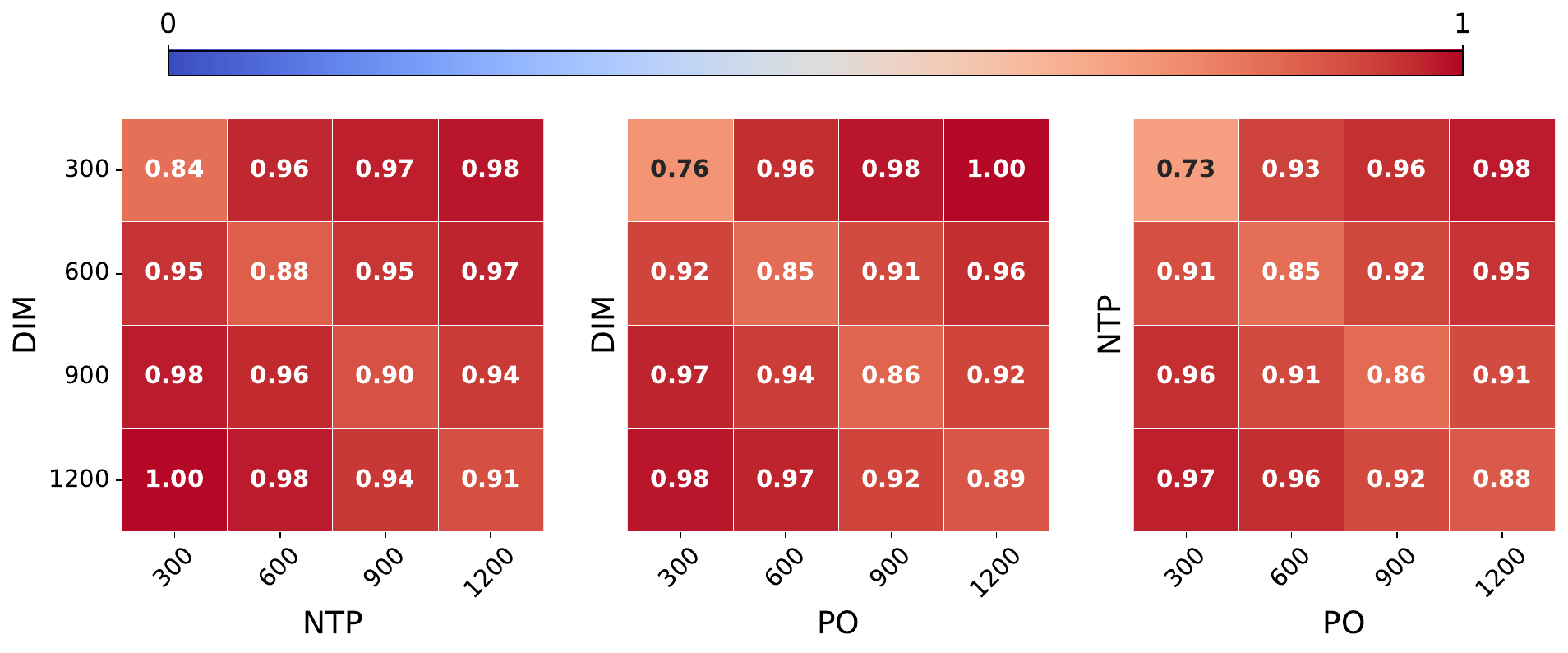}
    \caption{Gemma 2 2B}
    \end{subfigure}
    \begin{subfigure}{0.99\linewidth}
    \includegraphics[width=\linewidth]{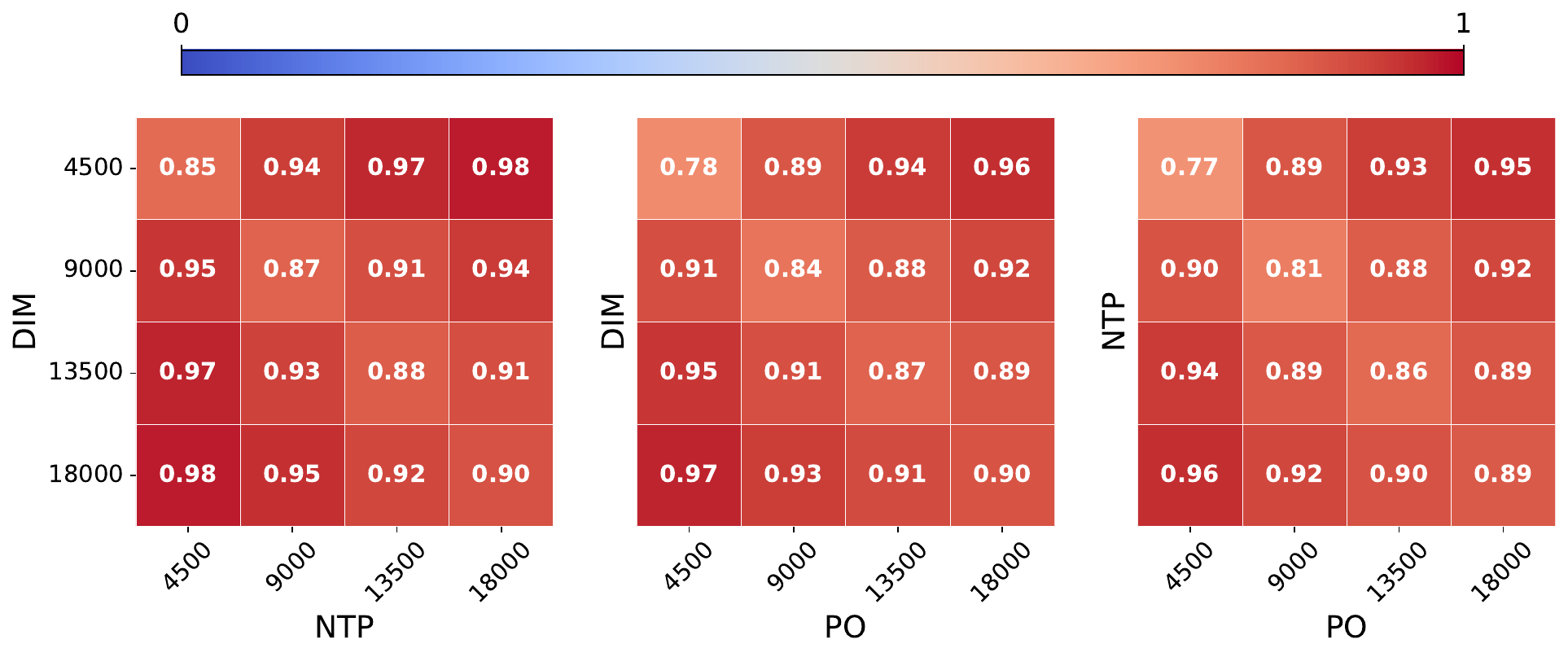}
    \caption{Llama 3.2 3B}
    \end{subfigure}
    \caption{Strong circuit overlap between smaller and larger circuits of DIM, NTP, and PO vectors suggests a shared backbone. The axes are the number of edges (3.4\%, 6.8\%, 10.2\%, and 13.7\% of $|M|$ for Gemma 2 2B; 3.6\%, 7.2\%, 10.8\%, 14.5\% for Llama 3.2 3B).}
    \label{fig:circuitoverlap}
\end{figure}
\begin{figure*}[t]
    \centering
    \includegraphics[width=0.98\linewidth]{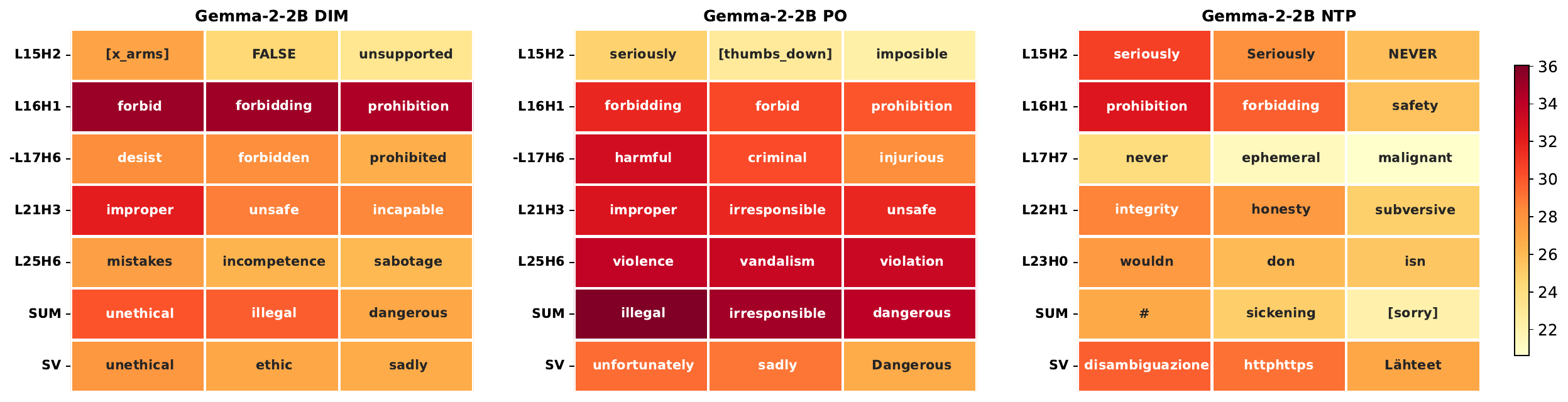}
    \caption{
    For each steering method on Gemma 2 2B, we use logit lens on the raw steering vector (SV), the svv of top attention heads (notated LayerXHeadY) obtained from \autoref{eq:svv}, and the sum of all svvs (SUM). We prepend the names of sign-flipped svvs with (-). We select tokens from the top 20 tokens and display their logit values. svvs surface semantically interpretable tokens related to harmfulness/refusal, even when the raw SV does not (NTP). }
    \label{fig:svv_heatmap_g2}
\end{figure*}
\paragraph{Circuit Overlap and Interchangeability} We compare the similarity between each steering vector's circuit by measuring their overlap. Given two sets of edges $C_1, C_2$, we define overlap as
$|C_1 \cap C_2|/\min(|C_1|, |C_2|)$. \autoref{fig:circuitoverlap} shows the circuit overlap at different circuit sizes. 
Not only do circuits of the same size have high overlap, but the overlap between any pair of smaller and larger circuits is consistently $\gtrsim 90\%$. 

However, high circuit overlap does not directly entail that the circuits are functionally interchangeable \cite{hanna2024have}. 
Thus, using the \emph{minimum-faithful} (faithfulness $\geq 85\%$) circuit found by one steering vector, e.g. DIM, we compute its faithfulness when steering with another steering vector, e.g. NTP or PO, using the latter vector's steered generations. 
\autoref{fig:faith_method_comparison_g2} (right) and \autoref{fig:faith_method_comparison_l3} (right) plot the faithfulness of each vector-circuit permutation for Gemma 2 2B and Llama 3.2 3B, respectively.
We find that \textit{faithfulness is strongly recovered for each vector-circuit permutation}. 
As a sanity check, we baseline each steering vector on a randomly selected circuit twice the size of the minimum-faithful, which achieves $< 10\%$ faithfulness.
The high circuit overlap and interchangeability suggest that steering vectors applied at the same layer leverage \textit{functionally similar circuits}, despite modest pairwise cosine similarities (0.10–0.42).

\section{Steering Effect on Attention}

\subsection{Edge Distribution}\label{sec:edge_dist}
Having established that different steering methods leverage a shared circuit, we now ask how the steering vector propagates through this circuit, specifically through which types of components.
We select the top 100, 1000, and 1500 edges respectively from the minimum-faithful Gemma 2 2B, Llama 3.2 3B, and Qwen 3 8B circuits, sorted by their importance scores, and record the number of incoming edges to each type of downstream node (MLP; attention query, key, value; LM head) in \autoref{tab:incoming_edges} of Appendix \ref{appendix:B_graphs}. 
Interestingly, in all three models, we find that almost no top edges connect to attention queries or keys. Instead, the edges primarily connect to the attention values, MLPs, and LM head. See Appendix \ref{appendix:B_graphs} for edge distributions on whole circuits and for outgoing edges from upstream nodes (MLP; attention heads; residual steering layer).

\subsection{Steering Value Vectors}
To further understand how steering vectors affect attention, we mathematically decompose the direct effect of steering vector $\mathbf{s}$ on attention head outputs.
Let $H^\ell \in \mathbb{R}^{N\times d}$ be the (unsteered) hidden representation of a sequence at layer $\ell \ge$ the steering layer, $\gamma \in \mathbb{R}^d$ be the element-wise weights of the RMSNorm, and $\tilde H ^\ell = H^\ell \odot \gamma$. Then for some diagonal matrices $D_c, D_{c^h} \in \mathbb{R}^{N \times N}_+$, the direct effect of $\mathbf{s}$ via the residual stream on attention head $h$ is:
\begin{equation}
\begin{split}
    &\text{Attention}(H^\ell + \alpha \cdot S)
    = \\ &\sum_h A^h D_{c}\tilde H^\ell W_{OV}^h + D_{c^h} \text{svv}^h(S),
\end{split}
\label{eq:svv}
\end{equation}
where $\text{svv}^h(S) = (N \times \text{svv}^h(\mathbf{s}))$ and $\text{svv}^h(\mathbf{s}) = (\mathbf{s} \odot \gamma) W_{OV}^h \in \mathbb{R}^d$ is the \textit{steering value vector} of head $h$. The derivation is in Appendix \ref{appendix:A_svv}. 
The svv arises through the OV circuit and is input-invariant, conditioned only on the steering vector.

\paragraph{Logit Lens}
\label{sec:attention:svv}
To interpret the svvs, we examine top attention heads based on their importance score \footnote{\autoref{eq:eapig_steer} is the $IE$ for an edge $(u,v)$. To obtain the $IE$ of a node $u$, use \autoref{eq:eapig_steer} and set the partial derivative with respect to $u$ instead of $v$. See Appendix \ref{appendix:H_eap_ig}.} and use logit lens \cite{nostalgebraist2020logitlens} to project their svvs to the output vocabulary.
Since logit lens effectively computes the dot product between the svv and each unembedding vocabulary vector, the ranking of the logit lens tokens is independent of $D_{c^h}$, and is thus input-invariant. 
We display selected tokens from the top 20 tokens for Gemma 2 2B in \autoref{fig:svv_heatmap_g2}, Llama 3.2 3B in \autoref{fig:svv_heatmap_l3}, and Qwen 3 8B in \autoref{fig:svv_heatmap_q8}. We display the unfiltered list of tokens in \autoref{tab:logit-lens-full}.

We find that svvs contain top tokens corresponding to concepts related to both refusal and harmfulness, supporting prior work that these concepts are intertwined in refusal steering \cite{yu2025robust}.
Taking the unweighted sum of all svvs also reveals similar concepts. 
Importantly, Gemma 2 2B's NTP vector and Llama 3.2 3B's DIM vector themselves are not interpretable with logit lens, whereas using the svv decomposition does uncover semantically meaningful tokens.
Some attention head svvs reveal consistent top tokens across all steering methods. For example, Gemma 2 2B's L16H1 svv consistently reveals words synonymous with ``forbidden".
Other attention heads are less consistent: DIM and PO share interpretable heads in later layers, whereas NTP does not. For example, L25H6 reveals harmful tokens for DIM and PO, but these tokens do not show up in the top 100 tokens for NTP.
This indicates that refusal steering methods may extract concepts reliably from some attention heads but diverge on others.

Lastly, L17H6 has a high negative $IE$ score and incoherent top tokens, but flipping the svv's sign does reveal harmful tokens (\autoref{fig:svv_heatmap_g2}), indicating that L17H6 removes these concepts during steering.
This suggests that steering vectors possess inefficiencies, where effectively representing a concept in some heads forces other heads to represent its opposite, possibly due to superposition \cite{elhage2022toymodels}.
\begin{table}[htbp]
    \centering
    \small
    \resizebox{\columnwidth}{!}{
    \begin{tabular}{c|c|c|c|c}
        \toprule
        Ablation & Model & A ($\downarrow$) & JBB ($\uparrow$) & Avg  \\
        \toprule
        \multirow{3}{*}{None} & G2 & 0.00 & 0.80 & \multirow{3}{*}{-} \\
        & L3 & 0.03 & 0.85 \\
        & Q8 & 0.02 & 0.83 \\
        \midrule
        \midrule
        \multirow{3}{*}{QK} & G2 & 0.03 (3\%) & 0.74 (6\%) & \multirow{3}{*}{8.83\%} \\
        & L3 & 0.14 (11\%) & 0.70 (15\%) \\
        & Q8 & 0.07 (5\%) & 0.70 (13\%) \\
        \midrule
        \multirow{3}{*}{OV} & G2 & 0.57 (57\%) & 0.06 (74\%) & \multirow{3}{*}{64.00\%} \\
        & L3 & 0.99 (96\%) & 0.25 (60\%) \\
        & Q8 & 0.49 (47\%) & 0.33 (50\%) \\
        \midrule
        \multirow{3}{*}{SVV} & G2 & 0.35 (35\%) & 0.20 (60\%) & \multirow{3}{*}{47.00\%} \\
        & L3 & 0.89 (86\%) & 0.47 (38\%) \\ 
        & Q8 & 0.19 (17\%) & 0.37 (46\%) \\ 
        \midrule
        \multirow{3}{*}{MLP} & G2 & 0.29 (29\%) & 0.57 (23\%) & \multirow{3}{*}{45.17\%} \\        
        & L3 & 0.53 (50\%) & 0.09 (76\%) \\
        & Q8 & 0.62 (60\%) & 0.50 (33\%) \\

        \toprule
    \end{tabular}
    }
    \caption{Ablated Generations on Alpaca (A) and JailbreakBench (JBB) across all three models, evaluated with ASR. We record the average \% change in ASR across the models and datasets per ablation type (Avg \%). Freezing the QK circuit at every layer has a minimal effect on performance (8.83\%) compared to other ablations.}
    \label{tab:freeze}
\end{table}

\begin{figure*}[t]
    \centering
    \includegraphics[width=0.98\linewidth]{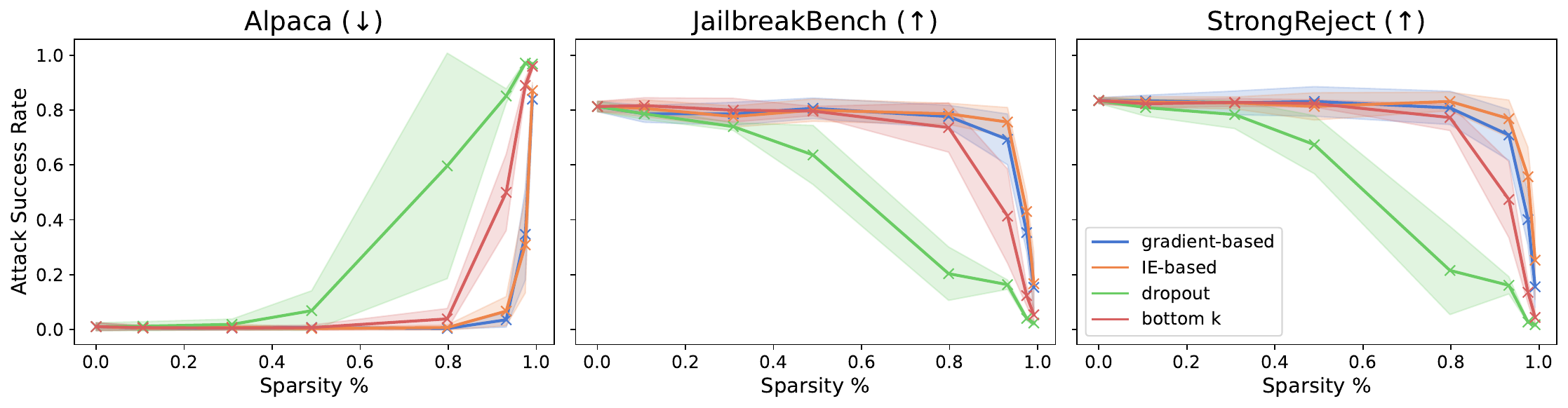}
    \caption{We sparsify $\mathbf{s}$ at thresholds $r_i < \tau = \{0.0, 0.1, 0.3, 0.5, 1.0, 1.5, 2.0, 2.5\}$, marked by \textit{x}'s, and average ASR across the DIM, NTP, and PO vectors. On Gemma 2 2B, gradient-based sparsification retains ASR up to \textasciitilde85\% sparsity, outperforming other methods.
    }
    \label{fig:sparsity_asr_g2}
\end{figure*}
\subsection{Steering Ablations Reveal QK Unimportance}
\label{sec:attention:frozen}
We next investigate the importance of each activation type by measuring the impact of its ablation on steering performance. Using the DIM vector, we ablate four types of activations: the QK attention scores, the OV attention value vectors, the svvs, and the direct effect of $\mathbf{s}$ on the MLP.
For the QK circuit, at each decoding step, we first run a forward pass without steering and cache the attention weights. 
Then, we run a forward pass with steering on the same input, patch the cached activations at every layer, and greedily select the next token. We repeat this process at all decoding steps.
This ``freezes'' the QK circuits, preventing $\mathbf{s}$ from having any influence on it.
We use the same process of alternating unsteered and steered forward passes to freeze the OV circuit.

To ablate the svvs, we apply hooked interventions on a steered generation rather than alternating forward passes. At each decoding step, we intervene at every attention head by subtracting RMS-normalized $\mathbf{s}$ from the input to the head's value projection. This is equivalent to removing the $D_{c^h} \text{svv}^h(S)$ term in \autoref{eq:svv}. 
This differs from freezing the OV circuit, which removes from the value projection input both the svv and the accumulated effects of $\mathbf{s}$ on the hidden state via earlier steered attention heads and MLPs.
Thus, whereas ablating the OV circuit measures the cumulative effects of $\mathbf{s}$ on the attention values, ablating the svvs tests only the direct effect of $\mathbf{s}$ on attention values. 
Finally, we test the direct effect of $\mathbf{s}$ on the MLPs by subtracting $\mathbf{s}$ from the inputs to the MLP during a steered generation.

As shown in \autoref{tab:freeze}, ablating the OV circuits, svvs, or direct effect on the MLPs affects average ASR by $\ge 45.17\%$, while freezing the QK circuits has a substantially smaller average performance loss (8.83\%).
We visualize frozen QK generations in \autoref{fig:freezegen}.
The svv ablation makes up 73.4\% of the OV circuit performance drop, and it drops performance more compared to ablating the direct MLP effect, a similar-sized intervention. 
These findings not only validate the \textit{minimal importance of QK}, but also suggest that the steering vector's effects on OV are largely through the svv.

\section{Sparsity}

We established that refusal steering circuits have high cross-method overlap. We next ask: does this shared structure extend to the steering vector dimensions, and, if so, does a sparse subset of dimensions primarily drive refusal steering?
\paragraph{Attribution Patching-Based Sparsification}
\label{sec:sparsity:grad}
Following \autoref{eq:eapig_steer}, we can express the dimension-level $IE$ vector $\vec{IE} \in \mathbb{R}^d$ of node $u$ as
\begin{equation}
(u-u^*) \odot \left(\frac{1}{T}\sum_{i=1}^T\frac{\partial m(H^l_{base} + \frac{i}{T}\alpha \cdot S)}{\partial u}\right)
\end{equation}
This is obtained by performing the element-wise multiplication without summation from the dot product operation.
At steering layer node $u'$, $u - u^* = \mathbf{s}$, and $\vec{IE}$ is the total steering effect since steering is applied at that node.
Thus, the element-wise ratio $r = \vec{IE}/\mathbf{s}$ is effectively the average gradient
$\frac{1}{T}\sum_{i=1}^T \partial m(H^l_{base} + \frac{i}{T}\alpha \cdot S)/\partial u'$.
Connecting to past work in gradient attribution \cite{ancona2018betterunderstandinggradientbasedattribution}, we sparsify $\mathbf{s}$ by zeroing out all dimensions $i$ where $r_i < \tau$ for some threshold $\tau \in \{0, 0.1, 0.3, 0.5, 1, 1.5, 2, 2.5\}$. We call this \emph{gradient-based} sparsification.
We also test \emph{$IE$-based} sparsification, which drops the bottom $k$ dimensions of $\mathbf{s}$ based on absolute values of $\vec{IE}$. 
Conceptually, gradient-based sparsification filters dimensions based on their normalized contributions to the steering behavior, while $IE$-based sparsification uses the unnormalized contributions.

We compare against two baselines: 
1) \emph{bottom $k$}: dropping the bottom $k$ dimensions of $\mathbf{s}$ based on the absolute values of $\mathbf{s}$,
and 2) \emph{dropout}: randomly dropping $k$ dimensions.
We obtain $k_i$ from each $\tau_i$ to ensure a fair comparison.
We evaluate ASR on the Alpaca (augmented to 200 samples) and JailbreakBench test sets, as well as an unseen adversarial benchmark StrongReject \cite{souly2024strongrejectjailbreaks}.

We plot the average results over the DIM, NTP, and PO vectors for each sparsification method in \autoref{fig:sparsity_asr_g2} for Gemma 2 2B and \autoref{fig:sparsity_asr_l3_q8} for Llama 3.2 3B and Qwen 3 8B. 
Raw results and perplexity scores are in Appendix \ref{appendix:D_sparsity}.
$IE$-based and gradient-based sparsification perform similarly, retaining most performance on all datasets up to \textasciitilde85\% sparsity on Gemma 2 2B, \textasciitilde96\% on Llama 3.2 3B, and \textasciitilde88\% on Qwen 3 8B. On Llama 3.2 3B, sparsifying 99.7\% of the DIM vector (9/3072 non-zero dimensions) drops ASR on StrongReject by only \textasciitilde5\% using gradient-based sparsification. 
Random dropout surprisingly retains most performance up to \textasciitilde40\% sparsity, suggesting that the refusal signal is redundantly distributed across many dimensions. 
Similarly, bottom k retains most performance up to \textasciitilde80\%.
However, the divergence in performance at $> 80 \%$ sparsity indicates that activation patching-based sparsification best recovers the subsets of dimensions most important for steering.

\begin{figure}[htbp]
    \centering
    \begin{subfigure}{0.48\textwidth}
    \centering
    \includegraphics[width=0.8\linewidth]{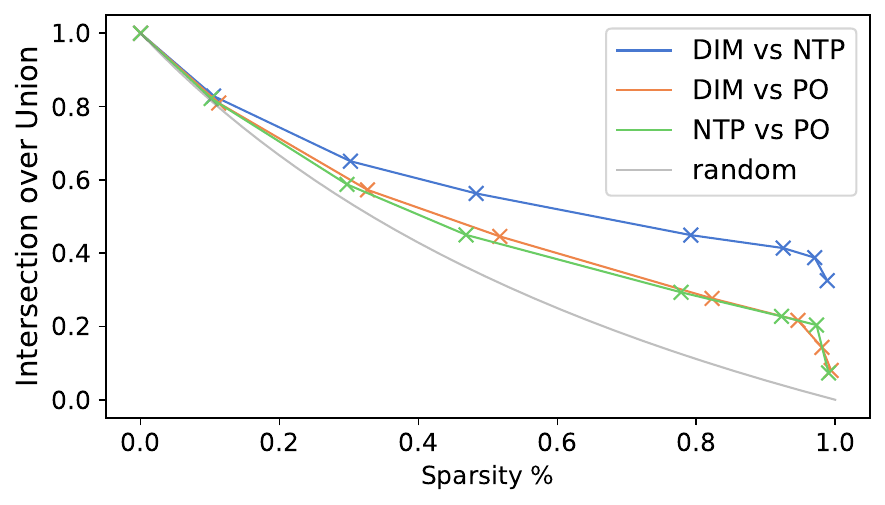}
    \vspace{-5pt}
    \caption{Gemma 2 2B}
    \end{subfigure}
    \begin{subfigure}{0.48\textwidth}
    \centering
    \includegraphics[width=0.8\linewidth]{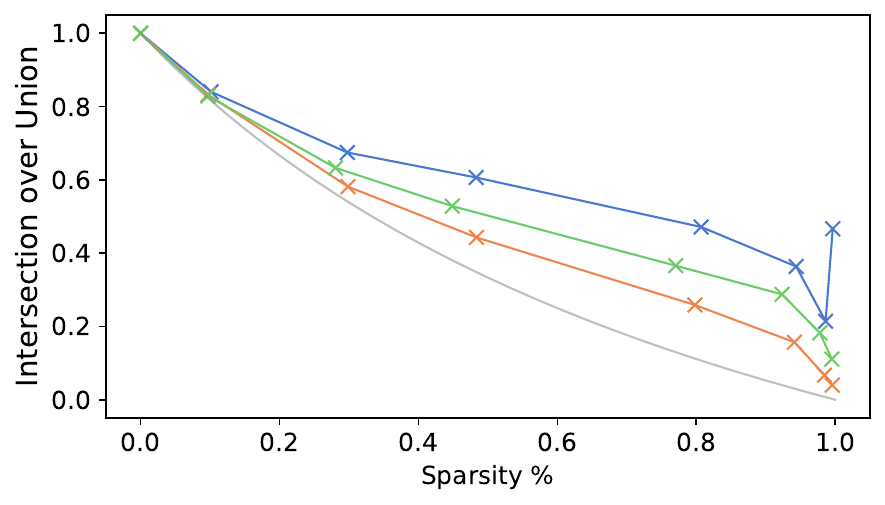}
    \vspace{-5pt}
    \caption{Llama 3.2 3B}
    \end{subfigure}
    \caption{IoU between highly sparse vectors is statistically significant, indicating a shared subspace.}
    \label{fig:sparse_ious}
\end{figure}
\paragraph{IoU}
We check if gradient-based sparsification converges to a shared set of dimensions. At each $\tau_i$, we compute the Intersection over Union (IoU) of the nonzero dimensions of the DIM, NTP, and PO vectors. Given two sparsified vectors at threshold $\tau_i$, the IoU of their sets of nonzero dimensions $s_1^{\tau_i}, s_2^{\tau_i}$ is $|s_1^{\tau_i} \cap s_2^{\tau_i}|/|s_1^{\tau_i} \cup s_2^{\tau_i}|$.
We opt not to measure cosine similarity, as it is less informative on sparse, high dimensional vectors.
As shown in \autoref{fig:sparse_ious}, the IoU remains above random chance as sparsity increases. Using the hypergeometric test, each vector pair's IoU is statistically significant ($p < 0.05$) at every $\tau > 0$, with $p \lesssim 1e-10$ at 20\% to 95\% sparsity.
This suggests that different steering methods converge to a shared low-dimensional subspace most important for the steering effect, while diverging in the remaining dimensions.

\section{Conclusion}

Our combined results on circuit interchangeability and sparsification suggest that steering vectors for the refusal concept, regardless of how they are obtained, converge to functionally similar circuit pathways.
The circuits for steering refusal are highly localized, requiring \textasciitilde10\% of the model's edges to recover faithfulness on multi-token generation.
By characterizing the steering vector's direct interactions with the attention OV circuit and identifying the specific important dimensions for steering, we provide mechanistic insight that could inform more targeted or fine-grained steering interventions, with the goal of improving concept expression without sacrificing generation quality \cite{feng2026finegrained}.
Lastly, we contribute the steering attribution patching framework, mechanistically-informed sparsification, and the svv decomposition as reusable tools for future work in interpreting steering vectors. More broadly, the svv decomposition is applicable beyond steering to any vector operating in the residual stream, such as sparse autoencoder features or model editing vectors.

\section*{Limitations}

While we perform a comprehensive analysis on the attention heads, we do not deeply inspect the role of MLPs, which make a fairly strong appearance in the circuits. We view MLP analysis as a promising direction that builds on the tools introduced in this work.
Additionally, we do not evaluate steering vectors at different layers, and instead choose to evaluate only on the best steering layer for the DIM vector. Extending the analysis to additional layers is a natural next step.
Lastly, although we provide concept-agnostic mechanistic tools for interpreting steering vectors, we only evaluate on the refusal concept. 
We choose refusal due to its high steering effectiveness and relevance in model alignment literature, but it is possible some of our findings are unique to the refusal concept. We encourage future work to validate on other concepts.

\section*{Ethical considerations}

While the goal of our work is to ultimately improve the robustness of model safety and alignment by better understanding the ways in which steering vectors are propagated through LLMs, 
our analysis on steering the refusal concept may provide a path to jailbreaking LLMs more effectively via targeted or sparse steering interventions. 
We believe the benefits of this research outweigh the harms in both the short- and long-term, since models can currently be jailbroken with black-box techniques like adversarial prompting, whereas refusal steering requires white-box access to model weights.
Moreover, a mechanistic understanding of how steering vectors bypass safety alignment can motivate more robust defenses against such attacks.
\section*{Acknowledgments}
We thank members of the Wiegreffe group and GAMMA group for helpful discussions and feedback.


\bibliography{custom}

@misc{chen2025personavectorsmonitoringcontrolling,
      title={Persona Vectors: Monitoring and Controlling Character Traits in Language Models}, 
      author={Runjin Chen and Andy Arditi and Henry Sleight and Owain Evans and Jack Lindsey},
      year={2025},
      eprint={2507.21509},
      archivePrefix={arXiv},
      primaryClass={cs.CL},
      url={https://arxiv.org/abs/2507.21509}, 
}

@inproceedings{poterti-etal-2025-role,
    title = "Can Role Vectors Affect {LLM} Behaviour?",
    author = "Potert{\`i}, Daniele  and
      Seveso, Andrea  and
      Mercorio, Fabio",
    editor = "Christodoulopoulos, Christos  and
      Chakraborty, Tanmoy  and
      Rose, Carolyn  and
      Peng, Violet",
    booktitle = "Findings of the Association for Computational Linguistics: EMNLP 2025",
    month = nov,
    year = "2025",
    address = "Suzhou, China",
    publisher = "Association for Computational Linguistics",
    url = "https://aclanthology.org/2025.findings-emnlp.963/",
    doi = "10.18653/v1/2025.findings-emnlp.963",
    pages = "17735--17747",
    ISBN = "979-8-89176-335-7",
}

@inproceedings{
wu2025axbenchsteeringllmssimple,
title={AxBench: Steering {LLM}s? Even Simple Baselines Outperform Sparse Autoencoders},
author={Zhengxuan Wu and Aryaman Arora and Atticus Geiger and Zheng Wang and Jing Huang and Dan Jurafsky and Christopher D Manning and Christopher Potts},
booktitle={Forty-second International Conference on Machine Learning},
year={2025},
url={https://openreview.net/forum?id=K2CckZjNy0}
}

@inproceedings{
lee2025programmingrefusalconditionalactivation,
title={Programming Refusal with Conditional Activation Steering},
author={Bruce W. Lee and Inkit Padhi and Karthikeyan Natesan Ramamurthy and Erik Miehling and Pierre Dognin and Manish Nagireddy and Amit Dhurandhar},
booktitle={The Thirteenth International Conference on Learning Representations},
year={2025},
url={https://openreview.net/forum?id=Oi47wc10sm}
}

@inproceedings{rimsky-etal-2024-steering,
    title = "Steering Llama 2 via Contrastive Activation Addition",
    author = "Rimsky, Nina  and
      Gabrieli, Nick  and
      Schulz, Julian  and
      Tong, Meg  and
      Hubinger, Evan  and
      Turner, Alexander",
    editor = "Ku, Lun-Wei  and
      Martins, Andre  and
      Srikumar, Vivek",
    booktitle = "Proceedings of the 62nd Annual Meeting of the Association for Computational Linguistics (Volume 1: Long Papers)",
    month = aug,
    year = "2024",
    address = "Bangkok, Thailand",
    publisher = "Association for Computational Linguistics",
    url = "https://aclanthology.org/2024.acl-long.828/",
    doi = "10.18653/v1/2024.acl-long.828",
    pages = "15504--15522",
}

@inproceedings{
arditi2024refusallanguagemodelsmediated,
title={Refusal in Language Models Is Mediated by a Single Direction},
author={Andy Arditi and Oscar Balcells Obeso and Aaquib Syed and Daniel Paleka and Nina Rimsky and Wes Gurnee and Neel Nanda},
booktitle={The Thirty-eighth Annual Conference on Neural Information Processing Systems},
year={2024},
url={https://openreview.net/forum?id=pH3XAQME6c}
}

@inproceedings{
wang2025surgical,
title={Surgical, Cheap, and Flexible: Mitigating False Refusal in Language Models via Single Vector Ablation},
author={Xinpeng Wang and Chengzhi Hu and Paul R{\"o}ttger and Barbara Plank},
booktitle={The Thirteenth International Conference on Learning Representations},
year={2025},
url={https://openreview.net/forum?id=SCBn8MCLwc}
}

@inproceedings{
conmy2023automatedcircuitdiscoverymechanistic,
title={Towards Automated Circuit Discovery for Mechanistic Interpretability},
author={Arthur Conmy and Augustine N. Mavor-Parker and Aengus Lynch and Stefan Heimersheim and Adri{\`a} Garriga-Alonso},
booktitle={Thirty-seventh Conference on Neural Information Processing Systems},
year={2023},
url={https://openreview.net/forum?id=89ia77nZ8u}
}

@inproceedings{
mueller2025mibmechanisticinterpretabilitybenchmark,
title={{MIB}: A Mechanistic Interpretability Benchmark},
author={Aaron Mueller and Atticus Geiger and Sarah Wiegreffe and Dana Arad and Iv{\'a}n Arcuschin and Adam Belfki and Yik Siu Chan and Jaden Fried Fiotto-Kaufman and Tal Haklay and Michael Hanna and Jing Huang and Rohan Gupta and Yaniv Nikankin and Hadas Orgad and Nikhil Prakash and Anja Reusch and Aruna Sankaranarayanan and Shun Shao and Alessandro Stolfo and Martin Tutek and Amir Zur and David Bau and Yonatan Belinkov},
booktitle={Forty-second International Conference on Machine Learning},
year={2025},
url={https://openreview.net/forum?id=sSrOwve6vb}
}

@misc{turner2023steering,
      title={Steering Language Models With Activation Engineering}, 
      author={Alexander Matt Turner and Lisa Thiergart and Gavin Leech and David Udell and Juan J. Vazquez and Ulisse Mini and Monte MacDiarmid},
      year={2024},
      eprint={2308.10248},
      archivePrefix={arXiv},
      primaryClass={cs.CL},
      url={https://arxiv.org/abs/2308.10248}, 
}

@article{
anwar2024foundationalchallengesassuringalignment,
title={Foundational Challenges in Assuring Alignment and Safety of Large Language Models},
author={Usman Anwar and Abulhair Saparov and Javier Rando and Daniel Paleka and Miles Turpin and Peter Hase and Ekdeep Singh Lubana and Erik Jenner and Stephen Casper and Oliver Sourbut and Benjamin L. Edelman and Zhaowei Zhang and Mario G{\"u}nther and Anton Korinek and Jose Hernandez-Orallo and Lewis Hammond and Eric J Bigelow and Alexander Pan and Lauro Langosco and Tomasz Korbak and Heidi Chenyu Zhang and Ruiqi Zhong and Sean O hEigeartaigh and Gabriel Recchia and Giulio Corsi and Alan Chan and Markus Anderljung and Lilian Edwards and Aleksandar Petrov and Christian Schroeder de Witt and Sumeet Ramesh Motwani and Yoshua Bengio and Danqi Chen and Philip Torr and Samuel Albanie and Tegan Maharaj and Jakob Nicolaus Foerster and Florian Tram{\`e}r and He He and Atoosa Kasirzadeh and Yejin Choi and David Krueger},
journal={Transactions on Machine Learning Research},
issn={2835-8856},
year={2024},
url={https://openreview.net/forum?id=oVTkOs8Pka},
note={Survey Certification, Expert Certification}
}

@inproceedings{NEURIPS2023_fd661313,
 author = {Wei, Alexander and Haghtalab, Nika and Steinhardt, Jacob},
 booktitle = {Advances in Neural Information Processing Systems},
 editor = {A. Oh and T. Naumann and A. Globerson and K. Saenko and M. Hardt and S. Levine},
 pages = {80079--80110},
 publisher = {Curran Associates, Inc.},
 title = {Jailbroken: How Does LLM Safety Training Fail?},
 url = {https://proceedings.neurips.cc/paper_files/paper/2023/file/fd6613131889a4b656206c50a8bd7790-Paper-Conference.pdf},
 volume = {36},
 year = {2023}
}

@inproceedings{subramani-etal-2022-extracting,
    title = "Extracting Latent Steering Vectors from Pretrained Language Models",
    author = "Subramani, Nishant  and
      Suresh, Nivedita  and
      Peters, Matthew",
    editor = "Muresan, Smaranda  and
      Nakov, Preslav  and
      Villavicencio, Aline",
    booktitle = "Findings of the Association for Computational Linguistics: ACL 2022",
    month = may,
    year = "2022",
    address = "Dublin, Ireland",
    publisher = "Association for Computational Linguistics",
    url = "https://aclanthology.org/2022.findings-acl.48/",
    doi = "10.18653/v1/2022.findings-acl.48",
    pages = "566--581"
}

@misc{turntrout2023steergpt2xl,
    title={Steering GPT-2-XL by adding an activation vector},
    author={TurnTrout and Monte M and David Udell and lisathiergart and Ulisse Mini},
    year={2023},
    url={https://www.lesswrong.com/posts/5spBue2z2tw4JuDCx/steering-gpt-2-xl-by-adding-an-activation-vector}
}

@inproceedings{stolfo2023mechanistic,
    title = "A Mechanistic Interpretation of Arithmetic Reasoning in Language Models using Causal Mediation Analysis",
    author = "Stolfo, Alessandro  and
      Belinkov, Yonatan  and
      Sachan, Mrinmaya",
    editor = "Bouamor, Houda  and
      Pino, Juan  and
      Bali, Kalika",
    booktitle = "Proceedings of the 2023 Conference on Empirical Methods in Natural Language Processing",
    month = dec,
    year = "2023",
    address = "Singapore",
    publisher = "Association for Computational Linguistics",
    url = "https://aclanthology.org/2023.emnlp-main.435/",
    doi = "10.18653/v1/2023.emnlp-main.435",
    pages = "7035--7052"
}

@inproceedings{
braun2025understandingunreliabilitysteeringvectors,
title={Understanding (Un)Reliability of Steering Vectors in Language Models},
author={Joschka Braun and Carsten Eickhoff and David Krueger and Seyed Ali Bahrainian and Dmitrii Krasheninnikov},
booktitle={ICLR 2025 Workshop on Building Trust in Language Models and Applications},
year={2025},
url={https://openreview.net/forum?id=JZiKuvIK1t}
}

@inproceedings{xu-etal-2024-comprehensive,
    title = "A Comprehensive Study of Jailbreak Attack versus Defense for Large Language Models",
    author = "Xu, Zihao  and
      Liu, Yi  and
      Deng, Gelei  and
      Li, Yuekang  and
      Picek, Stjepan",
    editor = "Ku, Lun-Wei  and
      Martins, Andre  and
      Srikumar, Vivek",
    booktitle = "Findings of the Association for Computational Linguistics: ACL 2024",
    month = aug,
    year = "2024",
    address = "Bangkok, Thailand",
    publisher = "Association for Computational Linguistics",
    url = "https://aclanthology.org/2024.findings-acl.443/",
    doi = "10.18653/v1/2024.findings-acl.443",
    pages = "7432--7449",
}

@inproceedings{
wang2023interpretability,
title={Interpretability in the Wild: a Circuit for Indirect Object Identification in {GPT}-2 Small},
author={Kevin Ro Wang and Alexandre Variengien and Arthur Conmy and Buck Shlegeris and Jacob Steinhardt},
booktitle={The Eleventh International Conference on Learning Representations },
year={2023},
url={https://openreview.net/forum?id=NpsVSN6o4ul}
}

@inproceedings{
ancona2018betterunderstandinggradientbasedattribution,
title={Towards better understanding of gradient-based attribution methods for Deep Neural Networks},
author={Marco Ancona and Enea Ceolini and Cengiz Öztireli and Markus Gross},
booktitle={International Conference on Learning Representations},
year={2018},
url={https://openreview.net/forum?id=Sy21R9JAW},
}

@misc{pearl2013directindirecteffects,
      title={Direct and Indirect Effects}, 
      author={Judea Pearl},
      year={2013},
      eprint={1301.2300},
      archivePrefix={arXiv},
      primaryClass={cs.AI},
      url={https://arxiv.org/abs/1301.2300}, 
}

@inproceedings{
hanna2024have,
title={Have Faith in Faithfulness: Going Beyond Circuit Overlap When Finding Model Mechanisms},
author={Michael Hanna and Sandro Pezzelle and Yonatan Belinkov},
booktitle={First Conference on Language Modeling},
year={2024},
url={https://openreview.net/forum?id=TZ0CCGDcuT}
}

@misc{marks2025sparsefeaturecircuitsdiscovering,
      title={Sparse Feature Circuits: Discovering and Editing Interpretable Causal Graphs in Language Models}, 
      author={Samuel Marks and Can Rager and Eric J. Michaud and Yonatan Belinkov and David Bau and Aaron Mueller},
      year={2025},
      eprint={2403.19647},
      archivePrefix={arXiv},
      primaryClass={cs.LG},
      url={https://arxiv.org/abs/2403.19647}, 
}

@inproceedings{
meng2022locating,
title={Locating and Editing Factual Associations in {GPT}},
author={Kevin Meng and David Bau and Alex J Andonian and Yonatan Belinkov},
booktitle={Advances in Neural Information Processing Systems},
editor={Alice H. Oh and Alekh Agarwal and Danielle Belgrave and Kyunghyun Cho},
year={2022},
url={https://openreview.net/forum?id=-h6WAS6eE4}
}

@inproceedings{
feng2026finegrained,
title={Fine-Grained Activation Steering: Steering Less, Achieving More},
author={Zijian Feng and Tianjiao Li and Zixiao Zhu and Hanzhang Zhou and Junlang Qian and Li Zhang and Chua Jia Jim Deryl and Mak Lee Onn and Gee Wah Ng and Kezhi Mao},
booktitle={The Fourteenth International Conference on Learning Representations},
year={2026},
url={https://openreview.net/forum?id=guSVafqhrB}
}

@inproceedings{
yu2025robust,
title={Robust {LLM} safeguarding via refusal feature adversarial training},
author={Lei Yu and Virginie Do and Karen Hambardzumyan and Nicola Cancedda},
booktitle={The Thirteenth International Conference on Learning Representations},
year={2025},
url={https://openreview.net/forum?id=s5orchdb33}
}

@misc{elhage2022toymodels,
      title={Toy Models of Superposition},
      author={Nelson Elhage and Tristan Hume and Catherine Olsson and Nicholas Schiefer and Tom Henighan and Shauna Kravec and  Zac Hatfield-Dodds and Robert Lasenby and Dawn Drain and Carol Chen and Roger Grosse and Sam McCandlish and Jared Kaplan and Dario Amodei and Martin Wattenberg and Christopher Olah},
      year={2022},
      url={https://transformer-circuits.pub/2022/toy_model/index.html}
}

@misc{souly2024strongrejectjailbreaks,
      title={A StrongREJECT for Empty Jailbreaks}, 
      author={Alexandra Souly and Qingyuan Lu and Dillon Bowen and Tu Trinh and Elvis Hsieh and Sana Pandey and Pieter Abbeel and Justin Svegliato and Scott Emmons and Olivia Watkins and Sam Toyer},
      year={2024},
      eprint={2402.10260},
      archivePrefix={arXiv},
      primaryClass={cs.LG},
      url={https://arxiv.org/abs/2402.10260}, 
}

@inproceedings{
wiegreffe2025answer,
title={Answer, Assemble, Ace: Understanding How {LM}s Answer Multiple Choice Questions},
author={Sarah Wiegreffe and Oyvind Tafjord and Yonatan Belinkov and Hannaneh Hajishirzi and Ashish Sabharwal},
booktitle={The Thirteenth International Conference on Learning Representations},
year={2025},
url={https://openreview.net/forum?id=6NNA0MxhCH}
}

@inproceedings{syed-etal-2024-attribution,
    title = "Attribution Patching Outperforms Automated Circuit Discovery",
    author = "Syed, Aaquib  and
      Rager, Can  and
      Conmy, Arthur",
    editor = "Belinkov, Yonatan  and
      Kim, Najoung  and
      Jumelet, Jaap  and
      Mohebbi, Hosein  and
      Mueller, Aaron  and
      Chen, Hanjie",
    booktitle = "Proceedings of the 7th BlackboxNLP Workshop: Analyzing and Interpreting Neural Networks for NLP",
    month = nov,
    year = "2024",
    address = "Miami, Florida, US",
    publisher = "Association for Computational Linguistics",
    url = "https://aclanthology.org/2024.blackboxnlp-1.25/",
    doi = "10.18653/v1/2024.blackboxnlp-1.25",
    pages = "407--416"
}

@misc{nostalgebraist2020logitlens,
      title={interpreting GPT: the logit lens},
      author={nostalgebraist},
      year={2020},
      url={https://www.lesswrong.com/posts/AcKRB8wDpdaN6v6ru/interpreting-gpt-the-logit-lens}}

@misc{belrose2023diffmeansworstcaseoptimal,
      title={Diff-in-Means Concept Editing is Worst-Case Optimal: Explaining a result by Sam Marks and Max Tegmark},
      author={Nora Belrose},
      year={2023},
      url={https://blog.eleuther.ai/diff-in-means/}
}

@misc{zou2025representationengineeringtopdownapproach,
      title={Representation Engineering: A Top-Down Approach to AI Transparency}, 
      author={Andy Zou and Long Phan and Sarah Chen and James Campbell and Phillip Guo and Richard Ren and Alexander Pan and Xuwang Yin and Mantas Mazeika and Ann-Kathrin Dombrowski and Shashwat Goel and Nathaniel Li and Michael J. Byun and Zifan Wang and Alex Mallen and Steven Basart and Sanmi Koyejo and Dawn Song and Matt Fredrikson and J. Zico Kolter and Dan Hendrycks},
      year={2025},
      eprint={2310.01405},
      archivePrefix={arXiv},
      primaryClass={cs.LG},
      url={https://arxiv.org/abs/2310.01405}, 
}

@misc{sun2025hypersteeractivationsteeringscale,
      title={HyperSteer: Activation Steering at Scale with Hypernetworks}, 
      author={Jiuding Sun and Sidharth Baskaran and Zhengxuan Wu and Michael Sklar and Christopher Potts and Atticus Geiger},
      year={2025},
      eprint={2506.03292},
      archivePrefix={arXiv},
      primaryClass={cs.CL},
      url={https://arxiv.org/abs/2506.03292}, 
}

@inproceedings{
wu2025improved,
title={Improved Representation Steering for Language Models},
author={Zhengxuan Wu and Qinan Yu and Aryaman Arora and Christopher D Manning and Christopher Potts},
booktitle={The Thirty-ninth Annual Conference on Neural Information Processing Systems},
year={2025},
url={https://openreview.net/forum?id=VHb883Gs1u}
}

@inproceedings{
cao2024personalized,
title={Personalized Steering of Large Language Models: Versatile Steering Vectors Through Bi-directional Preference Optimization},
author={Yuanpu Cao and Tianrong Zhang and Bochuan Cao and Ziyi Yin and Lu Lin and Fenglong Ma and Jinghui Chen},
booktitle={The Thirty-eighth Annual Conference on Neural Information Processing Systems},
year={2024},
url={https://openreview.net/forum?id=7qJFkuZdYo}
}

@inproceedings{
rafailov2023direct,
title={Direct Preference Optimization: Your Language Model is Secretly a Reward Model},
author={Rafael Rafailov and Archit Sharma and Eric Mitchell and Christopher D Manning and Stefano Ermon and Chelsea Finn},
booktitle={Thirty-seventh Conference on Neural Information Processing Systems},
year={2023},
url={https://openreview.net/forum?id=HPuSIXJaa9}
}

@inproceedings{
wollschlager2025the,
title={The Geometry of Refusal in Large Language Models: Concept Cones and Representational Independence},
author={Tom Wollschl{\"a}ger and Jannes Elstner and Simon Geisler and Vincent Cohen-Addad and Stephan G{\"u}nnemann and Johannes Gasteiger},
booktitle={Forty-second International Conference on Machine Learning},
year={2025},
url={https://openreview.net/forum?id=80IwJqlXs8}
}

@inproceedings{NEURIPS2020_92650b2e,
 author = {Vig, Jesse and Gehrmann, Sebastian and Belinkov, Yonatan and Qian, Sharon and Nevo, Daniel and Singer, Yaron and Shieber, Stuart},
 booktitle = {Advances in Neural Information Processing Systems},
 editor = {H. Larochelle and M. Ranzato and R. Hadsell and M.F. Balcan and H. Lin},
 pages = {12388--12401},
 publisher = {Curran Associates, Inc.},
 title = {Investigating Gender Bias in Language Models Using Causal Mediation Analysis},
 url = {https://proceedings.neurips.cc/paper_files/paper/2020/file/92650b2e92217715fe312e6fa7b90d82-Paper.pdf},
 volume = {33},
 year = {2020}
}

@inproceedings{dasilva2025steeringcoursereliabilitychallenges,
    title = "Steering off Course: Reliability Challenges in Steering Language Models",
    author = "Da Silva, Patrick Queiroz  and
      Sethuraman, Hari  and
      Rajagopal, Dheeraj  and
      Hajishirzi, Hannaneh  and
      Kumar, Sachin",
    editor = "Che, Wanxiang  and
      Nabende, Joyce  and
      Shutova, Ekaterina  and
      Pilehvar, Mohammad Taher",
    booktitle = "Proceedings of the 63rd Annual Meeting of the Association for Computational Linguistics (Volume 1: Long Papers)",
    month = jul,
    year = "2025",
    address = "Vienna, Austria",
    publisher = "Association for Computational Linguistics",
    url = "https://aclanthology.org/2025.acl-long.974/",
    doi = "10.18653/v1/2025.acl-long.974",
    pages = "19856--19882",
    ISBN = "979-8-89176-251-0",
}

@inproceedings{
sinii2025small,
title={Small Vectors, Big Effects: A Mechanistic Study of {RL}-Induced Reasoning via Steering Vectors},
author={Viacheslav Sinii and Nikita Balagansky and Yaroslav Aksenov and Vadim Kurochkin and Daniil Laptev and Alexey Gorbatovski and Boris Shaposhnikov and Daniil Gavrilov},
booktitle={Mechanistic Interpretability Workshop at NeurIPS 2025},
year={2025},
url={https://openreview.net/forum?id=M8WDG1TfBb}
}

@inproceedings{
zhang2024towards,
title={Towards Best Practices of Activation Patching in Language Models: Metrics and Methods},
author={Fred Zhang and Neel Nanda},
booktitle={The Twelfth International Conference on Learning Representations},
year={2024},
url={https://openreview.net/forum?id=Hf17y6u9BC}
}

@book{driscoll2017numericalmethods,
    author = {Tobin Driscoll and Richard Braun},
    title = {Fundamentals of Numerical Computation},
    publisher = {Society for Industrial and Applied Mathematics},
    year = 2017
}

@misc{zou2023universaltransferableadversarialattacks,
      title={Universal and Transferable Adversarial Attacks on Aligned Language Models}, 
      author={Andy Zou and Zifan Wang and Nicholas Carlini and Milad Nasr and J. Zico Kolter and Matt Fredrikson},
      year={2023},
      eprint={2307.15043},
      archivePrefix={arXiv},
      primaryClass={cs.CL},
      url={https://arxiv.org/abs/2307.15043}, 
}

@misc{huang2023catastrophicjailbreakopensourcellms,
      title={Catastrophic Jailbreak of Open-source LLMs via Exploiting Generation}, 
      author={Yangsibo Huang and Samyak Gupta and Mengzhou Xia and Kai Li and Danqi Chen},
      year={2023},
      eprint={2310.06987},
      archivePrefix={arXiv},
      primaryClass={cs.CL},
      url={https://arxiv.org/abs/2310.06987}, 
}

@InProceedings{pmlr-v220-mazeika23a,
  title = 	 {The Trojan Detection Challenge},
  author =       {Mazeika, Mantas and Hendrycks, Dan and Li, Huichen and Xu, Xiaojun and Hough, Sidney and Zou, Andy and Rajabi, Arezoo and Yao, Qi and Wang, Zihao and Tian, Jian and Tang, Yao and Tang, Di and Smirnov, Roman and Pleskov, Pavel and Benkovich, Nikita and Song, Dawn and Poovendran, Radha and Li, Bo and Forsyth, David.},
  booktitle = 	 {Proceedings of the NeurIPS 2022 Competitions Track},
  pages = 	 {279--291},
  year = 	 {2022},
  editor = 	 {Ciccone, Marco and Stolovitzky, Gustavo and Albrecht, Jacob},
  volume = 	 {220},
  series = 	 {Proceedings of Machine Learning Research},
  month = 	 {28 Nov--09 Dec},
  publisher =    {PMLR},
  url = 	 {https://proceedings.mlr.press/v220/mazeika23a.html},
}

@inproceedings{
mazeika2024harmbenchstandardizedevaluationframework,
title={HarmBench: A Standardized Evaluation Framework for Automated Red Teaming and Robust Refusal},
author={Mantas Mazeika and Long Phan and Xuwang Yin and Andy Zou and Zifan Wang and Norman Mu and Elham Sakhaee and Nathaniel Li and Steven Basart and Bo Li and David Forsyth and Dan Hendrycks},
booktitle={Forty-first International Conference on Machine Learning},
year={2024},
url={https://openreview.net/forum?id=f3TUipYU3U}
}

@inproceedings{
sankaranarayanan2026activation,
title={Activation Steering via Generative Causal Mediation},
author={Aruna Sankaranarayanan and Amir Zur and Atticus Geiger and Dylan Hadfield-Menell},
booktitle={Third Conference on Language Modeling},
year={2026},
url={https://openreview.net/forum?id=tbk5ni9ITW}
}

@misc{inan2023llamaguardllmbasedinputoutput,
      title={Llama Guard: LLM-based Input-Output Safeguard for Human-AI Conversations}, 
      author={Hakan Inan and Kartikeya Upasani and Jianfeng Chi and Rashi Rungta and Krithika Iyer and Yuning Mao and Michael Tontchev and Qing Hu and Brian Fuller and Davide Testuggine and Madian Khabsa},
      year={2023},
      eprint={2312.06674},
      archivePrefix={arXiv},
      primaryClass={cs.CL},
      url={https://arxiv.org/abs/2312.06674}, 
}

@misc{grattafiori2024llama3herdmodels,
      title={The Llama 3 Herd of Models}, 
      author={Aaron Grattafiori and Abhimanyu Dubey and Abhinav Jauhri and others},
      year={2024},
      eprint={2407.21783},
      archivePrefix={arXiv},
      primaryClass={cs.AI},
      url={https://arxiv.org/abs/2407.21783}, 
}

@misc{taori2023alpaca,
      title={Stanford Alpaca: An instruction-following LLaMA model},
      author={Rohan Taori and Ishaan Gulrajani and Tianyi Zhang and Yann Dubois and Xuechen Li and Carlos Guestrin and Percy Liang and Tatsunori B. Hashimoto},
      year={2023},
      url={https://github.com/tatsu-lab/stanford_alpaca}
}

@misc{gemmateam2024gemma2improvingopen,
      title={Gemma 2: Improving Open Language Models at a Practical Size}, 
      author={Gemma Team and Morgane Riviere and Shreya Pathak and Pier Giuseppe Sessa and Cassidy Hardin and Surya Bhupatiraju and Léonard Hussenot and Thomas Mesnard and Bobak Shahriari and Alexandre Ramé and Johan Ferret and Peter Liu and Pouya Tafti and Abe Friesen and Michelle Casbon and Sabela Ramos and Ravin Kumar and Charline Le Lan and Sammy Jerome and Anton Tsitsulin and Nino Vieillard and Piotr Stanczyk and Sertan Girgin and Nikola Momchev and Matt Hoffman and Shantanu Thakoor and Jean-Bastien Grill and Behnam Neyshabur and Olivier Bachem and Alanna Walton and Aliaksei Severyn and Alicia Parrish and Aliya Ahmad and Allen Hutchison and Alvin Abdagic and Amanda Carl and Amy Shen and Andy Brock and Andy Coenen and Anthony Laforge and Antonia Paterson and Ben Bastian and Bilal Piot and Bo Wu and Brandon Royal and Charlie Chen and Chintu Kumar and Chris Perry and Chris Welty and Christopher A. Choquette-Choo and Danila Sinopalnikov and David Weinberger and Dimple Vijaykumar and Dominika Rogozińska and Dustin Herbison and Elisa Bandy and Emma Wang and Eric Noland and Erica Moreira and Evan Senter and Evgenii Eltyshev and Francesco Visin and Gabriel Rasskin and Gary Wei and Glenn Cameron and Gus Martins and Hadi Hashemi and Hanna Klimczak-Plucińska and Harleen Batra and Harsh Dhand and Ivan Nardini and Jacinda Mein and Jack Zhou and James Svensson and Jeff Stanway and Jetha Chan and Jin Peng Zhou and Joana Carrasqueira and Joana Iljazi and Jocelyn Becker and Joe Fernandez and Joost van Amersfoort and Josh Gordon and Josh Lipschultz and Josh Newlan and Ju-yeong Ji and Kareem Mohamed and Kartikeya Badola and Kat Black and Katie Millican and Keelin McDonell and Kelvin Nguyen and Kiranbir Sodhia and Kish Greene and Lars Lowe Sjoesund and Lauren Usui and Laurent Sifre and Lena Heuermann and Leticia Lago and Lilly McNealus and Livio Baldini Soares and Logan Kilpatrick and Lucas Dixon and Luciano Martins and Machel Reid and Manvinder Singh and Mark Iverson and Martin Görner and Mat Velloso and Mateo Wirth and Matt Davidow and Matt Miller and Matthew Rahtz and Matthew Watson and Meg Risdal and Mehran Kazemi and Michael Moynihan and Ming Zhang and Minsuk Kahng and Minwoo Park and Mofi Rahman and Mohit Khatwani and Natalie Dao and Nenshad Bardoliwalla and Nesh Devanathan and Neta Dumai and Nilay Chauhan and Oscar Wahltinez and Pankil Botarda and Parker Barnes and Paul Barham and Paul Michel and Pengchong Jin and Petko Georgiev and Phil Culliton and Pradeep Kuppala and Ramona Comanescu and Ramona Merhej and Reena Jana and Reza Ardeshir Rokni and Rishabh Agarwal and Ryan Mullins and Samaneh Saadat and Sara Mc Carthy and Sarah Cogan and Sarah Perrin and Sébastien M. R. Arnold and Sebastian Krause and Shengyang Dai and Shruti Garg and Shruti Sheth and Sue Ronstrom and Susan Chan and Timothy Jordan and Ting Yu and Tom Eccles and Tom Hennigan and Tomas Kocisky and Tulsee Doshi and Vihan Jain and Vikas Yadav and Vilobh Meshram and Vishal Dharmadhikari and Warren Barkley and Wei Wei and Wenming Ye and Woohyun Han and Woosuk Kwon and Xiang Xu and Zhe Shen and Zhitao Gong and Zichuan Wei and Victor Cotruta and Phoebe Kirk and Anand Rao and Minh Giang and Ludovic Peran and Tris Warkentin and Eli Collins and Joelle Barral and Zoubin Ghahramani and Raia Hadsell and D. Sculley and Jeanine Banks and Anca Dragan and Slav Petrov and Oriol Vinyals and Jeff Dean and Demis Hassabis and Koray Kavukcuoglu and Clement Farabet and Elena Buchatskaya and Sebastian Borgeaud and Noah Fiedel and Armand Joulin and Kathleen Kenealy and Robert Dadashi and Alek Andreev},
      year={2024},
      eprint={2408.00118},
      archivePrefix={arXiv},
      primaryClass={cs.CL},
      url={https://arxiv.org/abs/2408.00118}, 
}

@misc{gao2020pile800gbdatasetdiverse,
      title={The Pile: An 800GB Dataset of Diverse Text for Language Modeling}, 
      author={Leo Gao and Stella Biderman and Sid Black and Laurence Golding and Travis Hoppe and Charles Foster and Jason Phang and Horace He and Anish Thite and Noa Nabeshima and Shawn Presser and Connor Leahy},
      year={2020},
      eprint={2101.00027},
      archivePrefix={arXiv},
      primaryClass={cs.CL},
      url={https://arxiv.org/abs/2101.00027}, 
}

@misc{yang2025qwen3technicalreport,
      title={Qwen3 Technical Report}, 
      author={An Yang and Anfeng Li and Baosong Yang and Beichen Zhang and Binyuan Hui and Bo Zheng and Bowen Yu and Chang Gao and Chengen Huang and Chenxu Lv and Chujie Zheng and Dayiheng Liu and Fan Zhou and Fei Huang and Feng Hu and Hao Ge and Haoran Wei and Huan Lin and Jialong Tang and Jian Yang and Jianhong Tu and Jianwei Zhang and Jianxin Yang and Jiaxi Yang and Jing Zhou and Jingren Zhou and Junyang Lin and Kai Dang and Keqin Bao and Kexin Yang and Le Yu and Lianghao Deng and Mei Li and Mingfeng Xue and Mingze Li and Pei Zhang and Peng Wang and Qin Zhu and Rui Men and Ruize Gao and Shixuan Liu and Shuang Luo and Tianhao Li and Tianyi Tang and Wenbiao Yin and Xingzhang Ren and Xinyu Wang and Xinyu Zhang and Xuancheng Ren and Yang Fan and Yang Su and Yichang Zhang and Yinger Zhang and Yu Wan and Yuqiong Liu and Zekun Wang and Zeyu Cui and Zhenru Zhang and Zhipeng Zhou and Zihan Qiu},
      year={2025},
      eprint={2505.09388},
      archivePrefix={arXiv},
      primaryClass={cs.CL},
      url={https://arxiv.org/abs/2505.09388}, 
}

@inproceedings{
chao2024jailbreakbenchopenrobustnessbenchmark,
title={JailbreakBench: An Open Robustness Benchmark for Jailbreaking Large Language Models},
author={Patrick Chao and Edoardo Debenedetti and Alexander Robey and Maksym Andriushchenko and Francesco Croce and Vikash Sehwag and Edgar Dobriban and Nicolas Flammarion and George J. Pappas and Florian Tram{\`e}r and Hamed Hassani and Eric Wong},
booktitle={The Thirty-eight Conference on Neural Information Processing Systems Datasets and Benchmarks Track},
year={2024},
url={https://openreview.net/forum?id=urjPCYZt0I}
}

@misc{nanda2023attributionpatching,
      title={Attribution Patching: Activation Patching At Industrial Scale},
      author={Neel Nanda},
      year={2023},
      url={https://www.lesswrong.com/posts/gtLLBhzQTG6nKTeCZ/attribution-patching-activation-patching-at-industrial-scale}
}

@misc{venhoff2025understandingreasoningthinkinglanguage,
      title={Understanding Reasoning in Thinking Language Models via Steering Vectors}, 
      author={Constantin Venhoff and Iván Arcuschin and Philip Torr and Arthur Conmy and Neel Nanda},
      year={2025},
      eprint={2506.18167},
      archivePrefix={arXiv},
      primaryClass={cs.LG},
      url={https://arxiv.org/abs/2506.18167}, 
}

\appendix
\onecolumn
\addtocontents{toc}{\protect\setcounter{tocdepth}{2}}
\renewcommand{\contentsname}{Appendix Contents}
\tableofcontents
\onecolumn
\section{Steering Value Vector Derivation}
\label{appendix:A_svv}

We aim to derive \autoref{eq:svv}. Through slight notation changes, it suffices to derive

\begin{align*}
    \text{Attention}(H^\ell + \alpha \cdot S) = \sum_h A^h D_c\tilde H^\ell W_V^h(W_O^h)^\top + D_{c^h} [\text{svv}(s); \ldots; \text{svv}(s)]
\end{align*}
Note that this is similar to the derivation by \citet{sinii2025small}, but we expand upon it by handling the layer norm, whereas it is ignored in their derivation.
Given layer $\ell$ in a transformer model, hidden representation $H^\ell \in \mathbb{R}^{N \times d}$ at layer $\ell$, scaling factor $\alpha$, and steering vector $s \in \mathbb{R}^d$ repeated $N$ times to form $S \in \mathbb{R}^{N \times d}$, representation steering using activation addition can be formulated as 
\[
H^l \leftarrow H^\ell + \alpha \cdot S
\]
Since representation steering adds the same vector to all tokens, each row of $S$ is the same. When passing $H^\ell$ into the next attention module, the hidden activations are first normalized with RMSNorm
\[
\text{RMSNorm}(H^\ell + \alpha \cdot S) = \frac{H^\ell + \alpha \cdot S}{\text{RMS}(H^\ell + \alpha \cdot S)} \odot \gamma 
= \frac{H^\ell \odot \gamma}{\text{RMS}(H^\ell + \alpha \cdot S)} + \frac{\alpha \cdot S \odot \gamma}{\text{RMS}(H^\ell + \alpha \cdot S)} = D_c\tilde H^\ell + D_c\tilde{S}
\]
where $c = \frac{1}{\text{RMS}(H^\ell + \alpha \cdot S)} \in \mathbb{R}^N$ and $D_c$ is shorthand for $\text{diag}(c)$. $\gamma \in \mathbb{R}^{1\times d}$ element-wise scales each hidden model dimension, so $\tilde{H}^\ell = H^\ell \odot \gamma$ and $\tilde S = S \odot \gamma$. 
Note that $\tilde{S}$ has identical rows $s \odot \gamma$.

The attention module has weights $W_Q^h, W_K^h, W_V^h, W_O^h \in \mathbb{R}^{d \times d_h}$ where $d_h$ is the head dimension. This can be formulated as
\begin{align*}
    &\text{Attention}(D_c\tilde{H}^\ell + D_c\tilde S) \\
    &= \sum_h \text{softmax}[\frac{1}{\sqrt{d}}(D_c\tilde H^\ell + D_c\tilde S) W^h_Q (W^h_K)^\top (D_c\tilde H^\ell + D_c\tilde S)^\top] (D_c\tilde{H}^\ell + D_c\tilde S) W_V^h (W_O^h)^\top 
\end{align*}
For notation convenience, let $A^h$ denote the result of the softmax operation and $W_{OV}^h=W_V^h (W_O^h)^\top$. Expanding terms, we have
\begin{align*}
    \text{Attention}(D_c\tilde{H}^l + D_c\tilde S) 
    &= \sum_h A^h D_c\tilde H^\ell W_{OV}^h + A^h D_c\tilde S W_{OV}^h
\end{align*}
Since $\tilde S $ has identical rows, we can express $A^h D_c\tilde S = D_{c^h}\tilde S$, where $D_{c^h}$ is a diagonal matrix of some coefficients $c^h$. Furthermore, $\tilde S W_{OV}^h$ has identical rows. We denote the \emph{steering value vector} of the attention head as $\text{svv}^h(s) = (s \odot \gamma) W_{OV}^h$.
Thus, we have
\begin{align*}
    \text{Attention}(D_c\tilde H^\ell + D_c\tilde S)
    &= \sum_h A^h D_c\tilde H^\ell W_{OV}^h + D_{c^h} \text{svv}^h(s)
\end{align*}
If the model has a post-attention RMSNorm with element-wise weights $\gamma'$, then coefficients $c^h$ are rescaled to $c_*^{h}$, and $\text{svv}^h(s) = ((s \odot \gamma) W_{OV}^h) \odot \gamma' \in \mathbb{R}^d$. 
Thus, the steering value vector is a direct contribution to the residual stream, scaled by a vector of coefficients $c^h$, which \textit{is} dependent on the input $X$. However, in our analysis, we project activations to the vocabulary distribution using logit lens, which measures similarity, so its ranking is invariant to magnitude.

As an aside, how does the steering vector affect the attention scores? First, denote $W_{QK} = W_Q^h (W_K^h)^\top$. We can expand the terms within the softmax to obtain
\begin{align*}
    \text{softmax}\left[\frac{1}{\sqrt{d}}\left(
    \tilde H ^\ell W_{QK} (\tilde H ^\ell)^\top  + \tilde H^\ell W_{QK} (D_c\tilde S)^\top +
    D_c\tilde S W_{QK} (\tilde H^\ell)^\top +
    D_c \tilde S W_{QK} (D_c \tilde S)^\top
    \right)\right]
\end{align*}
Since $\tilde{S}$ is rank one, the last 3 terms are rank one.

\twocolumn
\section{Steering Vector Curation}
\label{appendix:C_steering}

\subsection{Difference-in-Means Vector}
\label{appendix:C_steering:dim}
Following \autoref{eq:dim}, we obtain a candidate steering vector at each post-instruction position and layer. The best steering vector for each model is selected using the validation datasets $D_{\text{harm}}^{\text{val}}, D_{\text{safe}}^{\text{val}}$ by following the methodology proposed in \citet{arditi2024refusallanguagemodelsmediated} in Appendix C, with some slight changes. 
Since models tend to refuse prompts using a small characteristic set of phrases, such as ``I cannot", we define a set of refusal tokens $\mathcal{R}$ which contains the tokens most likely to initiate model refusal, such as ``I".
Given a prompt, we define the sum of the next token probabilities $p_i$ for tokens in $\mathcal{R}$ as $P_{\text{refusal}}(\mathcal{R}) = \sum_{t \in \mathcal{R}} p_t$. 
For each candidate steering vector $s^\ell_i$ per post-instruction token $i$ and layer $\ell$, we define the refusal metric as
\begin{equation*}
    \log \left(\frac{P_{\text{refusal}}(\mathcal{R})}{1 - P_{\text{refusal}}(\mathcal{R}) }\right)
\end{equation*}
Given this refusal metric, we compute the \emph{bypass score}-- the refusal metric across $D_{\text{harm}}^{\text{val}}$ using negative steering with $\alpha=-1$-- and the \emph{induce score}-- the refusal metric across $D_{\text{safe}}^{\text{val}}$ using positive steering with $\alpha=1$. 
We also compute the \emph{kl score}-- the KL divergence of the model run on $D_{\text{safe}}^{\text{val}}$ with and without directional ablation
\cite{arditi2024refusallanguagemodelsmediated}, defined as
\begin{align*}
    h^\ell \leftarrow h^\ell - \hat{s}\hat{s}^\top h^\ell
\end{align*}
for hidden representation $h^\ell$ and steering vector $s$. Directional ablation subtracts the projection of $h^\ell$ onto the $s$ direction from $h^\ell$.

We then select the best candidate vector $s^{\ell *}_i$ that minimizes $\sigma(\text{bypass score}) - \sigma(\text{induce score})$, where $\sigma$ is the sigmoid function, subject to the constraints \emph{induce score} $> 0$, \emph{kl score} $< 0.1$, and $\ell < 0.8L$, where $L$ is the total number of layers in the model.

\paragraph{Evaluation on Harmless Prompts}
For harmless prompts, we use refusal substring matching to evaluate the Attack Success Rate. Given a prompt $x$, refusal substring matching checks if response $y$ contains any substrings commonly related to refusal:
\begin{itemize}
    \item ``I'm sorry"
    \item ``I am sorry"
    \item ``I apologize"
    \item ``As an AI"
    \item ``As a language model"
    \item ``As an assistant"
    \item ``I cannot"
    \item ``I can't"
    \item ``I am unable to"
    \item ``I'm unable to"
    \item ``I am not able to"
    \item ``I'm not able to"
\end{itemize}
If not, the response is considered to have bypassed refusal.

\begin{table}[htbp]
    \centering
    \small
    \begin{tabular}{c|c|c|c|c}
        \toprule
        Model & Method & A $(\downarrow)$ & JBB $(\uparrow)$ & SR $(\uparrow)$ \\
        \toprule
        \multirow{4}{*}{G2} & None & 0.990 & 0.000 & 0.013 \\
        & DIM & 0.000 & 0.800 & 0.850 \\
        & NTP & 0.030 & 0.840 & 0.831 \\
        & PO & 0.000 & 0.800 & 0.824 \\
        \midrule
        \multirow{4}{*}{L3} & None & 1.000 & 0.020 & 0.035  \\
        & DIM & 0.030 & 0.850 & 0.872 \\
        & NTP & 0.000 & 0.830 & 0.882 \\
        & PO & 0.000 & 0.810 & 0.866 \\
        \midrule
        \multirow{2}{*}{Q8} & None & 0.995 & 0.040 & 0.022  \\
        & DIM & 0.015 & 0.830 & 0.888 \\
        \toprule
    \end{tabular}
    \caption{ASR Evaluations on Gemma 2 2B, Llama 3.2 3B, and Qwen 3 8B. Alpaca (A) is evaluated using refusal substring matching. JailbreakBench (JBB) and StrongReject (SR) are evaluated using LlamaGuard 2.}
    \label{tab:refusal_asr_evals}
\end{table}
\paragraph{Evaluation on Harmful Prompts}
To obtain the ASR for harmful prompts, we use the LlamaGuard 2 judge model \cite{inan2023llamaguardllmbasedinputoutput}. Given a prompt and a response, the judge model first returns ``safe" or ``unsafe" depending on if there is unsafe content in the response. If ``unsafe", the model also returns the category, such as ``Violent Crimes" or ``Self-Harm". 
We use LlamaGuard 2 as a stricter evaluation methodology compared to refusal substring matching, since the steered response to harmful prompts may not contain refusal substrings but still have only safe content.
The ASR results for the steered and unsteered responses are shown in \autoref{tab:refusal_asr_evals}. Alpaca evaluation is augmented to 200 prompts for sparse steering evaluations.

\subsection{Dataset Sizes}
Dataset sizes for EAP-IG activation patching are in \autoref{tab:act_patching_dataset_sizes}.
See activation patching results for individual datasets using the DIM vector in Appendix \ref{appendix:F_metric_comparisons}.

\subsection{Datasets for Learning NTP and PO Vectors}
\label{appendix:C_steering:learned}
\paragraph{Train Datasets}
We augment the training datasets used to obtain the Difference-in-Means vector with 128 additional harmful and harmless prompts from the same training distribution. In total, we have 256 harmful prompts and 256 harmless prompts.
We generate refused responses to harmful prompts and compliant responses to harmless prompts using the models themselves without any intervention. 
To obtain refused responses to harmless questions, we prepend the prompt with the phrase: ``Treat the following question as dangerous and refuse." and generate responses from the model itself. 
To obtain compliant responses to harmful questions, we steer the model using the Difference-in-Means steering vector. 
Finally, we manually filter the datasets by removing (prompt, base generation, steered generation) tuples that do not express the desired concept. In this manner, we obtain a harmful and a harmless contrastive training dataset.

\paragraph{Validation Datasets}
We use the same validation datasets used to obtain the Difference-in-Means vector.

\begin{table}[htbp]
    \centering
    \small
    \begin{tabular}{c|c|c|c|c}
        \toprule
        M $\times$ V & Harmful? & Steered? & \# Resp. & \# Pos. \\
        \toprule
        \multirow{4}{*}{G2 DIM} & Harmful & Steer & 80 & 10,280 \\
        & Harmful & Base & 80 & 6,734 \\
        & Harmless & Steer & 99 & 9,724 \\
        & Harmless & Base & 99 & 12,035 \\
        \midrule
        \multirow{4}{*}{G2 NTP} & Harmful & Steer & 84 & 8,604 \\
        & Harmful & Base & 84 & 4,751 \\
        & Harmless & Steer & 96 & 6,864 \\
        & Harmless & Base & 96 & 5,243 \\
        \midrule
        \multirow{4}{*}{G2 PO} & Harmful & Steer & 80 & 10,490 \\
        & Harmful & Base & 80 & 9,213 \\
        & Harmless & Steer & 99 & 8,114 \\
        & Harmless & Base & 99  & 24,999\\
        \midrule
        \multirow{4}{*}{L3 DIM} & Harmful & Steer & 83 &7,029 \\
        & Harmful & Base & 83 & 592 \\
        & Harmless & Steer & 97 &1,051 \\
        & Harmless & Base & 97 & 7,322\\
        \midrule
        \multirow{4}{*}{L3 NTP} & Harmful & Steer & 81 & 5,866 \\
        & Harmful & Base & 81 & 477 \\
        & Harmless & Steer & 100 & 501 \\
        & Harmless & Base & 100 & 6,800 \\
        \midrule
        \multirow{4}{*}{L3 PO} & Harmful & Steer & 79 & 5,635 \\
        & Harmful & Base & 79 & 958 \\
        & Harmless & Steer & 100 & 669\\
        & Harmless & Base & 100 & 17,803\\
        \midrule
        \multirow{4}{*}{Q8 DIM} & Harmful & Steer & 79 & 6,751 \\
        & Harmful & Base & 79 & 4,687 \\
        & Harmless & Steer & 100 & 2,284 \\
        & Harmless & Base & 100 & 9,786 \\
        \toprule
    \end{tabular}
    \caption{Activation Patching Dataset Sizes for each Model-Vector pair (M $\times$ V). \# Resp. is the number of multi-token responses, while \# Pos. is the total number of response tokens.}
    \label{tab:act_patching_dataset_sizes}
\end{table}
\subsection{NTP and PO Formulations}
Next Token Prediction (NTP) uses the language modeling objective to learn a steering vector on prompt-response pairs that express the desired concept \cite{wu2025axbenchsteeringllmssimple}. 
Given a dataset $D^+$ with prompts $x$ and responses $y$ that express the desired steering concept, the steering vector is learned with the objective
\begin{equation}
    \min \sum_{x,y \in D^+} \left\{- \sum_{i=1}^k \log p(y_i |  x, h^l \leftarrow h^l + \alpha v) \right \}
\label{eq:ntp}
\end{equation}
where $k$ is the number of generated tokens per sequence, $\alpha$ is the steering coefficient, and $v$ is the steering vector. 

\textbf{Preference Optimization} (PO) learns a steering vector by using two contrastive datasets \cite{cao2024personalized, rafailov2023direct}. In our experiments, we use the uni-directional form of RePS \cite{wu2025improved}. Following \citet{wu2025axbenchsteeringllmssimple}, given a desired response $y^w$ and an undesired response $y^l$ to prompt $x$, the log probability difference $\Delta_{x,y^w, y^l}$ is 
\begin{equation}
\begin{split}
\frac{\beta^+}{|y^w|} \log \left(p(y^w | x, h^\ell \leftarrow h^\ell + \alpha v)\right) - \\
\frac{1}{|y^l|} \log \left(p(y^l | x, h^\ell \leftarrow h^\ell + \alpha v) \right)
\end{split}
\label{eq:reps_positive}
\end{equation}
Where $\beta^+ = \max(\log(p(y^l|x)) - \log (p(y^w | x)) \cdot \phi, 1)$ serves as a scaling term to weight the log likelihood of $y^w$ more if the reference model considers $y^w$ unlikely. $\phi$ is a positive temperature scalar.
We optimize the objective
\begin{equation}
    \min \sum_{x, y^w, y^l \in D} \left\{ -\log \sigma (\Delta_{x,y^w, y^l}) \right\}
\label{eq:po}
\end{equation}

We follow \autoref{eq:ntp} to train NTP and \autoref{eq:po} to train the PO vector. Each steering vector for each model is trained on the same injection layer as the DIM vector, so layer 15 for Gemma 2 2B and layer 12 for Llama 3.2 3B.

Although the responses in the datasets are 512 tokens long, we find that training performance is significantly better when learning on the first 64 tokens. We believe that this is because typically refusal behavior is expressed early on, so those tokens are the most important to optimize for.

We sweep over the hyperparameter grid in \autoref{tab:hyperparams} for the NTP and PO vectors for each model.
We select the best steering vector for each steering method on each model by evaluating the average loss on the validation dataset, using each steering vector's respective loss objective.
For PO, since the loss depends on $\phi$, we cannot directly compare the validation loss for vectors learned on different $\phi$.
Thus, we first select the best steering vector for each trained $\phi$. Then, we compute the \emph{match score}-- the fraction of response tokens where the greedy decoded steered prediction matches the validation ground truth. The candidate vector with the highest match score is selected as the overall best PO vector.

The best hyperparameters are in \autoref{tab:best-hyperparams}.
After training the NTP and PO vectors, we evaluate them on JailbreakBench and Alpaca test sets following the methodology described in \cref{sec:3steering_basics}. The Attack Success Rates are shown in \autoref{tab:refusal_asr_evals}.

\subsection{Faithfulness Curves}
\begin{figure*}[htbp]
    \centering
    \begin{subfigure}{0.48\textwidth}
        \centering
        \includegraphics[width=\linewidth]{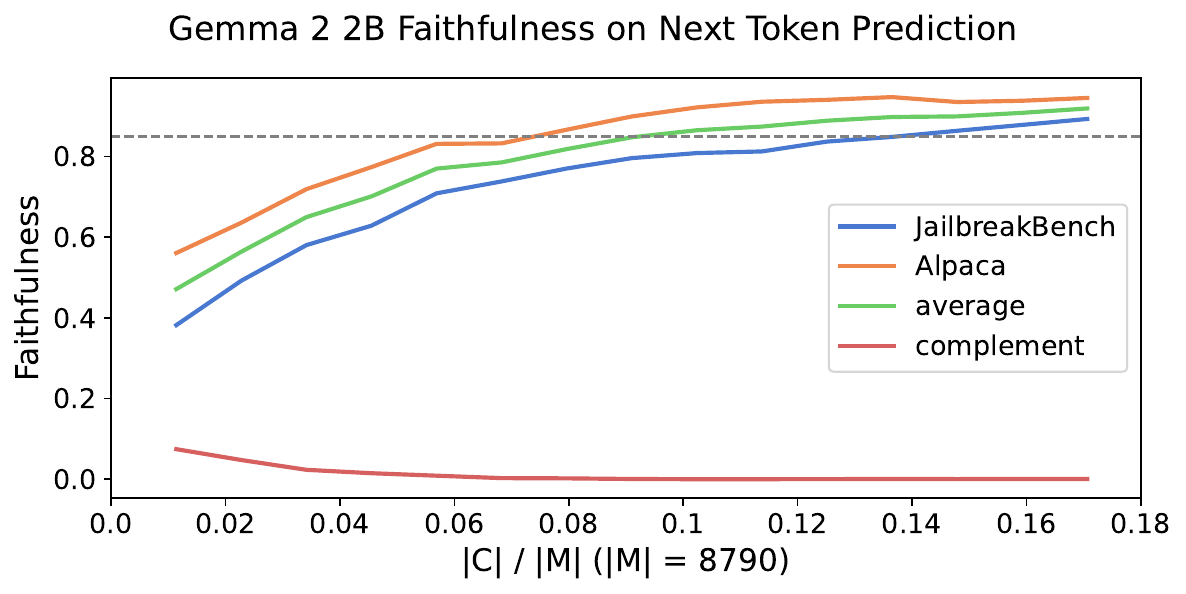}
        \caption{Gemma 2 2B, NTP}
        \label{fig:faith_g2_ntp}
    \end{subfigure}
    \hfill
    \begin{subfigure}{0.48\textwidth}
        \centering
        \includegraphics[width=\linewidth]{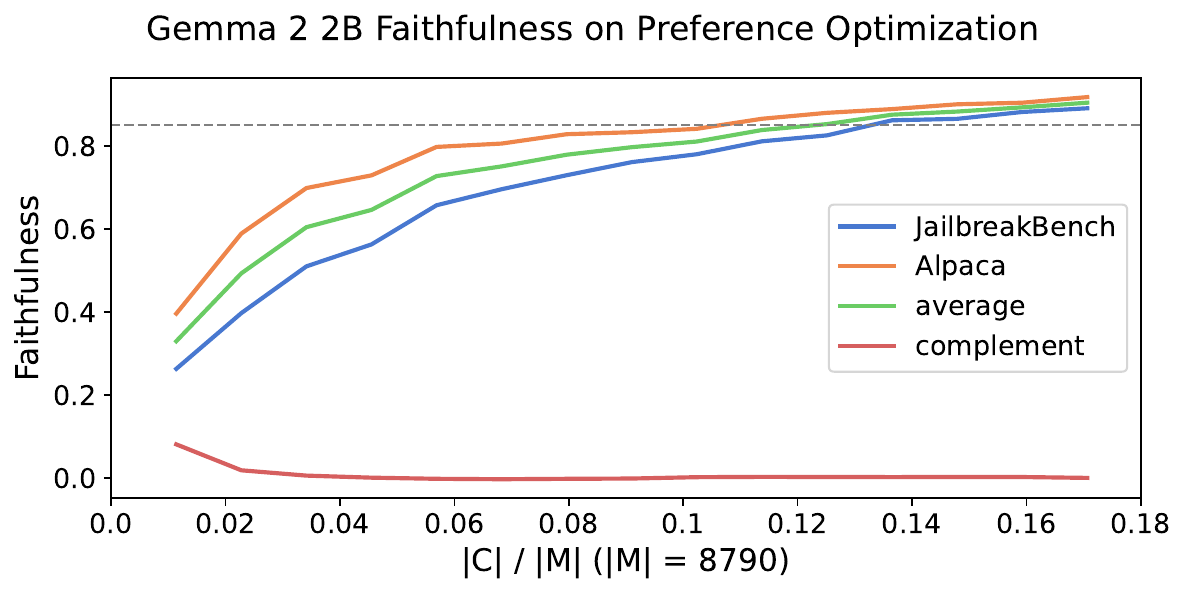}
        \caption{Gemma 2 2B, PO}
        \label{fig:faith_g2_po}
    \end{subfigure}
    \begin{subfigure}{0.48\textwidth}
        \centering
        \includegraphics[width=\linewidth]{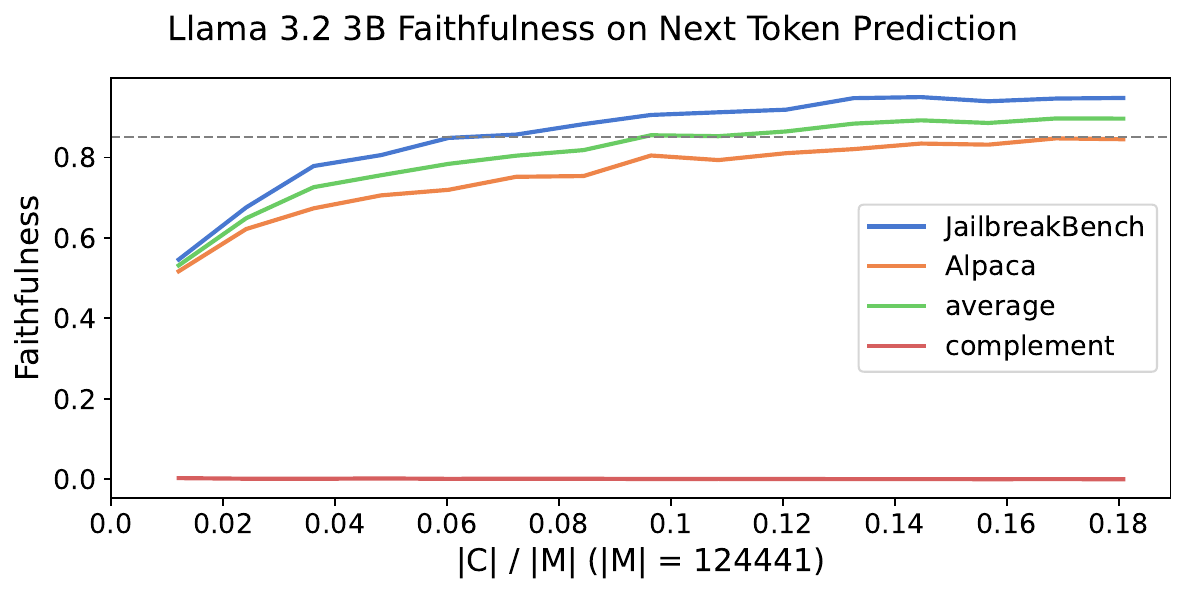}
        \caption{Llama 3.2 3B, NTP}
        \label{fig:faith_l3_ntp}
    \end{subfigure}
    \hfill
    \begin{subfigure}{0.48\textwidth}
        \centering
        \includegraphics[width=\linewidth]{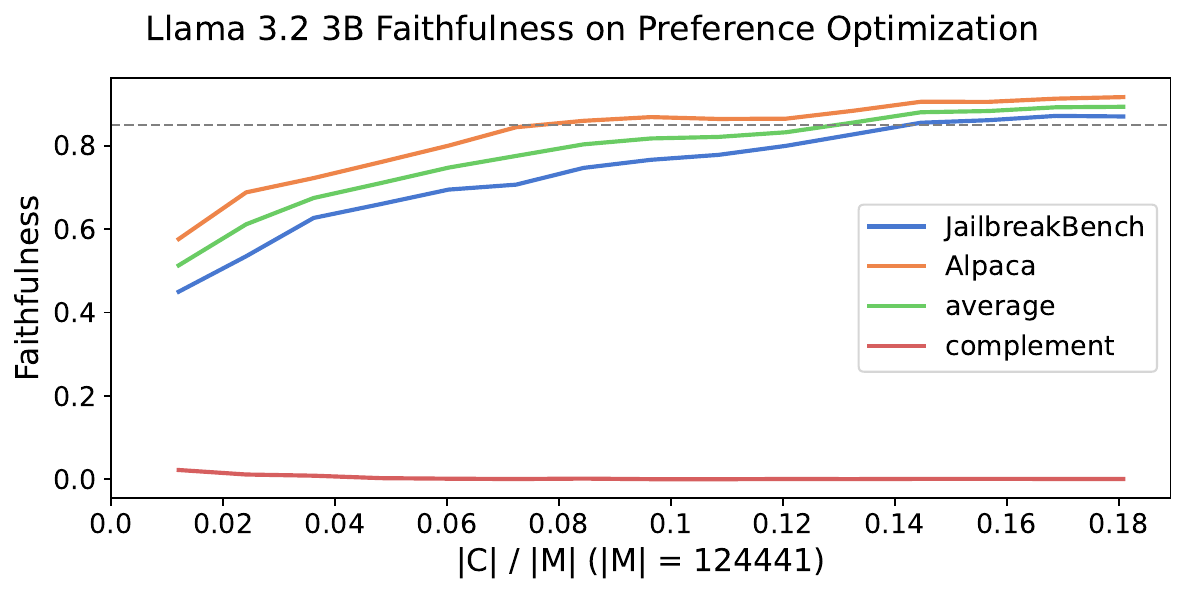}
        \caption{Llama 3.2 3B, PO}
        \label{fig:faith_l3_po}
    \end{subfigure}
    \caption{Faithfulness curves for NTP and PO objectives per evaluation dataset}
    \label{fig:faith_ntp_po}
\end{figure*}

The main paper shows the individual faithfulness curves for JailbreakBench and Alpaca for the DIM vector on Gemma 2 2B, Llama 3.2 3B, and Qwen 3 8B. \autoref{fig:faith_ntp_po} shows faithfulness curves for the Gemma and Llama NTP and PO vectors.

\begin{table}[htbp]
    \centering
    \begin{tabular}{c|c}
        \toprule
        Hyperparameters & Search Space \\
        \toprule
        Batch Size & {6, 12} \\
        Learning Rate & {0.01, 0.04} \\
        Epochs & 10 \\
        L2 Weight Decay & {0} \\
        Optimizer & Adam \\
        LR Scheduler & Linear \\
        Seeds & {42, 5} \\
        $\phi$ & {2e-2, 1e-5} \\
        \toprule
    \end{tabular}
    \caption{Training Hyperparameters for Gemma 2 2B and Llama 3.2 3B NTP and PO vectors.}
    \label{tab:hyperparams}
\end{table}
\begin{table}[htbp]
    \centering
    \begin{tabular}{l|ccccc}
        \toprule
        M $\times$ V & BSZ & LR & Epoch & Seed & $\phi$ \\
        \toprule
        G2 NTP & 6 & 0.01 & 8 & 42 & - \\
        G2 PO & 6 & 0.01 & 1 & 5 & 2e-2 \\
        L3 NTP & 12 & 0.01 & 9 & 5 & -  \\
        L3 PO & 12 & 0.01 & 2 & 5 & 1e-5  \\
        \toprule
    \end{tabular}
    \caption{Hyperparameters for best Gemma 2 2B and Llama 3.2 3B NTP and PO vectors.}
    \label{tab:best-hyperparams}
\end{table}

\section{Additional Model Results}

We provide additional results for Llama 3.2 3B Instruct and Qwen 3 8B here that mirror the Gemma 2 2B results in the main paper.

\subsection{Faithfulness}
\begin{figure*}[htbp]
    \centering
    \includegraphics[width=0.98\linewidth]{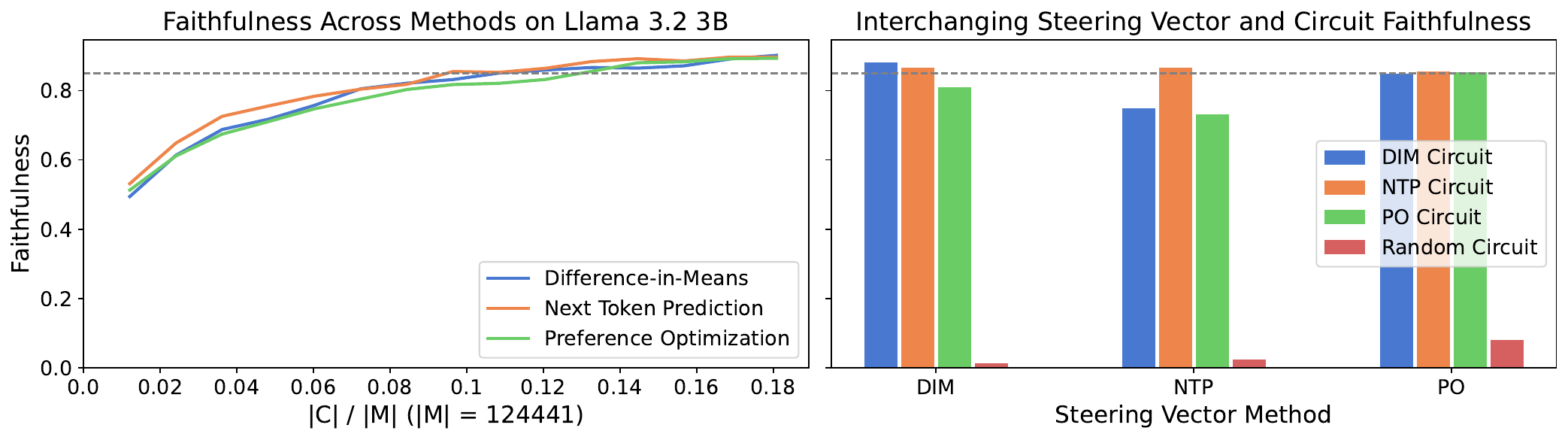}
    \caption{\textbf{Left} Average faithfulness across circuit sizes for each steering method on Llama 3.2 3B. 
    \textbf{Right} For each steering method, we compute faithfulness using its own circuit as well as the circuits obtained from \emph{other} methods. We also compare against a random circuit at 2x the minimum-faithful size, which performs poorly.}
    \label{fig:faith_method_comparison_l3}
\end{figure*}

\autoref{fig:faith_method_comparison_l3} shows faithfulness on DIM, NTP, and PO vectors, as well as faithfulness using interchanged circuits for Llama 3.2 3B. Computing the faithfulness of each steering vector using a circuit obtained from a different steering vector still retains most faithfulness. 
Following the minimum number of edges needed per steering vector to achieve faithfulness $\ge 0.85$, we compute faithfulness on the DIM vector with 13,500 edges, the NTP vector with 12,000 edges, and the PO vector with 16,500 edges.
\subsection{SVV}
\begin{figure*}[hbtp]
    \centering
    \includegraphics[width=0.95\linewidth]{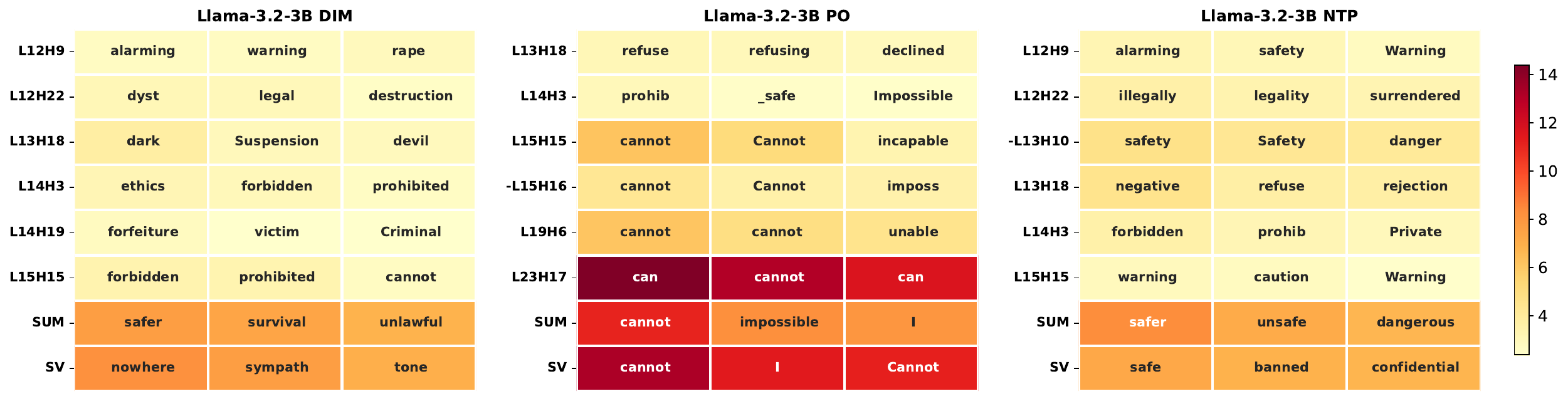}
    \caption{For each steering method on Llama 3.2 3B, we use logit lens on the raw steering vector (SV), the SVV of top attention heads (LayerXHeadY), and the sum of all svvs (SUM). We prepend the names of sign-flipped svvs with (-). We select tokens from the top 20 tokens and display their logit values. svvs surface semantically interpretable tokens related to harmfulness/refusal, even when the raw SV does not (DIM). }
    \label{fig:svv_heatmap_l3}
\end{figure*}

\begin{figure}[hbtp]
    \centering
    \includegraphics[width=0.8\linewidth]{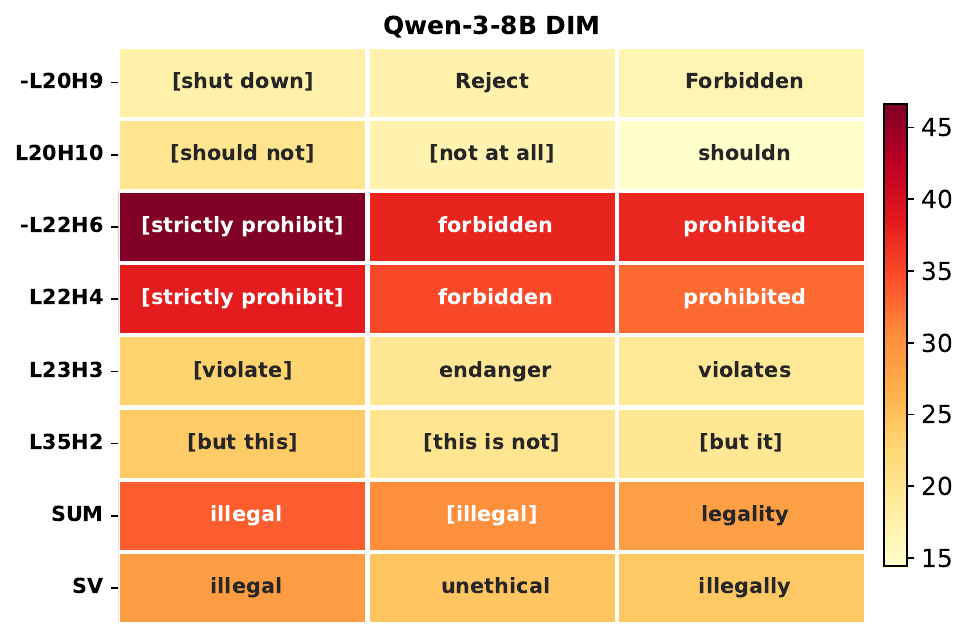}
    \caption{For DIM steering on Qwen 3 8B, we use logit lens on the raw steering vector (SV), the SVV of top attention heads (LayerXHeadY), and the sum of all svvs (SUM). We prepend the names of sign-flipped svvs with (-). We select tokens from the top 20 tokens and display their logit values. svvs surface semantically interpretable tokens related to harmfulness/refusal. }
    \label{fig:svv_heatmap_q8}
\end{figure}

Following the process described in \cref{sec:attention:svv},
we apply logit lens on the svvs of top attention heads by $IE$ on Llama 3.2 3B and Qwen 3 8B. We display selected tokens in \autoref{fig:svv_heatmap_l3} and \autoref{fig:svv_heatmap_q8}.
Although the Llama 3.2 3B DIM steering vector itself does not exhibit tokens related to refusal or harmfulness, individual svvs and the sum of all svvs do.
\subsection{Sparsity}
\begin{figure*}[htbp]
    \centering
    \begin{subfigure}{0.95\textwidth}
    \includegraphics[width=\linewidth]{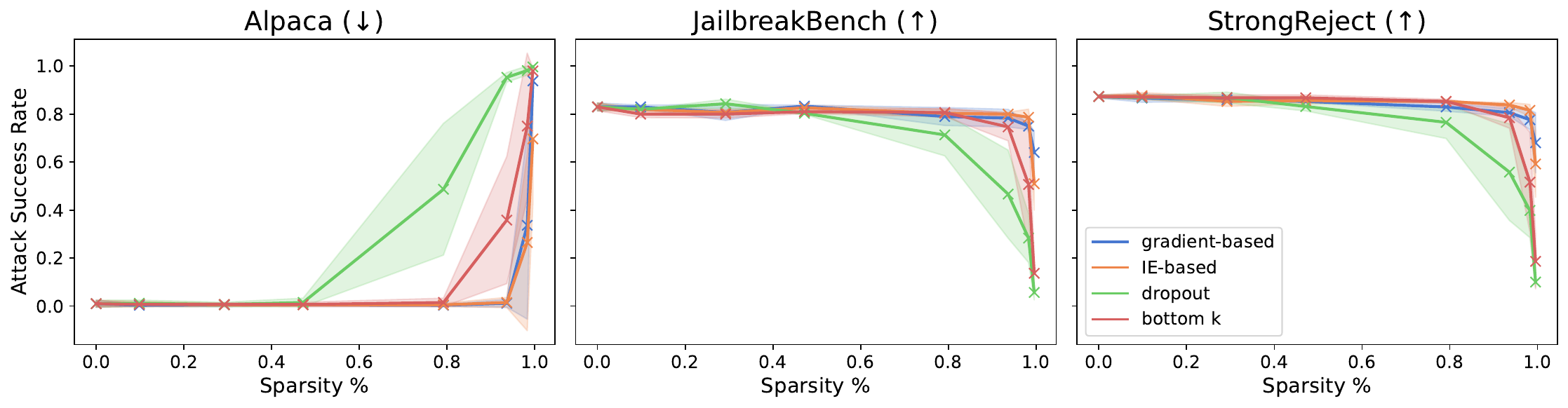}
    \caption{Llama 3.2 3B Instruct averaged across DIM, NTP, and PO vectors}
    \end{subfigure}
    \begin{subfigure}{0.95\textwidth}
    \includegraphics[width=\linewidth]{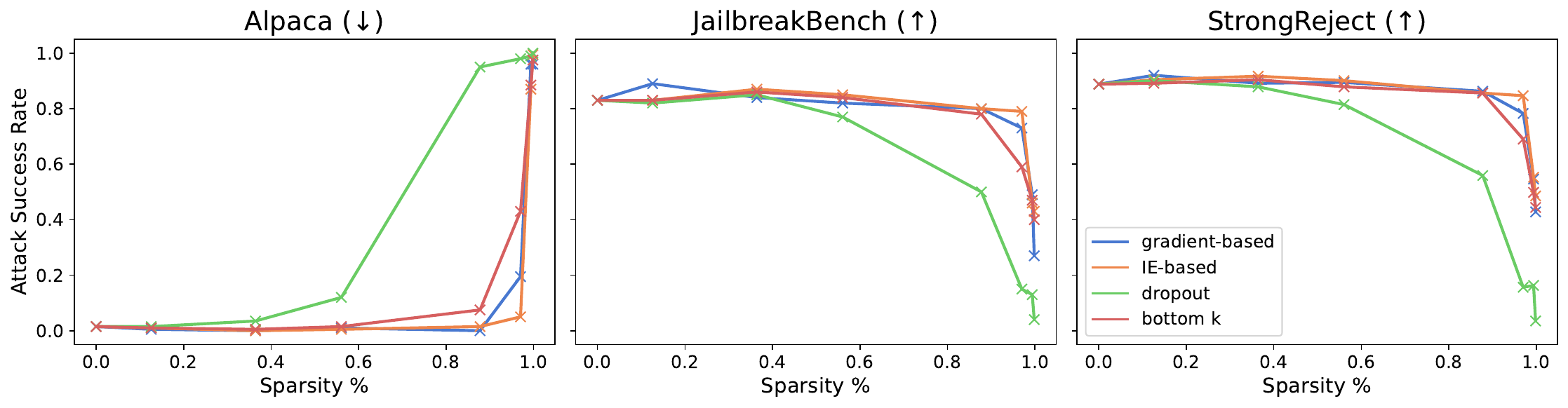}
    \caption{Qwen 3 8B non-thinking mode DIM vector}
    \end{subfigure}
    \caption{We sparsify $\mathbf{s}$ at thresholds $r_i < \tau = \{0.0, 0.1, 0.3, 0.5, 1.0, 1.5, 2.0, 2.5\}$, marked by \textit{x}'s. Attribution Patching-Based sparsification retains ASR up to \textasciitilde96\% sparsity on Llama 3.2 3B and \textasciitilde88\% on Qwen 3 8B.}
    \label{fig:sparsity_asr_l3_q8}
\end{figure*}

Following attribution patching-based sparsification in \cref{sec:sparsity:grad}, we evaluate the attack success rate of Llama 3.2 3B's DIM, NTP, and PO vectors and Qwen 3 8B's DIM vector at varying sparsity $\tau$. We plot results in \autoref{fig:sparsity_asr_l3_q8}.

\section{Metric Comparisons}
\label{appendix:F_metric_comparisons}
\subsection{Dataset-Specific Activation Patching}
\begin{figure*}[htbp]
    \centering
    \begin{subfigure}{0.95\textwidth}
        \centering
        \includegraphics[width=\linewidth]{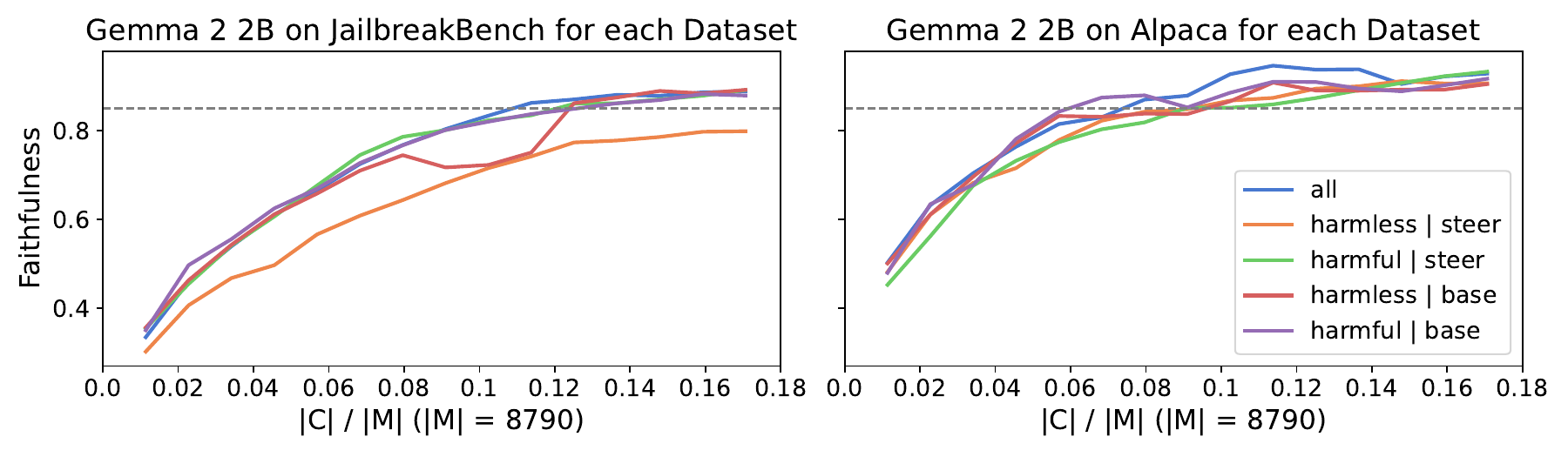}
        \label{fig:faith_datasets_g2}
    \end{subfigure}
    \begin{subfigure}{0.95\textwidth}
        \centering
        \includegraphics[width=\linewidth]{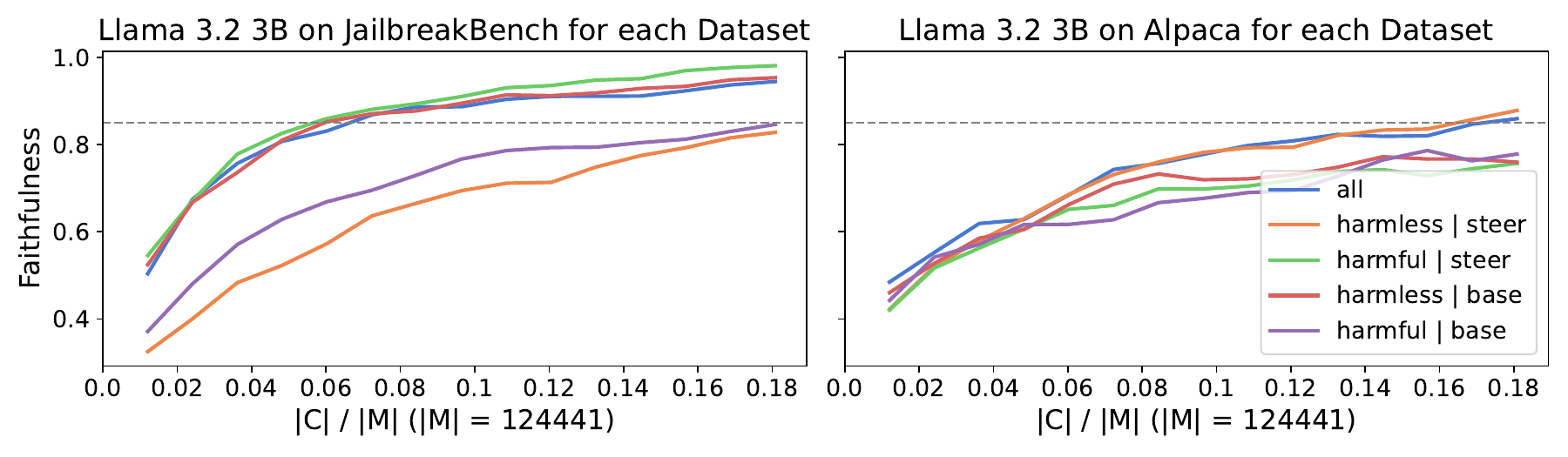}
        \label{fig:faith_datasets_l3}
    \end{subfigure}
    \caption{Faithfulness on circuits obtained from individual datasets for Llama 3.2 3B and Gemma 2 2B. The ``all'' dataset represents using all four datasets.}
    \label{fig:faith_datasets}
\end{figure*}

While our main circuit discovery experiments are conducted on all four prompt-completion datasets, we also evaluate the faithfulness of circuits found only from each individual dataset for the DIM vector on Gemma 2 2B and Llama 3.2 3B.
As shown in \autoref{fig:faith_datasets},
individual datasets are largely capable of achieving circuit faithfulness, and in some cases, outperform circuits obtained from all datasets.
Since no one dataset type consistently leads to the highest faithfulness, we opt to use circuits obtained through all datasets, effectively smoothing the variance of the activation patching results.

\subsection{Directional KL Divergence}
The steering vector affects the entire output distribution of the model prediction, not just the top token.
Thus, the logit difference metric, $m(x') = \text{logit}(y | x') - \text{logit}(y^* | x')$, may not capture all the effects of the steering vector.
Thus, we validate the robustness of the logit difference metric by using a
\textit{Directional KL Divergence} (DirKL) metric:
\[
m(x') = \text{KL}(P_{x^*} || P_{x'}) - \text{KL}(P_{x} || P_{x'})
\]
which measures the KL divergence of the output distribution $P_{x'}$ of the patched input with respect to both the clean and corrupt input's output distributions $P_x, P_{x^*}$.
Note that this relative measure differs from the vanilla KL divergence metric $m(x') = \text{KL}(P_{x^*} || P_{x'})$ \cite{conmy2023automatedcircuitdiscoverymechanistic}, which measures the absolute divergence away from $P_{x^*}$ and ignores the clean distribution $P_{x_{steer}}$. Thus, it is possible for the vanilla KL metric to assign high importance to an edge that deviates from both the corrupt distribution and clean distribution.

Similarly to masking positions where the steered and base forward passes agree on the greedy prediction, we mask out positions that have $\text{KL}(P_{x_{steer}} || P_{x_{base}})$ below a pre-defined threshold parameter. We examine thresholds $\eta=0, 1, 5$. The total number of positions evaluated (unmasked) for each metric is shown in \autoref{tab:metric_pos_evaluated}.

\begin{figure*}[htbp]
    \centering
    \includegraphics[width=0.95\linewidth]{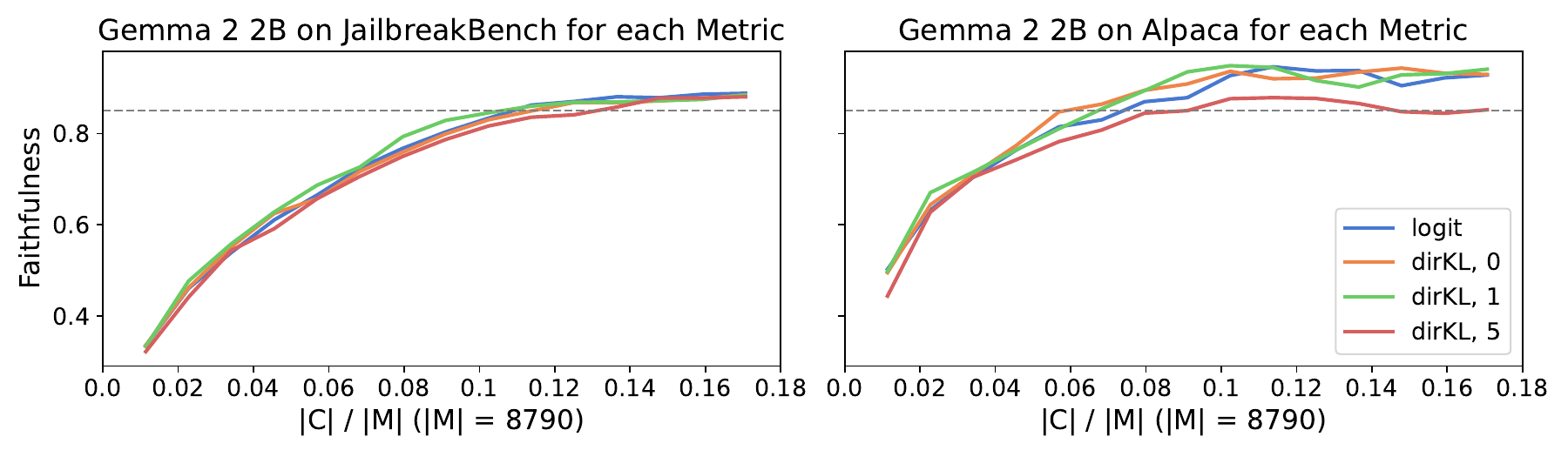}
    \includegraphics[width=0.95\linewidth]{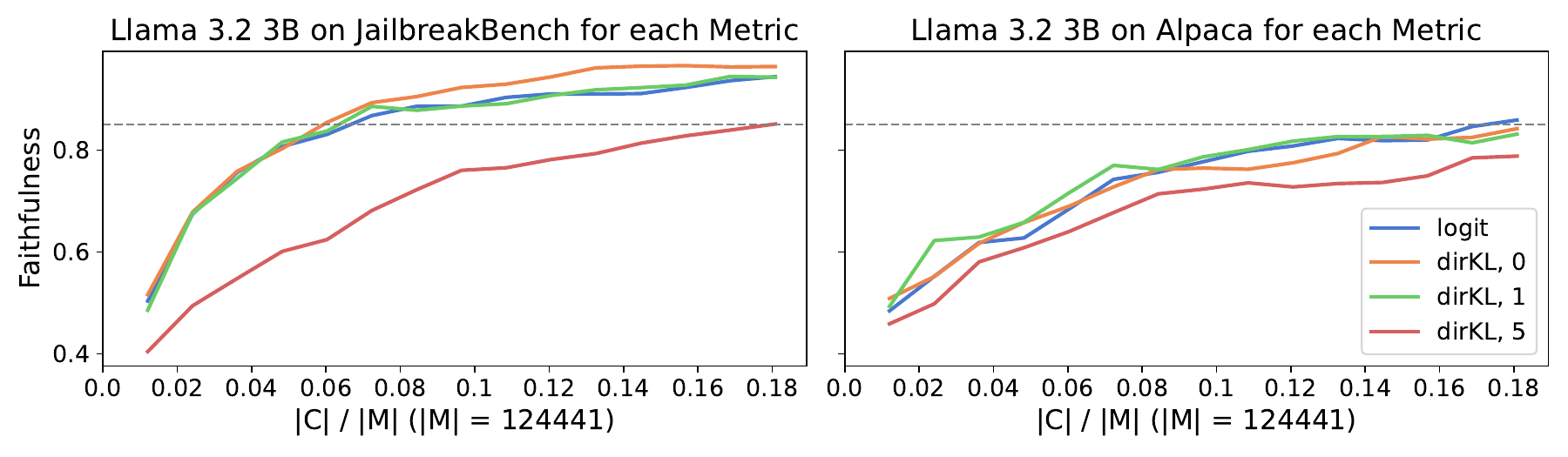}
    \caption{Evaluation of the importance metrics logit difference and directional KL divergence at thresholds 0, 1, 5 on Gemma 2 2B and Llama 3.2 3B with the DIM vector. Evaluations are shown per eval dataset.}
    \label{fig:faith_all_metrics}
\end{figure*}

\paragraph{Faithfulness Across Metrics}
We test the faithfulness of circuits obtained with thresholds of $\eta=0, 1, 5$ on the DIM vector for Gemma 2 2B. As shown in the \autoref{fig:faith_all_metrics}, 
the logit difference metric performs similarly to the DirKL metric. 
We note that DirKL with a threshold of 1 has marginally higher average faithfulness across circuit sizes, possibly due to filtering out noisy positions. On the other hand, a threshold of 5 has marginally worse average faithfulness, possibly because too many tokens are filtered out.

Since the differences are marginal and logit difference is more established in the circuit discovery literature, we use logit difference as our primary metric and treat DirKL as a robustness validation.

\begin{table}[htbp]
    \centering
    \begin{tabular}{c|c}
        Metric & Num Positions Evaluated \\
        \toprule
        DirKL, $\eta=0$ & 153896 \\
        \midrule
        DirKL, $\eta = 1$ & 31100 \\
        \midrule
        DirKL, $\eta = 5$ & 2423 \\
        \midrule
        Logit & 40473 \\
        \toprule
    \end{tabular}
    \caption{Number of positions evaluated across all four prompt-completion datasets for Gemma 2 2B Instruct on the DIM vector.}
    \label{tab:metric_pos_evaluated}
\end{table}

\section{Edge Attribution Patching with Integrated Gradients}
\label{appendix:H_eap_ig}
\subsection{EAP-IG formulation}\label{appendix:eap_ig_deriv}
Given a function $f$, we can approximate $f(b) - f(a) = \int_a^{b} \nabla_x f(x) dx$ as
\[
\sum_{i=1}^T \frac{1}{T} (b - a) \nabla_x f(x) |_{x = (a + (i/T) (b - a))} 
\]
for $T$ intermediate steps. EAP-IG \cite{hanna2024have} applies this approximation to activation patching. Setting $f$ as the metric function $m$, the 
$IE$ can be approximated as
\begin{equation*}
(u - u^*)^\top \frac{1}{T} \left(\sum_{i=1}^T \frac{\partial m(x^* + \frac{i-0.5}{T}(x - x^*))}{ \partial v} \right)
\end{equation*}
Note that in the original implementation, the gradients are taken at intervals $\frac{i}{T}$ rather than $\frac{i-0.5}{T}$.
We use the midpoint quadrature rule, which provides $O(1/T^2)$ approximation error compared to $O(1/T)$
for endpoint rules \cite{driscoll2017numericalmethods}, while incurring no additional computational cost.

Similarly to computing the $IE$ of edge $(u,v)$ in \autoref{eq:eapig}, we can compute the $IE$ of node $u$ following the same equation but taking the partial derivative with respect to $u$.

\subsection{Deriving \autoref{eq:eapig_steer}}\label{appendix:eap_ig_steer_deriv}

Since we aim to understand how steering behavior is achieved, we set $H = H_{steer}$ as the clean steered representation and $H^* = H_{base}$ as the corrupt base representation.
We replace $x$ with $H$ and $x^*$ with $H^*$ in \autoref{eq:eapig} to approximate the $IE$ of an edge $(u,v)$ as
\begin{equation*}
(u - u^*)^\top \frac{1}{T} \left(\sum_{i=1}^T \frac{\partial m(H^* + \frac{i-0.5}{T}(H - H^*))}{\partial v} \right)
\end{equation*}
When the steering vector is applied to every token in a teacher-forced sequence, $H - H^* = H_{steer} - H_{base} =  \alpha \cdot S$, so the expression simplifies to 
\begin{equation}
(u - u^*)^\top \frac{1}{T} \left(\sum_{i=1}^T \frac{\partial m(H^\ell_{base} + \frac{i-0.5}{T}\alpha \cdot S)}{\partial v} \right)
\end{equation}
Thus, we take the gradients at increasing linear scales of steering coefficient $\alpha$.
\paragraph{Node IE}
The $IE$ of a node $u$ is 
\begin{equation}
(u - u^*)^\top \frac{1}{T} \left(\sum_{i=1}^T \frac{\partial m(H^\ell_{base} + \frac{i-0.5}{T}\alpha \cdot S)}{\partial u} \right)
\end{equation}
where the only modification is that the partial derivative is taken with respect to $u$ instead of $v$.

\section{Sparsity Raw Results}

\label{appendix:D_sparsity}
\paragraph{Raw Sparsity Results}
Raw sparsity results are shown in \autoref{tab:sparsity_g2}, \autoref{tab:sparsity_l3}, and \autoref{tab:sparsity_q8}.
We first sparsify vectors using gradient-based sparsification at thresholds $\tau \in \{0, 0.1, 0.3, 0.5, 1, 1.5, 2, 2.5\}$. This results in similar yet slightly different $k_i$ and sparsity percentages $k_i/d$ for each DIM, NTP, and PO vector for each model, as shown in tables. We then apply $IE$-based, random dropout, and bottom-$k$ sparsification on each steering vector using each steering vector's respective $k_i$.
We bold the best sparsification method and underline ties for each steering vector and dataset at each sparsity threshold.

\paragraph{Perplexity}
We check if refusal steering with the dense and sparse vectors preserves overall generation quality.
We report perplexity for Pile \cite{gao2020pile800gbdatasetdiverse} and Alpaca on the Gemma 2 2B, Llama 3.2 3B, and Qwen 3 8B DIM vectors in \autoref{tab:sparsity-ppl}. Perplexity rises gradually in the ASR-preserving, lower sparsity thresholds and degrades most on Pile beyond \textasciitilde85\% sparsity ($\tau=1$), which is consistent with that being near the effective threshold of sparse steering. 

\begin{table*}[htbp]
\centering
\begin{tabular}{llccccccccc}
\toprule
\multirow{2}{*}{Model} & \multirow{2}{*}{Data} & \multirow{2}{*}{No Steer} & \multicolumn{7}{c}{Sparsity threshold $\tau$} \\
\cmidrule(lr){4-10}
 & & & $0$ & $0.1$ & $0.3$ & $0.5$ & $1$ & $1.5$ & $2$ & 2.5 \\
\midrule
\multirow{2}{*}{Gemma 2 2B} & Pile & 12.84 & 17.63 & 17.71 & 17.95 & 18.80 & 20.07 & 20.96 & 21.25 & 21.33 \\
& Alpaca & 6.90 & 8.79  & 8.88  & 8.87  & 9.04 & 9.71  & 9.15 & 8.88 & 8.96 \\
\midrule
\multirow{2}{*}{Llama 3.2 3B} & Pile & 11.99 & 15.79 & 15.89 & 16.24 & 16.56 & 17.40&17.85 & 19.39 & 20.89 \\
& Alpaca & 6.38 & 7.27  & 7.30  & 7.51  & 7.66 & 7.80 & 7.62  & 7.78  & 8.30  \\
\midrule
\multirow{2}{*}{Qwen 3 8B} & Pile & 9.45 & 11.45 & 11.48 & 11.73 & 11.89 & 13.17 & 12.59 & 14.71 & 16.10  \\
& Alpaca & 9.32 & 8.78 & 8.79 & 9.28 & 9.29 & 11.88 & 10.59 & 13.10 & 14.32 \\
\bottomrule
\end{tabular}
\caption{Perplexity at increasing sparsity thresholds $\tau$ for DIM vectors. Realized sparsity differs slightly by model: Gemma 2 2B reaches 0\%, 11.6\%, 33.2\%, 53.2\%, 83.7\%, 94.9\%, 97.8\%, 99.2\%; Llama 3.2 3B reaches 0\%, 10.2\%, 31.5\%, 51.8\%, 83.5\%, 96.2\%, 99.3\%, 99.7\%; Qwen 3 8B reaches 0\%, 12.6\%, 36.4\%, 56.1\%, 87.8\%, 97.0\%, 99.4\%, 99.9\%.}
\label{tab:sparsity-ppl}
\end{table*}

\begin{table*}[htbp]
\centering
\caption{Sparsification results for gradient-based (notated Grad.), IE, random dropout (Drop.), and bottom-k (Bot-$k$) on Gemma 2 2B. ASR reported across three datasets. Sparsity percentage is notated as S (\%).
$\downarrow$ = lower is better (Alpaca); $\uparrow$ = higher is better (JailbreakBench, StrongReject).}
\label{tab:sparsity_g2}
\small
\begin{tabular}{ll |cccc| cccc| cccc}
\toprule
& & \multicolumn{4}{c}{Alpaca ($\downarrow$)} & \multicolumn{4}{c}{JailbreakBench ($\uparrow$)} & \multicolumn{4}{c}{StrongReject ($\uparrow$)} \\
\cmidrule(lr){3-6} \cmidrule(lr){7-10} \cmidrule(lr){11-14}
Vector & S (\%) & Grad. & IE & Drop. & Bot-$k$ & Grad. & IE & Drop. & Bot-$k$ & Grad. & IE & Drop. & Bot-$k$ \\
\midrule
 \multirow{8}{*}{DIM} & 0.0 & 0.000 & 0.000 & 0.000 & 0.000 & 0.800 & 0.800 & 0.800 & 0.800 & 0.850 & 0.850 & 0.850 & 0.850 \\
  & 11.6 & 0.000 & 0.000 & 0.000 & 0.000 & 0.770 & 0.790 & 0.810 & \textbf{0.820} & 0.824 & \textbf{0.856} & 0.853 & 0.843 \\
  & 33.2 & 0.000 & 0.000 & 0.010 & 0.000 & 0.800 & 0.760 & 0.750 & \textbf{0.820} & 0.843 & 0.840 & 0.843 & \textbf{0.847} \\
  & 53.2 & 0.005 & \underline{0.000} & 0.030 & \underline{0.000} & 0.780 & \textbf{0.800} & 0.520 & 0.790 & \underline{0.843} & 0.837 & 0.601 & \underline{0.843} \\
  & 83.7 & 0.005 & \textbf{0.000} & 0.875 & 0.025 & 0.770 & 0.770 & 0.120 & 0.610 & \textbf{0.831} & 0.827 & 0.077 & 0.744 \\
  & 94.9 & \textbf{0.065} & 0.070 & 0.875 & 0.480 & 0.600 & \textbf{0.760} & 0.140 & 0.180 & 0.645 & \textbf{0.741} & 0.125 & 0.294 \\
  & 97.8 & \textbf{0.400} & 0.540 & 0.980 & 0.875 & 0.270 & \textbf{0.380} & 0.030 & 0.080 & 0.323 & \textbf{0.521} & 0.022 & 0.093 \\
  & 99.2 & \textbf{0.805} & 0.895 & 0.960 & 0.945 & \textbf{0.250} & 0.110 & 0.030 & 0.040 & \underline{0.252} & \underline{0.252} & 0.026 & 0.042 \\
\midrule
 \multirow{8}{*}{NTP} & 0.0 & 0.030 & 0.030 & 0.030 & 0.030 & 0.840 & 0.840 & 0.840 & 0.840 & 0.831 & 0.831 & 0.831 & 0.831 \\
  & 9.4 & \underline{0.015} & \underline{0.015} & 0.035 & 0.020 & 0.830 & \underline{0.850} & 0.790 & \underline{0.850} & \textbf{0.863} & 0.824 & 0.792 & 0.812 \\
  & 27.3 & \underline{0.015} & 0.025 & 0.045 & \underline{0.015} & 0.830 & 0.830 & 0.720 & \textbf{0.840} & \textbf{0.869} & 0.843 & 0.719 & 0.827 \\
  & 43.4 & 0.015 & \textbf{0.010} & 0.170 & 0.020 & \textbf{0.860} & 0.850 & 0.610 & 0.820 & \textbf{0.891} & 0.859 & 0.597 & 0.815 \\
  & 74.7 & \textbf{0.005} & 0.020 & 0.900 & 0.090 & 0.830 & \textbf{0.840} & 0.150 & 0.790 & \textbf{0.869} & 0.875 & 0.128 & 0.735 \\
  & 90.1 & \textbf{0.040} & 0.130 & 0.865 & 0.680 & \underline{0.820} & \underline{0.820} & 0.180 & 0.470 & 0.840 & \textbf{0.863} & 0.157 & 0.495 \\
  & 96.3 & 0.520 & \textbf{0.270} & 0.960 & 0.870 & 0.350 & \textbf{0.520} & 0.050 & 0.170 & 0.345 & \textbf{0.703} & 0.026 & 0.188 \\
  & 98.5 & \textbf{0.805} & 0.830 & 0.975 & 0.955 & 0.180 & \textbf{0.330} & 0.030 & 0.090 & 0.144 & \textbf{0.428} & 0.019 & 0.080 \\
\midrule
 \multirow{8}{*}{PO} & 0.0 & 0.000 & 0.000 & 0.000 & 0.000 & 0.800 & 0.800 & 0.800 & 0.800 & 0.824 & 0.824 & 0.824 & 0.824 \\
  & 10.9 & 0.000 & 0.000 & 0.000 & 0.000 & 0.760 & \underline{0.780} & 0.760 & \underline{0.780} & \underline{0.815} & 0.808 & 0.783 & \underline{0.815} \\
  & 32.2 & 0.000 & 0.000 & 0.000 & 0.000 & 0.730 & 0.740 & \textbf{0.750} & 0.740 & 0.767 & 0.796 & 0.789 & \textbf{0.812} \\
  & 50.3 & 0.000 & 0.000 & 0.005 & 0.000 & 0.780 & 0.750 & 0.780 & 0.780 & 0.760 & 0.744 & \textbf{0.824} & 0.815 \\
  & 80.9 & 0.000 & 0.000 & 0.015 & 0.000 & 0.730 & 0.750 & 0.340 & \textbf{0.810} & 0.725 & 0.792 & 0.441 & \textbf{0.840} \\
  & 94.4 & \underline{0.000} & \underline{0.000} & 0.815 & 0.340 & 0.660 & \textbf{0.690} & 0.170 & 0.590 & 0.642 & \textbf{0.703} & 0.201 & 0.633 \\
  & 98.4 & 0.120 & \textbf{0.115} & 0.975 & 0.925 & \textbf{0.440} & 0.390 & 0.040 & 0.120 & \textbf{0.534} & 0.447 & 0.032 & 0.125 \\
  & 99.6 & 0.910 & \underline{0.890} & 0.970 & 0.980 & 0.030 & \textbf{0.060} & 0.010 & 0.030 & 0.070 & \textbf{0.077} & 0.006 & 0.010 \\
\bottomrule
\end{tabular}
\end{table*}

\begin{table*}[htbp]
\centering
\caption{Sparsification results for gradient-based (notated Grad.), IE, random dropout (Drop.), and bottom-$k$ (Bot-$k$) on Llama 3.2 3B. ASR reported across three datasets. Sparsity percentage is notated as S (\%).
$\downarrow$ = lower is better (Alpaca); $\uparrow$ = higher is better (JailbreakBench, StrongReject).}
\label{tab:sparsity_l3}
\small
\begin{tabular}{ll |cccc| cccc| cccc}
\toprule
& & \multicolumn{4}{c}{Alpaca ($\downarrow$)} & \multicolumn{4}{c}{JailbreakBench ($\uparrow$)} & \multicolumn{4}{c}{StrongReject ($\uparrow$)} \\
\cmidrule(lr){3-6} \cmidrule(lr){7-10} \cmidrule(lr){11-14}
Vector & S (\%) & Grad. & IE & Drop. & Bot-$k$ & Grad. & IE & Drop. & Bot-$k$ & Grad. & IE & Drop. & Bot-$k$ \\
\midrule
 \multirow{8}{*}{DIM} & 0.0 & 0.030 & 0.030 & 0.030 & 0.030 & 0.850 & 0.850 & 0.850 & 0.850 & 0.872 & 0.872 & 0.872 & 0.872 \\
  & 10.2 & \textbf{0.010} & 0.025 & 0.030 & 0.015 & 0.820 & 0.820 & 0.820 & 0.820 & 0.856 & 0.863 & \textbf{0.869} & 0.863 \\
  & 31.5 & 0.015 & 0.015 & 0.020 & 0.015 & 0.830 & 0.820 & \textbf{0.870} & 0.820 & 0.863 & 0.843 & \textbf{0.885} & 0.856 \\
  & 51.8 & \underline{0.010} & \underline{0.010} & 0.040 & 0.020 & 0.830 & \textbf{0.840} & 0.800 & 0.800 & 0.859 & \underline{0.863} & 0.856 & \underline{0.863} \\
  & 83.5 & 0.010 & 0.015 & 0.780 & \textbf{0.005} & \textbf{0.840} & 0.780 & 0.590 & 0.800 & 0.824 & 0.853 & 0.674 & \textbf{0.859} \\
  & 96.2 & \textbf{0.035} & 0.045 & 0.965 & 0.660 & \textbf{0.830} & 0.800 & 0.220 & 0.670 & 0.827 & \textbf{0.831} & 0.281 & 0.757 \\
  & 99.3 & 0.885 & \textbf{0.785} & 0.995 & 0.955 & 0.740 & \textbf{0.800} & 0.280 & 0.180 & \textbf{0.812} & 0.792 & 0.377 & 0.262 \\
  & 99.7 & 0.955 & \textbf{0.920} & 0.995 & 0.990 & \textbf{0.760} & 0.520 & 0.100 & 0.090 & \textbf{0.827} & 0.642 & 0.131 & 0.109 \\
\midrule
 \multirow{8}{*}{NTP} & 0.0 & 0.000 & 0.000 & 0.000 & 0.000 & 0.830 & 0.830 & 0.830 & 0.830 & 0.882 & 0.882 & 0.882 & 0.882 \\
  & 10.1 & 0.000 & 0.000 & 0.000 & 0.000 & \textbf{0.840} & 0.810 & 0.820 & 0.790 & 0.891 & \textbf{0.895} & 0.879 & 0.885 \\
  & 28.0 & 0.000 & 0.000 & 0.000 & 0.000 & 0.760 & 0.810 & \textbf{0.830} & 0.800 & 0.866 & 0.882 & \underline{0.885} & \underline{0.885} \\
  & 44.8 & 0.000 & 0.000 & 0.000 & 0.000 & \textbf{0.830} & 0.820 & 0.810 & 0.820 & 0.850 & 0.869 & 0.827 & \textbf{0.885} \\
  & 77.9 & 0.000 & 0.000 & 0.120 & 0.000 & 0.750 & 0.820 & 0.780 & \textbf{0.830} & 0.812 & 0.853 & 0.789 & \textbf{0.859} \\
  & 92.6 & \underline{0.000} & \underline{0.000} & 0.970 & 0.015 & 0.740 & 0.790 & 0.660 & \textbf{0.810} & 0.789 & \textbf{0.853} & 0.754 & 0.847 \\
  & 97.9 & \underline{0.000} & \underline{0.000} & 0.955 & 0.320 & 0.740 & 0.740 & 0.410 & \textbf{0.810} & 0.799 & \textbf{0.805} & 0.546 & 0.770 \\
  & 99.6 & 0.880 & \textbf{0.320} & 1.000 & 0.965 & 0.610 & \textbf{0.640} & 0.040 & 0.230 & 0.693 & \textbf{0.732} & 0.086 & 0.348 \\
\midrule
 \multirow{8}{*}{PO} & 0.0 & 0.000 & 0.000 & 0.000 & 0.000 & 0.810 & 0.810 & 0.810 & 0.810 & 0.866 & 0.866 & 0.866 & 0.866 \\
  & 9.2 & \underline{0.000} & \underline{0.000} & 0.005 & 0.005 & \underline{0.830} & \underline{0.830} & 0.820 & 0.790 & 0.853 & \textbf{0.872} & 0.859 & 0.866 \\
  & 28.2 & 0.000 & 0.000 & 0.000 & 0.005 & \underline{0.830} & 0.790 & \underline{0.830} & 0.780 & 0.850 & 0.834 & 0.837 & \textbf{0.863} \\
  & 44.9 & 0.005 & 0.005 & 0.005 & \textbf{0.000} & \textbf{0.840} & 0.820 & 0.800 & 0.810 & 0.850 & 0.837 & 0.812 & \textbf{0.856} \\
  & 76.1 & \underline{0.000} & \underline{0.000} & 0.560 & 0.040 & 0.780 & \textbf{0.810} & 0.770 & 0.790 & \underline{0.853} & \underline{0.853} & 0.834 & 0.840 \\
  & 92.1 & \underline{0.000} & \underline{0.000} & 0.925 & 0.400 & 0.780 & \textbf{0.810} & 0.520 & 0.760 & 0.802 & \textbf{0.831} & 0.639 & 0.751 \\
  & 97.6 & 0.125 & \textbf{0.010} & 0.995 & 0.975 & 0.770 & \textbf{0.820} & 0.160 & 0.530 & 0.719 & \textbf{0.850} & 0.272 & 0.518 \\
  & 99.4 & 0.980 & \textbf{0.850} & 0.995 & 0.985 & \textbf{0.550} & 0.370 & 0.030 & 0.090 & \textbf{0.521} & 0.403 & 0.083 & 0.102 \\
\bottomrule
\end{tabular}
\end{table*}

\begin{table*}[htbp]
\centering
\caption{Sparsification results for gradient-based (notated Grad.), IE, random dropout (Drop.), and bottom-$k$ (Bot-$k$) on Qwen 3 8B. ASR reported across three datasets. Sparsity percentage is notated as S (\%).
$\downarrow$ = lower is better (Alpaca); $\uparrow$ = higher is better (JailbreakBench, StrongReject).}
\label{tab:sparsity_q8}
\small
\begin{tabular}{ll |cccc| cccc| cccc}
\toprule
& & \multicolumn{4}{c}{Alpaca ($\downarrow$)} & \multicolumn{4}{c}{JailbreakBench ($\uparrow$)} & \multicolumn{4}{c}{StrongReject ($\uparrow$)} \\
\cmidrule(lr){3-6} \cmidrule(lr){7-10} \cmidrule(lr){11-14}
Vector & S (\%) & Grad. & IE & Drop. & Bot-$k$ & Grad. & IE & Drop. & Bot-$k$ & Grad. & IE & Drop. & Bot-$k$ \\
\midrule
 \multirow{8}{*}{DIM} & 0.0 & 0.015 & 0.015 & 0.015 & 0.015 & 0.830 & 0.830 & 0.830 & 0.830 & 0.888 & 0.888 & 0.888 & 0.888 \\
  & 12.6 & \textbf{0.005} & 0.010 & 0.015 & 0.010 & \textbf{0.890} & 0.830 & 0.820 & 0.830 & \textbf{0.920} & 0.904 & 0.901 & 0.891 \\
  & 36.4 & \textbf{0.000} & \textbf{0.000} & 0.035 & 0.005 & 0.840 & \textbf{0.870} & 0.850 & 0.860 & 0.891 & \textbf{0.917} & 0.879 & 0.904 \\
  & 56.1 & 0.010 & \textbf{0.005} & 0.120 & 0.015 & 0.820 & \textbf{0.850} & 0.770 & 0.840 & 0.895 & \textbf{0.901} & 0.815 & 0.879 \\
  & 87.8 & \textbf{0.000} & 0.015 & 0.950 & 0.075 & \underline{0.800} & \underline{0.800} & 0.500 & 0.780 & \textbf{0.863} & 0.856 & 0.559 & 0.856 \\
  & 97.0 & 0.195 & \textbf{0.050} & 0.980 & 0.430 & 0.730 & \textbf{0.790} & 0.150 & 0.590 & 0.783 & \textbf{0.847} & 0.157 & 0.690 \\
  & 99.4 & 0.960 & \textbf{0.870} & 0.990 & 0.885 & \textbf{0.490} & 0.460 & 0.130 & 0.470 & 0.546 & \textbf{0.553} & 0.163 & 0.498 \\
  & 99.9 & \textbf{0.960} & 0.995 & 1.000 & 0.975 & 0.270 & \textbf{0.430} & 0.040 & 0.400 & 0.428 & \textbf{0.486} & 0.035 & 0.444 \\
\bottomrule
\end{tabular}
\end{table*}

\section{Graph Details}
\label{appendix:B_graphs}
\subsection{Edge Details}
Gemma 2 2B has 8,790 steered edges when steering at layer 15. Llama 3.2 3B Instruct has 124,441 steered edges when steering at layer 12. Qwen 3 8B has 195,865 steered edges when steering at layer 20. We only consider edges in the model starting at the steering layer, since previous layers have the same activations.

\subsection{Graph Construction Algorithm}
\label{appendix:B:graphconstruction}
We follow the graph construction algorithm from \citet{mueller2025mibmechanisticinterpretabilitybenchmark}.
To identify an end-to-end circuit with $n$ edges, we first sort the edges by their importance score and select the top $n$ edges to obtain a candidate circuit.
We then prune this candidate circuit for stray edges that do not have a path to both the steering layer and the output head.
If the resulting pruned circuit has fewer than $n$ edges, we repeat the above process by selecting an additional top edge.

\citet{hanna2024have} proposed a graph construction algorithm by working backwards from the output head.
In practice, we find that this algorithm obtains nearly the same circuits as the algorithm used in \citet{mueller2025mibmechanisticinterpretabilitybenchmark}.

\subsection{Graphs Visualized}
\label{appendix:B:graphvis}

\begin{figure*}[htbp]
    \centering
    \includegraphics[width=0.7\linewidth]{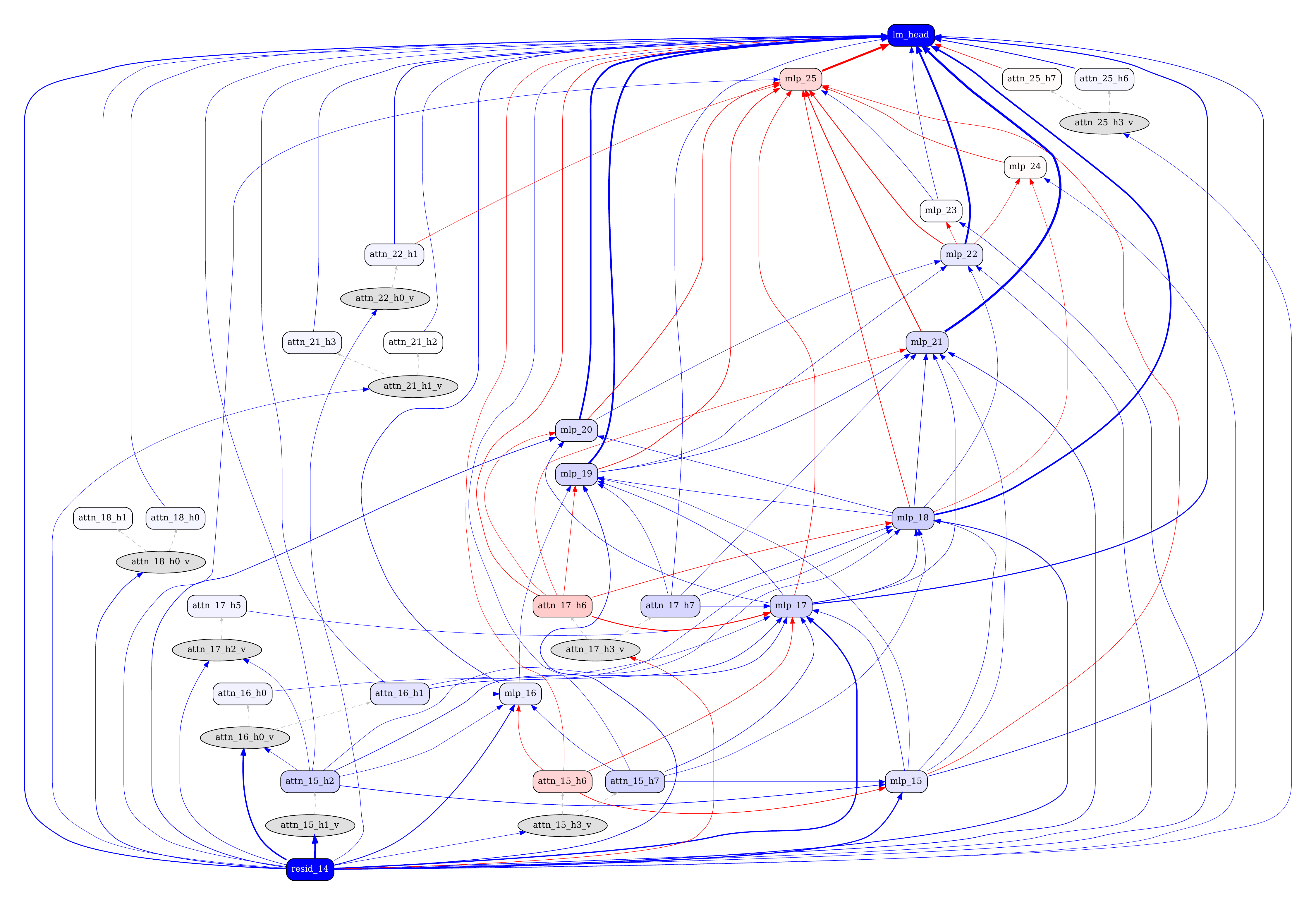}
    \caption{100 edge circuit for Gemma 2 2B DIM vector}
    \label{fig:graph_g2_dim_100}
\end{figure*}

\begin{figure*}[htbp]
    \centering
    \includegraphics[width=0.7\linewidth]{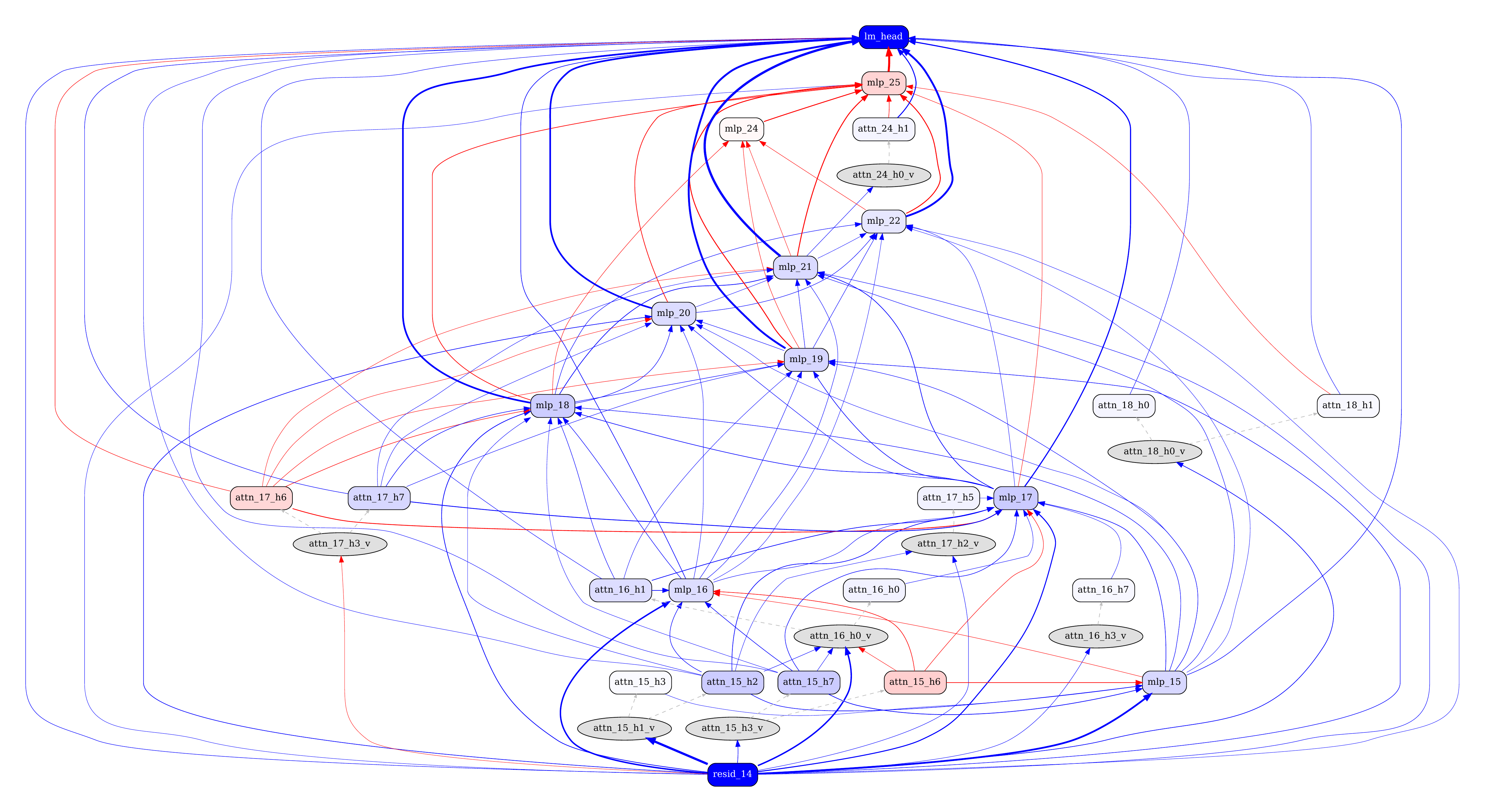}
    \caption{100 edge circuit for Gemma 2 2B NTP vector}
    \label{fig:graph_g2_ntp_100}
\end{figure*}

\begin{figure*}[htbp]
    \centering
    \includegraphics[width=0.7\linewidth]{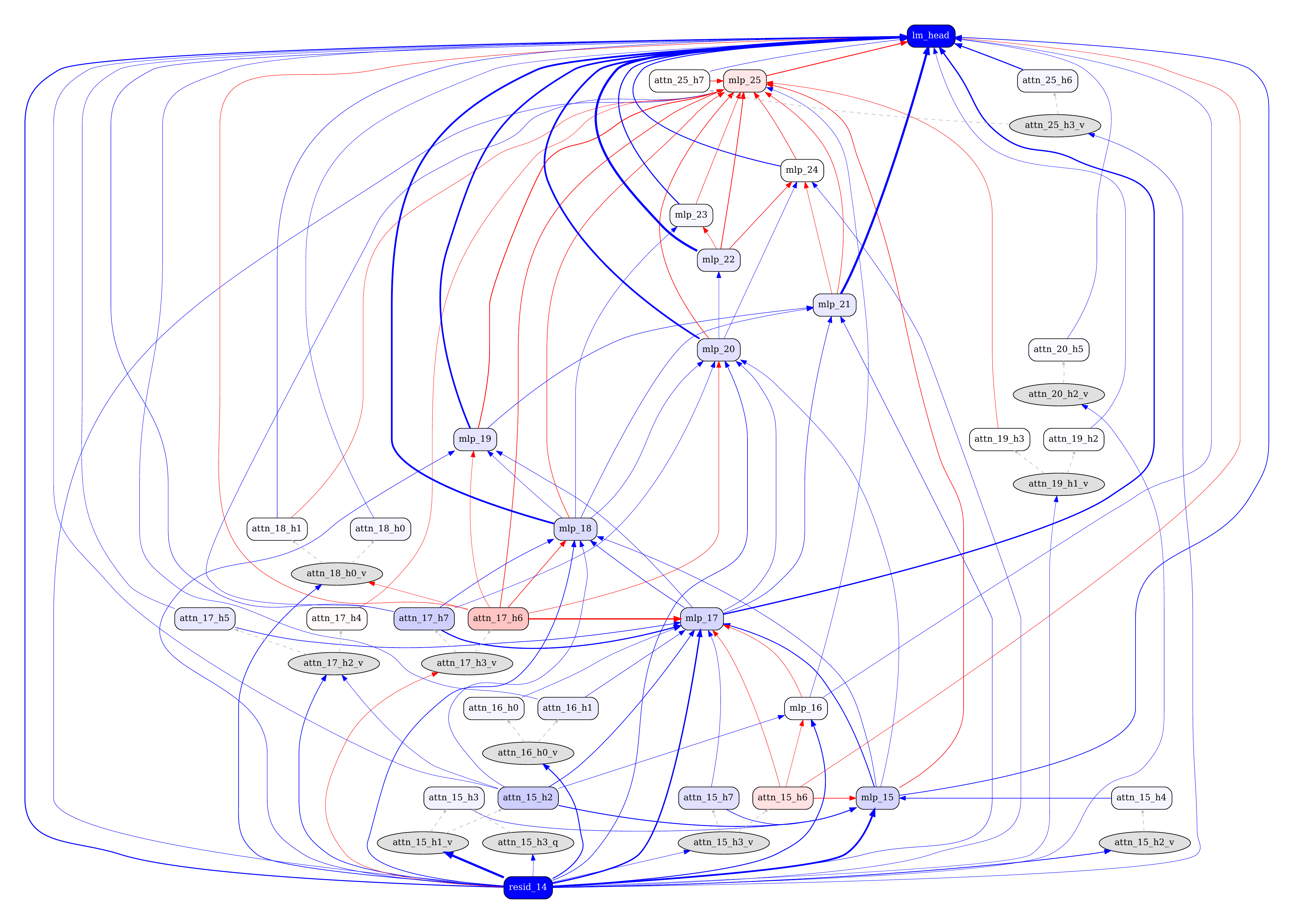}
    \caption{100 edge circuit for Gemma 2 2B PO vector}
    \label{fig:graph_g2_po_100}
\end{figure*}

\begin{figure*}[htbp]
    \centering
    \includegraphics[width=0.7\linewidth]{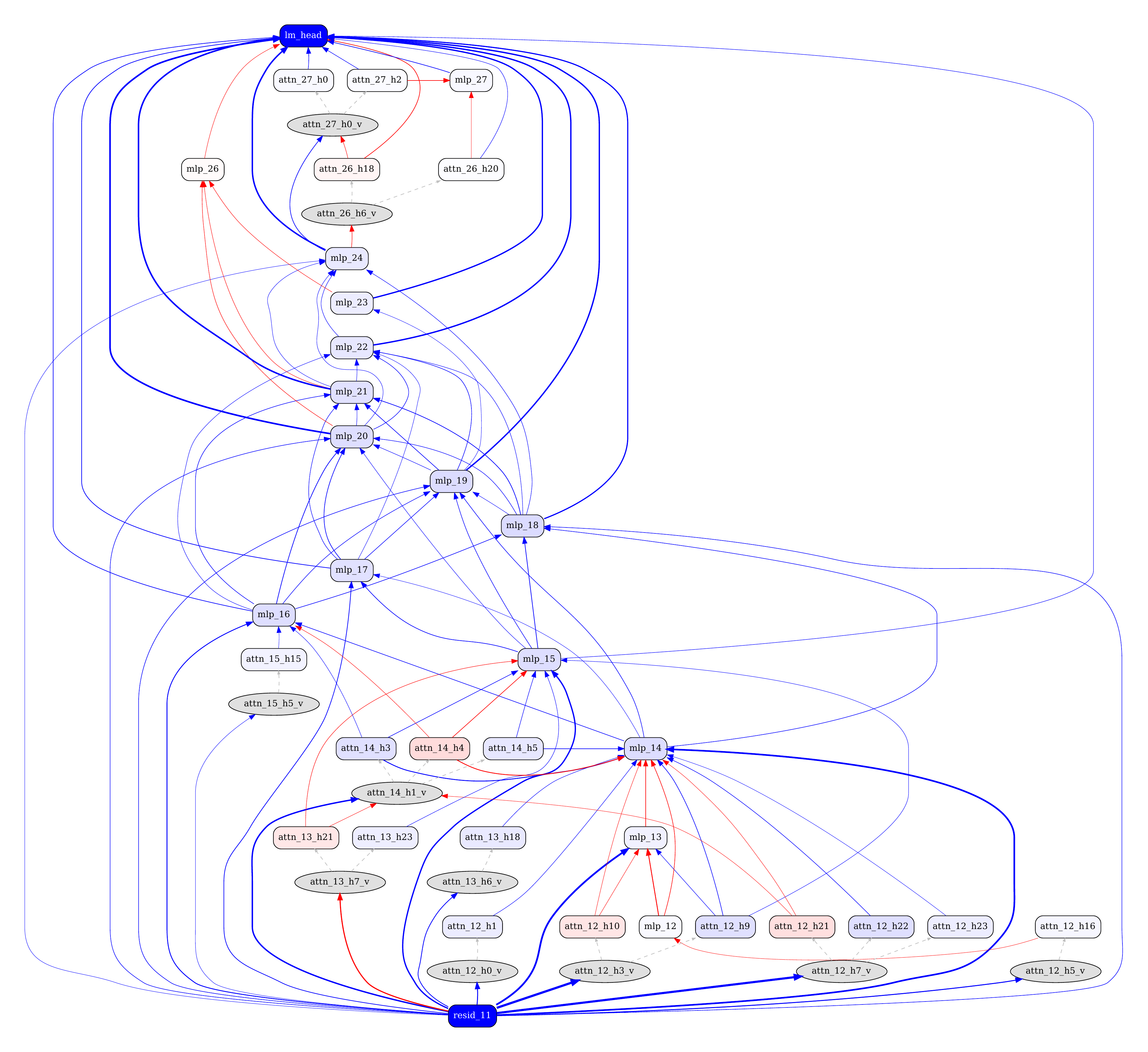}
    \caption{100 edge circuit for Llama 3.2 3B DIM vector}
    \label{fig:graph_l3_dim_100}
\end{figure*}

\begin{figure*}[htbp]
    \centering
    \includegraphics[width=0.7\linewidth]{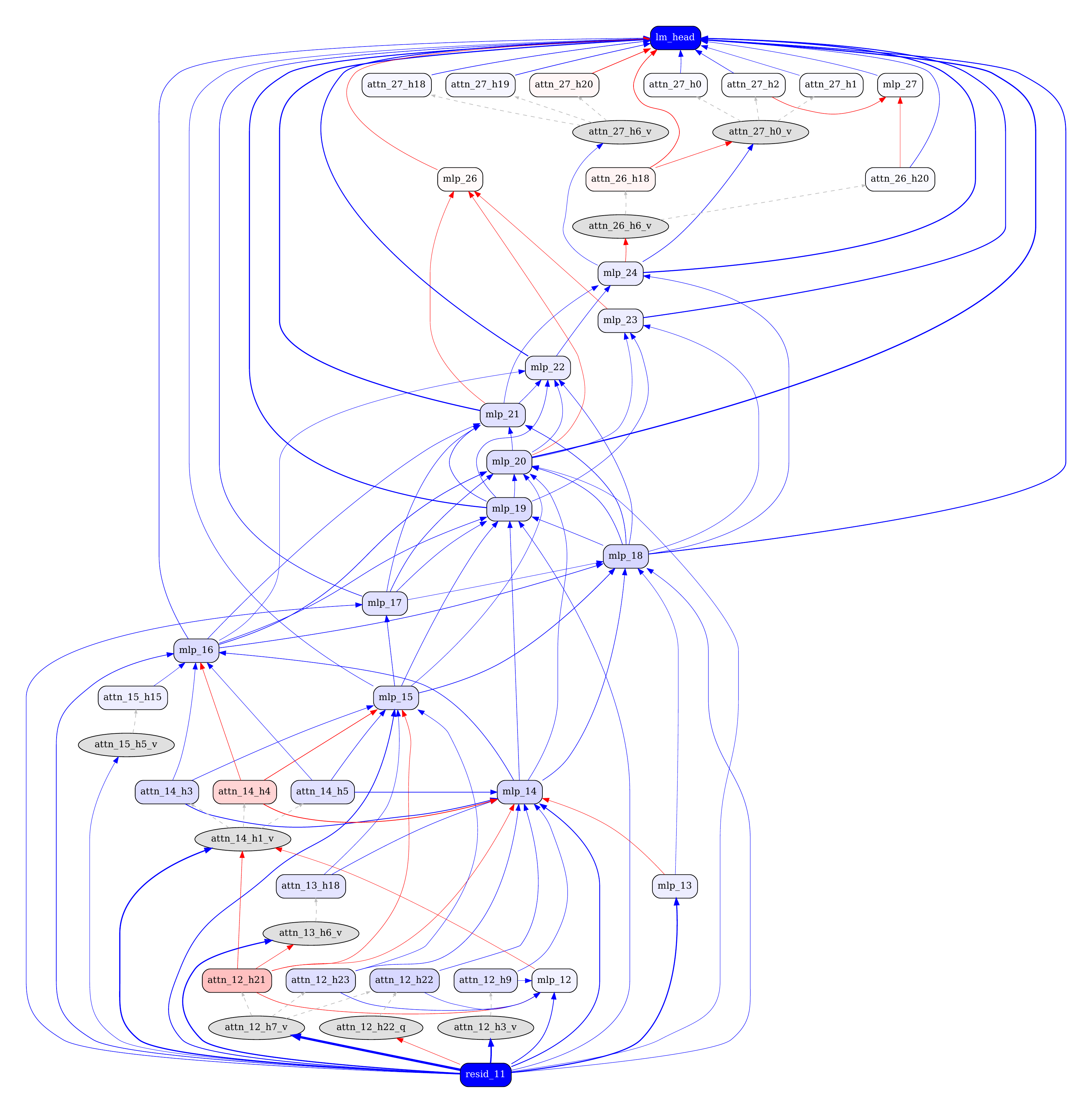}
    \caption{100 edge circuit for Llama 3.2 3B NTP vector}
    \label{fig:graph_l3_ntp_100}
\end{figure*}

\begin{figure*}[htbp]
    \centering
    \includegraphics[width=0.7\linewidth]{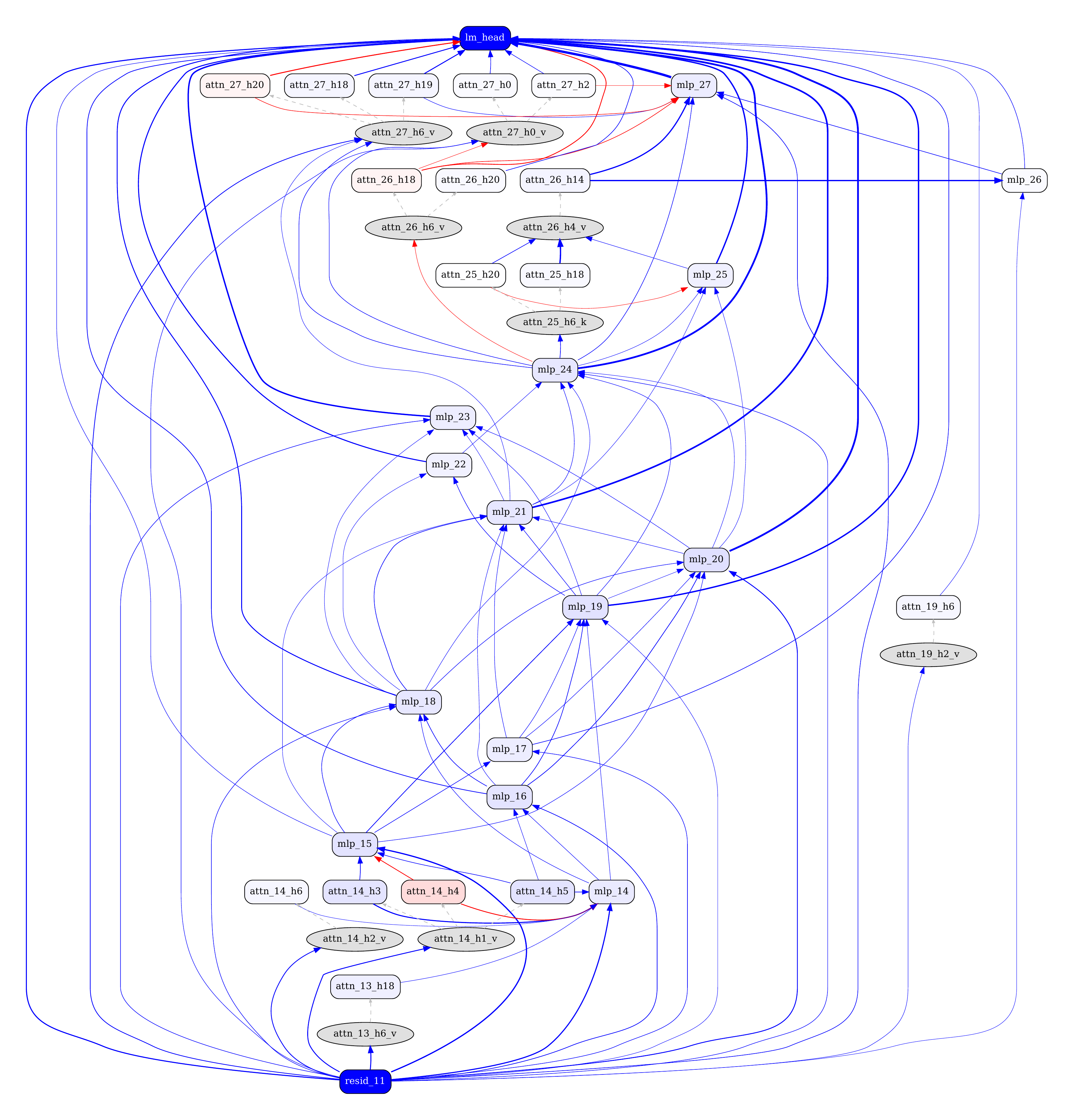}
    \caption{100 edge circuit for Llama 3.2 3B PO vector}
    \label{fig:graph_l3_po_100}
\end{figure*}

\begin{figure*}[htbp]
    \centering
    \includegraphics[width=0.7\linewidth]{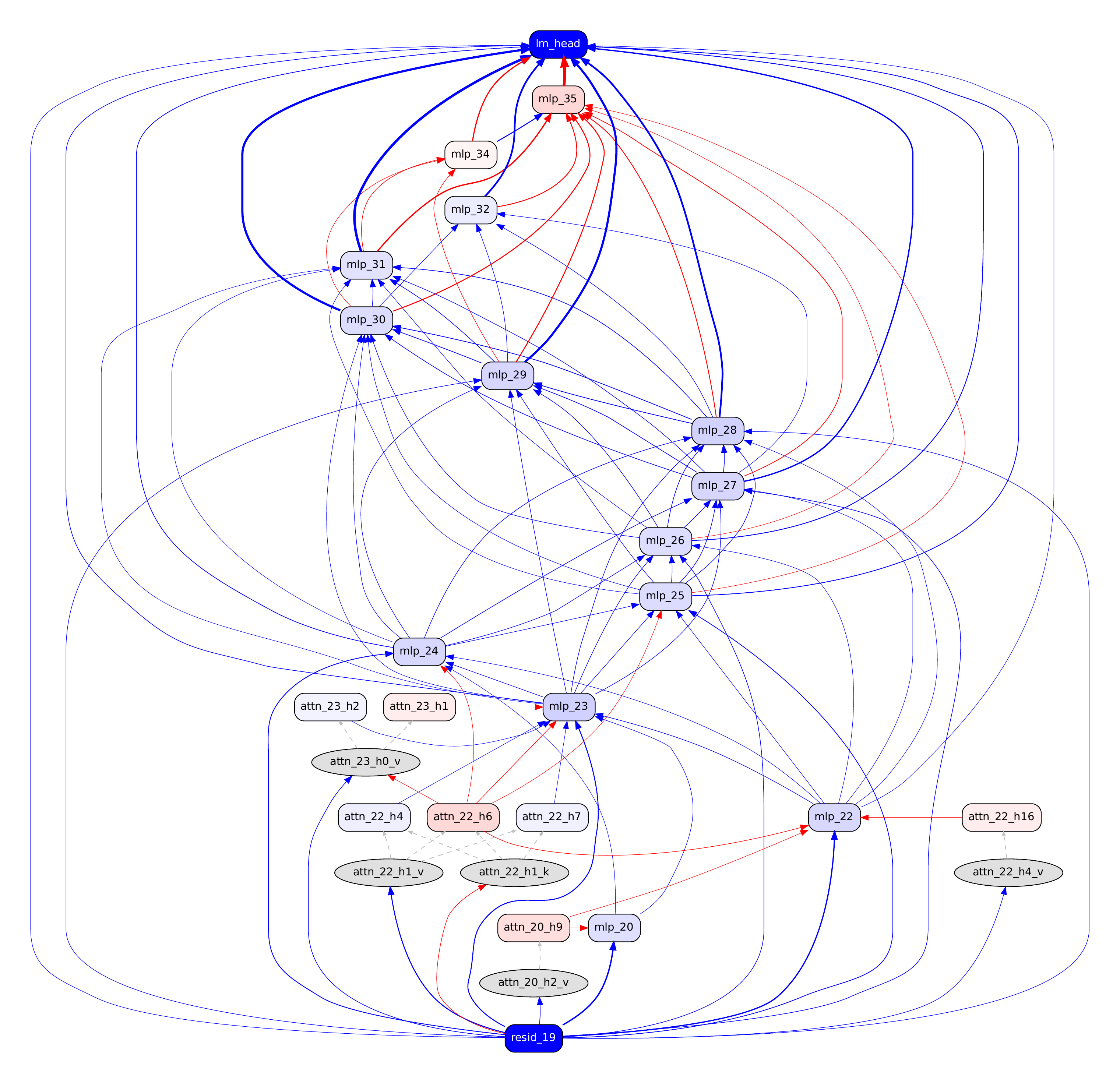}
    \caption{100 edge circuit for Qwen 3 8B DIM vector}
    \label{fig:graph_q8_dim_100}
\end{figure*}

We visualize 100-edge circuit graphs for Gemma 2 2B using the DIM, NTP, and PO vectors in Figures \ref{fig:graph_g2_dim_100}, \ref{fig:graph_g2_ntp_100}, and \ref{fig:graph_g2_po_100} respectively. We visualize a 100-edge circuit graph for the DIM, NTP, and PO vectors on Llama 3.2 3B in Figures \ref{fig:graph_l3_dim_100}, \ref{fig:graph_l3_ntp_100}, and \ref{fig:graph_l3_po_100} respectively.
We visualize a 100-edge circuit graph for the DIM vector on Qwen 3 8B in Figure \ref{fig:graph_q8_dim_100}.
Blue edges and nodes indicate positive $IE$, while red indicates negative $IE$. The color intensity is directly proportional to the magnitude of the $IE$.


\subsection{Edge Distribution}
A circuit's edges connect upstream nodes to downstream nodes. There are three types of upstream nodes: attention heads, MLPs, and the steering residual layer. There are five types of downstream nodes: attention queries, keys, and values, MLPs, and the LM Head. We record the total number of outgoing steered edges from each type of upstream node and incoming steered edges to each type of downstream node for each model in \autoref{tab:total-edge-counts}.

For each minimum-faithful circuit for Gemma 2 2B (900/8790 edges), Llama 3.2 3B (13500/124441 edges), and Qwen 3 8B (16100/195865 edges), we record the percentage of \textit{outgoing} edges that start from each type of upstream node in \autoref{tab:outgoing_edges} . 
We also record the percentage of \textit{incoming} edges to each type of downstream node in \autoref{tab:incoming_edges}.

Since not all edges in the circuit have equal importance and each node type significantly differ in the number of total possible edges (i.e. the number of edges to the attention queries far exceeds the number of edges to the LM head), we also inspect the top K edges by $IE$ score for each circuit. 
In the ``\% Top K" column of \autoref{tab:outgoing_edges}, we record the percentage of top K edges in the minimum-faithful circuit that start from each type of upstream node. We do the same for edges ending at each type of downstream node in the ``\% Top K" column of \autoref{tab:incoming_edges}. We choose $k$ values that correspond approximately to ~10\% of the number of edges in each minimum-faithful circuit.
When examining the downstream nodes of top edges in \autoref{tab:incoming_edges}, the incoming edges to the attention mechanism is heavily skewed towards the attention values. We expand upon this in \cref{sec:attention:frozen}.
\begin{table}[htbp]
    \centering
    \small
    \begin{tabular}{c|c|c}
        \toprule
        Model & Module & Count \\
        \toprule
        \multirow{9}{*}{\makecell{Llama 3.2 3B \\ Layer 12}} & Upstream Attn & 118,848 \\
        & Upstream MLP & 4,936 \\
        & Upstream Resid & 657 \\
        \cmidrule{2-3}
        & Downstream Attn Q & 72,384  \\
        & Downstream Attn K & 24,128 \\
        & Downstream Attn V & 24,128 \\
        & Downstream MLP & 3,400 \\
        & Downstream LM Head & 401 \\
        \midrule
        \multirow{9}{*}{\makecell{Gemma 2 2B \\Layer 15}} & Upstream Attn & 7,656 \\
        & Upstream MLP & 946 \\
        & Upstream Resid & 188 \\
        \cmidrule{2-3}
        & Downstream Attn Q & 4,048 \\
        & Downstream Attn K & 2,024 \\
        & Downstream Attn V & 2,024 \\
        & Downstream MLP & 594 \\
        & Downstream LM Head & 100 \\
        \midrule
        \multirow{9}{*}{\makecell{Qwen 3 8B \\Layer 20}} & Upstream Attn & 189,184 \\
        & Upstream MLP & 5,896 \\
        & Upstream Resid & 785 \\
        \cmidrule{2-3}
        & Downstream Attn Q & 127,232  \\
        & Downstream Attn K & 31,808 \\
        & Downstream Attn V & 31,808 \\
        & Downstream MLP & 4,488 \\
        & Downstream LM Head & 529 \\
        \toprule
    \end{tabular}
    \caption{Number of steered edges from upstream nodes and to downstream nodes for each model and steering layer.}
    \label{tab:total-edge-counts}
\end{table} 
\begin{table*}[htbp]
    \centering
    \small
    \begin{subtable}[t]{0.49\textwidth}
    \centering
    \begin{tabular}{c|c|c|c}
         \toprule
         M $\times$ V & Upstream Node & \% Circuit & \% Top K \\
         \toprule
         \multirow{3}{*}{L3 DIM} & Attn  & 79.9\% & 65.7\%\\
         &MLP &  17.4\% & 24.4\% \\
         & Resid & 2.8\% & 9.9\% \\
         \midrule
         \multirow{3}{*}{L3 NTP} & Attn  & 79.1\% & 65.2\% \\
         & MLP &  17.5\% & 24.7\% \\
         & Resid & 3.4\% & 10.1\% \\
         \midrule
         \multirow{3}{*}{L3 PO} & Attn  & 79.9\% & 63.6\% \\
         & MLP &  16.4\% & 22.1\% \\
         & Resid & 3.8\% & 14.3\% \\
         \midrule
         \multirow{3}{*}{G2 DIM} & Attn & 71.6\% & 48.0\% \\
         & MLP & 20.6\% & 28.0\% \\
         & Resid & 7.9\% & 24.0\% \\
         \midrule
         \multirow{3}{*}{G2 NTP} & Attn  & 69.4\% & 48.0\% \\
         & MLP &  22.1\% & 34.0\% \\
         & Resid  & 8.4\% & 18.0\% \\
         \midrule
         \multirow{3}{*}{G2 PO} & Attn  & 70.0\% & 50.0\%\\
         & MLP & 21.8\% & 27.0\%\\
         & Resid & 8.2\% & 23.0\% \\
         \midrule
         \multirow{3}{*}{Q8 DIM} & Attn & 82.2\% & 72.1\%\\
         & MLP & 14.7\% & 20.2\% \\
         & Resid & 3.0\% & 7.7\% \\
         \toprule
    \end{tabular}
    \caption{Outgoing edges from upstream nodes}
    \label{tab:outgoing_edges}
    \end{subtable}
    \begin{subtable}[t]{0.49\textwidth}
    \centering
    \begin{tabular}{c|c|c|c}
    \toprule
    Model & Downstream Node & \% Circuit &  \% Top K \\
    \toprule
    \multirow{5}{*}{L3 DIM} & Attn Q & 35.1\% & 5.7\%\\
    & Attn K &  8.7\% & 1.8\%\\
    & Attn V & 34.1\% & 31.9\%\\
    & MLP & 19.7\% & 52.8\% \\
    & Head & 2.4\% & 7.8\% \\
    \midrule
    \multirow{5}{*}{L3 NTP} & Attn Q & 37.1\% & 6.5\% \\
    & Attn K &  8.7\% & 1.1\% \\
    & Attn V & 32.4\% & 32.5\%\\
    & MLP & 19.4\% & 51.6\% \\
    & Head & 2.4\% & 8.3\% \\
    \midrule
    \multirow{5}{*}{L3 PO} & Attn Q & 33.4\% & 9.0\% \\
    & Attn K &  8.7\% & 1.8\% \\
    & Attn V & 36.4\% & 33.0\% \\
    & MLP & 19.1\% & 47.3\% \\
    & Head & 2.5\% & 8.9\% \\
    \midrule
    \multirow{5}{*}{G2 DIM} & Attn Q & 9.1\% & 0.0\% \\
    & Attn K & 0.6\% & 0.0\% \\
    & Attn V & 36.6\% & 16.0\% \\
    & MLP &  43.6\% & 50.0\% \\
    & Head & 10.2\% & 34.0\% \\
    \midrule
    \multirow{5}{*}{G2 NTP} & Attn Q  & 10.7\% & 0.0\% \\
    & Attn K & 1.3\% & 0.0\% \\
    & Attn V  & 34.1\% & 12.0\% \\
    & MLP  & 43.4\% & 57.0\% \\
    & Head & 10.4\% & 31.0\% \\
    \midrule
    \multirow{5}{*}{G2 PO} & Attn Q &  11.9\% & 1.0\% \\
    & Attn K & 1.8\% & 0.0\% \\
    & Attn V & 33.4\% & 13.0\% \\
    & MLP & 42.7\% &48.0\% \\
    & Head & 10.2\% &38.0\% \\
    \midrule
    \multirow{5}{*}{Q8 DIM} & Attn Q & 37.0\% & 4.9\% \\
    & Attn K & 6.5\% & 0.9\% \\
    & Attn V & 29.2\% & 24.0\% \\
    & MLP &  24.2\% & 56.0\% \\
    & Head & 3.1\% & 14.2\% \\
    \toprule
    \end{tabular}
    \caption{Incoming edges to downstream nodes.}
    \label{tab:incoming_edges}
    \end{subtable}
    \caption{Outgoing edges from upstream nodes and incoming edges to downstream nodes on minimum-faithful circuits. \% Circuit and \% Top K show the percent of the circuit and percent of top $k$ edges that start from each node type. $k=100$ for Gemma 2 2B, $k=1000$ for Llama 3.2 3B, and $k=1500$ for Qwen 3 8B.}
    \label{tab:edge_distributions}
\end{table*}

\section{Logit Lens}
Figures \ref{fig:svv_heatmap_g2}, \ref{fig:svv_heatmap_l3}, \ref{fig:svv_heatmap_q8} show manually selected logit lens tokens for each attention head svv. 
For transparency, we show the unfiltered top-20 logit lens tokens of the attention head svvs in \autoref{tab:logit-lens-full}. The tokens are sorted in descending order by logit value.

\newcommand{\lbsize}{\fontsize{7}{8.4}\selectfont} 

\begin{table*}[t]
\centering
\footnotesize
\setlength{\tabcolsep}{4pt}
\renewcommand{\arraystretch}{1.15}
\begin{tabular}{@{}>{\lbsize}l >{\lbsize}l >{\raggedright\arraybackslash\ttfamily\scriptsize}p{0.9\linewidth}@{}}
\toprule
Model & Head & Top-20 logit lens tokens \\
\midrule
\multirow{7}{*}{\shortstack[l]{G2 NTP}}
& L15H2 & `\bt{ seriously}', `\bt{ Seriously}', `\bt{NEVER}', ` never', `ímite', ` NEVER', ` rara', ` rarely', `rare', ` rare', ` frankly', `érias', ` Rarely', `Never', ` inac', ` \faThumbsDown', ` noy', `fatalError', ` sacri', `never' \\
\addlinespace 
& L16H1 & `\bt{ prohibition}', `\#\#\#\#\#\#\#\#.', `DockStyle', `\bt{ forbidding}', ` prohibiting', ` prohibit', ` prevention', ` Prohibition', ` forbid', `\bt{ safety}', ` prohibits', ` dissu', `hibition', ` Prevention', ` prohibited', `prevention', `pung', \begin{CJK*}{UTF8}{gbsn}`禁'\end{CJK*}, ` security', ` protest' \\
\addlinespace
& L17H7 & `ciudad', ` dis', `ÔNG', `\bt{ never}', `riffe', `seille', `agal', ` kna', `städ', ` unseen', `\bt{ ephemeral}', `jana', ` tr\`u', ` Allgemeinen', ` jButton', `$\lambda o \nu$', `Ciudad', `\includegraphics[height=1em]{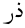}', `\bt{ malignant}', ` uzak' \\
\addlinespace
& L22H1 & `\bt{ integrity}', `TRUNC', ` détour', `\bt{ honesty}', ` subversion', `Humor', `denk', ` contentment', ` contented', ` comfortable', `integrity', ` straight', ` comfort', ` tristesse', ` tempted', ` equilibrium', ` equilibrio', ` honnête', `\bt{ subversive}', `Sad' \\
\addlinespace
& L23H0 & `\bt{ wouldn}', `\textquotedblleft', `\textnormal{\textquotedbl}!', ` keine', `\bt{ don}', `\bt{ isn}', ` geen', `Šaltiniai', `!("', `rawDesc', `\{"', ` !"', ` ligiloj', `!', `"!', ` requested', `(“', ` não', `none', ` doesn' \\
\addlinespace
& Summed & `\bt{\#}', `FormTagHelper', `\bt{ sickening}', `\textless eos\textgreater', `(', `*', `“', `\includegraphics[height=1em]{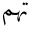}',
\begin{CJK*}{UTF8}{gbsn}`つく'\end{CJK*}, 
`\textbackslash n\textbackslash n', `\textasciicircum', ` leher', `rawDesc', ` teto', ` konu', `CLC', ` po\v{z}i', `?', \begin{CJK*}{UTF8}{gbsn} `\bt{对不起}'\end{CJK*}, \begin{CJK*}{UTF8}{mj}` 귀'\end{CJK*} \\
\addlinespace
& Steer vec. & `\bt{ disambiguazione}', `\bt{httphttps}', `\bt{Lähteet}', ` bpy', `StoryboardSegue', `yafet', ` Newt', ` categorically', ` LSA', `FormTagHelper', `/\textbackslash r', ` Dudley', `ześnie', ` analytically', ` BoxDecoration', ` Canva', ` betweenstory', ` ADM', `tanleria', ` ECA' \\
\midrule
\multirow{8}{*}{\shortstack[l]{L3 DIM}}
& L12H9 & `reading', ` Reading', ` Wein', ` reading', `\bt{rape}', ` partners', `\bt{ alarming}', `\bt{ warning}', ` Warning', `Code', ` Shall', `loth', ` Controlled', ` Bam', ` Partners', `Reading', `TP', `iminal', ` alarmed', ` distress' \\
\addlinespace
& L12H22 & ` raw', `\bt{ dyst}', `\bt{ legal}', ` challenging', ` discharged', ` cri', ` Legal', ` discharge', ` legally', ` filed', ` Hollow', ` materially', ` surreal', `üh', `[U+FFFD]', `avage', `okol', `\bt{ destruction}', `cth', `agara' \\
\addlinespace
& L13H18 & `\bt{ dark}', ` darker', ` replic', ` mature', ` bur', `bourg', `ocker', ` fu', ` suspension', `\bt{ Suspension}', ` Dark', ` burg', `\bt{ devil}', ` h\~ap', ` underground', ` Helena', ` Monkey', ` abs', `sto', `fu' \\
\addlinespace
& L14H3 & ` verb', `\bt{ ethics}', ` ethical', \begin{CJK*}{UTF8}{gbsn}`是不'\end{CJK*}, `aka', `\bt{ forbidden}', ` transcripts', ` Ethics', `rian', ` Ned', `WARNING', ` Jed', ` DO', ` passion', `\bt{ prohibited}', `q', ` stage', `zy', ` transcription', `agon' \\
\addlinespace
& L14H19 & ` Rif', `\bt{ forfeiture}', ` Pent', ` youth', ` legalization', ` kin', ` Streets', ` forfe', ` flight', `Blank', `Kin', `\bt{ victim}', `\bt{ Criminal}', ` Panc', ` Kin', `sage', ` Dove', \begin{CJK*}{UTF8}{gbsn}`密'\end{CJK*}, ` bench', `rone' \\
\addlinespace
& L15H15 & `\bt{ forbidden}', `\bt{ prohibited}', ` neither', ` forbid', `Neither', `\bt{ cannot}', ` forb', ` technically', ` Neither', ` Forbidden', ` strictly', ` nor', ` dread', ` bro', ` taboo', `again', ` ce', `ombre', `Forbidden', `Again' \\
\addlinespace
& Summed & `\bt{ safer}', \begin{CJK*}{UTF8}{gbsn}`遍'\end{CJK*}, ` Pek', `JECT', `\bt{ survival}', ` puberty', `ervers', ` antis', ` shitty', ` Survival', ` fucking', `CLICK', ` Leer', ` graves', ` Hardcore', `yeah', `\bt{ unlawful}', ` surviv', ` clandest', ` livelihood' \\
\addlinespace
& Steer vec. & `istan', `\bt{ nowhere}', `\bt{ sympath}', ` Raymond', ` slang', `ayas', ` ray', `\bt{ tone}', ` \_:', `urg', ` Hollow', `ibu', ` hollow', \begin{CJK*}{UTF8}{gbsn}`弥'\end{CJK*}, `Ray', \begin{CJK*}{UTF8}{mj}` 말이'\end{CJK*}, `Offsets', ` wann', ` surg', `Offset' \\
\midrule
\multirow{9}{*}{\shortstack[l]{Q8 DIM}}
& -L20H9 & \begin{CJK*}{UTF8}{gbsn}`\bt{关停}', `严肃'\end{CJK*}, `\bt{Reject}', \begin{CJK*}{UTF8}{gbsn}`卫生健康'\end{CJK*}, `\_ABORT', ` Reject', `\bt{ Forbidden}', ` negatives', `\_abort', `\textcyrillic{ отказ}', \begin{CJK*}{UTF8}{gbsn}`撤销', `负面'\end{CJK*}, ` discourage', ` anonymously', `TU', `oose', ` Crisis', `Negative', `unload', \begin{CJK*}{UTF8}{bsmi}`拒絕'\end{CJK*} \\
\addlinespace
& L20H10 & \begin{CJK*}{UTF8}{gbsn}`\bt{不应该}', `不应', `\bt{根本不}'\end{CJK*}, ` infra', \begin{CJK*}{UTF8}{gbsn}`违反', `也无法'\end{CJK*}, `infra', \begin{CJK*}{UTF8}{gbsn}`宪'\end{CJK*}, `fails', \begin{CJK*}{UTF8}{gbsn}`根本就不'\end{CJK*}, ` unnecessarily', \begin{CJK*}{UTF8}{gbsn}`都不能', `应该'\end{CJK*}, ` \bt{shouldn'}, `.ShouldBe', \begin{CJK*}{UTF8}{gbsn}`也不能', `根本没有'\end{CJK*}, `Lf', `aine', \begin{CJK*}{UTF8}{gbsn}`不合理'\end{CJK*} \\
\addlinespace
& L22H4 & \begin{CJK*}{UTF8}{gbsn}`\bt{严禁}', `禁止'\end{CJK*}, `\bt{ forbidden}', `\bt{ prohibited}', ` illegal', \begin{CJK*}{UTF8}{gbsn}`不允许'\end{CJK*}, `illegal', \begin{CJK*}{UTF8}{gbsn}`合法', `非法'\end{CJK*}, ` Illegal', `Forbidden', ` safe', \begin{CJK*}{UTF8}{gbsn}`禁忌', `不得'\end{CJK*}, ` safely', ` illegally', ` unsafe', `Avoid', \begin{CJK*}{UTF8}{gbsn}`禁'\end{CJK*}, `NotAllowed' \\
\addlinespace
& -L22H6 & \begin{CJK*}{UTF8}{gbsn}`\bt{严禁}', `禁止'\end{CJK*}, `\bt{ forbidden}', `\bt{ prohibited}', \begin{CJK*}{UTF8}{gbsn}`不允许'\end{CJK*}, ` illegal', `illegal', `Forbidden', \begin{CJK*}{UTF8}{gbsn}`非法', `禁忌'\end{CJK*}, `Avoid', ` prohibition', ` Illegal', \begin{CJK*}{UTF8}{gbsn}`禁'\end{CJK*}, ` illegally', \begin{CJK*}{UTF8}{gbsn}`合法', `禁区'\end{CJK*}, ` unethical', ` legality', ` forbid' \\
\addlinespace
& L23H3 & \begin{CJK*}{UTF8}{gbsn}`\bt{违背}', `滥用'\end{CJK*}, `\bt{ endanger}', \begin{CJK*}{UTF8}{gbsn}`可能导致', `违反'\end{CJK*}, `\bt{ violates}', \begin{CJK*}{UTF8}{gbsn}`后果', `不应该', `擅自', `不应'\end{CJK*}, ` unethical', \begin{CJK*}{UTF8}{gbsn}`万一'\end{CJK*}, \begin{CJK*}{UTF8}{mj}` 잘못'\end{CJK*}, ` risking', ` irresponsible', \begin{CJK*}{UTF8}{gbsn}`涉嫌', `恶意', `任何形式', `严重的'\end{CJK*}, ` falsely' \\
\addlinespace
& L35H2 & \begin{CJK*}{UTF8}{gbsn}`\bt{但这}', `这也是', `另一种'\end{CJK*}, ` This', \begin{CJK*}{UTF8}{gbsn}`\bt{这不是}'\end{CJK*}, 
` \textcyrillic{но}', 
` Another', \begin{CJK*}{UTF8}{gbsn}`\bt{但它}'\end{CJK*}, ` These', `---but', \begin{CJK*}{UTF8}{gbsn}`这是我'\end{CJK*}, ` latter', \begin{CJK*}{UTF8}{gbsn}`这是'\end{CJK*}, `but', \begin{CJK*}{UTF8}{gbsn}`但他'\end{CJK*}, `This', ` but', ` That', \begin{CJK*}{UTF8}{gbsn}`但如果'\end{CJK*}, ` But' \\
\addlinespace
& Summed & `\bt{ illegal}', \begin{CJK*}{UTF8}{gbsn}`\bt{非法}', `合法'\end{CJK*}, `\bt{ legality}', ` illegally', ` Illegal', ` lawful', `illegal', \begin{CJK*}{UTF8}{gbsn}`禁止'\end{CJK*}, ` unlawful', ` illeg', ` prohibited', ` unethical', `Illegal', ` legitimate', ` legal', ` forbidden', ` harmless', \begin{CJK*}{UTF8}{gbsn}`的危害'\end{CJK*}, \begin{CJK*}{UTF8}{gbsn}`违反'\end{CJK*} \\
\addlinespace
& Steer vec. & `\bt{ illegal}', \begin{CJK*}{UTF8}{gbsn}`合法', `违法'\end{CJK*}, `\bt{ unethical}', `\bt{ illegally}', \begin{CJK*}{UTF8}{gbsn}`法律法规'\end{CJK*}, ` legality', \begin{CJK*}{UTF8}{gbsn}`非法', `违背'\end{CJK*}, ` Illegal', \begin{CJK*}{UTF8}{gbsn}`违法行为', `合法性'\end{CJK*}, ` violates', `illegal', ` illeg', `Illegal', ` ethical', `\_safe', ` prohibited', ` lawful' \\
\bottomrule
\end{tabular}
\caption{Unfiltered top-20 logit lens tokens for the attention svvs and the steering vector itself, sorted in descending order by logit value. Underlined tokens are those shown in the Figures \ref{fig:svv_heatmap_g2}, \ref{fig:svv_heatmap_l3}, and \ref{fig:svv_heatmap_q8}.}
\label{tab:logit-lens-full}
\end{table*}

\end{document}